\documentclass[review,12pt,3p]{elsarticle}

\makeatletter
\def\ps@pprintTitle{%
 \let\@oddhead\@empty
 \let\@evenhead\@empty
 \def\@oddfoot{\centerline{\thepage}}%
 \let\@evenfoot\@oddfoot}
\makeatother

\usepackage{times}
\usepackage{balance}
\usepackage{setspace}
\usepackage{algorithm}
\usepackage{algorithmic}
\usepackage{placeins}

\usepackage{boxedminipage} 
\usepackage{graphicx}
\usepackage{verbatim} 
\usepackage{amsmath}
\usepackage{times} % Add all your packages here
\usepackage{gensymb}
\usepackage{dsfont}
\usepackage{multirow}
\usepackage{amsfonts}
\usepackage{float}
\usepackage{mathrsfs} 
\usepackage[T1]{fontenc}
\usepackage{url}
\usepackage{amssymb}
\usepackage[tight,footnotesize]{subfigure}
\usepackage[caption=false,font=footnotesize]{subfig}
\usepackage{epsfig}
\usepackage{bm}

\begin{document}
	
	% Title Page
	\title{A DEEP analysis of the META-DES framework for dynamic selection of ensemble of classifiers}
	\author[ets]{Rafael M. O. Cruz\corref{corr1}}
	\ead{rmoc@cin.ufpe.br}
	
	\author[ets]{Robert Sabourin}
	\ead{robert.sabourin@etsmtl.ca}
	
	\author[ufpe]{George D. C. Cavalcanti}
	\ead{gdcc@cin.ufpe.br}
	
	\address[ets]{LIVIA, \'{E}cole de Technologie Sup\'{e}rieure, University of Quebec, Montreal, Que., Canada - www.livia.etsmtl.ca
	}
	\address[ufpe]{Centro de Inform\'{a}tica, Universidade Federal de Pernambuco, Recife, PE, Brazil - www.cin.ufpe.br/$\sim$viisar\\
	}
	
	\cortext[corr1]{Corresponding author. Email Address: rafaelmenelau@gmail.com.}
	
	\begin{abstract}		% Insert abstract here
		
		Dynamic ensemble selection (DES) techniques work by estimating the level of competence of each classifier from a pool of classifiers. Only the most competent ones are selected to classify a given test sample. Hence, the key issue in DES is the criterion used to estimate the level of competence of the classifiers in predicting the label of a given test sample. In order to perform a more robust ensemble selection, we proposed the META-DES framework using meta-learning, where multiple criteria are encoded as meta-features and are passed down to a meta-classifier that is trained to estimate the competence level of a given classifier. In this technical report, we present a step-by-step analysis of each phase of the framework during training and test. We show how each set of meta-features is extracted as well as their impact on the estimation of the competence level of the base classifier. Moreover, an analysis of the impact of several factors in the system performance, such as the number of classifiers in the pool, the use of different linear base classifiers, as well as the size of the validation data. We show that using the dynamic selection of linear classifiers through the META-DES framework, we can solve complex non-linear classification problems where other combination techniques such as AdaBoost cannot. 
	\end{abstract} 
	
	\begin{keyword}
		
		Ensemble of Classifiers, Dynamic Ensemble Selection, Classifier competence, Meta-Learning, Linear Classifiers
		
	\end{keyword}

\maketitle
\onehalfspacing

\section{Introduction}

Multiple Classifier Systems (MCS) aim to combine classifiers in order to increase the recognition accuracy in pattern recognition systems~\cite{kittler,kuncheva}. MCS are composed of three phases~\cite{Alceu2014}: (1) Generation, (2) Selection, and (3) Integration. In the first phase, a pool of classifiers is generated. In the second phase, a single classifier or a subset having the best classifiers of the pool is(are) selected. We refer to the subset of classifiers as the Ensemble of Classifiers (EoC). In the last phase, integration, the predictions of the selected classifiers are combined to obtain the final decision~\cite{kittler}.

Recent works in MCS have shown that dynamic ensemble selection (DES) techniques achieve higher classification accuracy when compared to static ones~\cite{Alceu2014,CruzPR,knora}. This is specially true for ill-defined problems, i.e., for problems where the size of the training data is small, and there are not enough data available to train the classifiers~\cite{paulo2,logid}. The key issue in DES is to define a criterion to measure the level of competence of a base classifier. Most DES techniques~\cite{knora,lca,mcb,ijcnn2011} estimate the classifiers' local accuracy in small regions of the feature space surrounding the query instance, called the region of competence, as a search criterion for estimating the competence level of the base classifier. However, in our previous work~\cite{ijcnn2011}, we demonstrated that the use of local accuracy estimates alone is insufficient to provide higher classification performance. In addition, a dissimilarity analysis among eight dynamic selection techniques, performed in~\cite{Cruz2014ANNPR}, indicates that techniques based on different criteria for estimating the competence level of base classifiers yields different results. 

To tackle this issue, in~\cite{CruzPR} we proposed a novel DES framework, called META-DES, in which multiple criteria regarding the behavior of a base classifier are used to compute its level of competence. The framework is based on two environments: the classification environment, in which the input features are mapped into a set of class labels, and the meta-classification environment, where several properties from the classification environment, such as the classifier accuracy in a local region of the feature space, are extracted from the training data and encoded as meta-features. Given a test data, the meta-features are extracted using the test data as reference, and used as input to the meta-classifier. The meta-classifier decides whether the base classifier is competent enough to classify the test sample. 

One interesting properties of the META-DES framework is that it obtains higher classification accuracy using only linear classifiers. In this work, we perform a deep analysis of the training and classification steps of the META-DES framework. We perform step-by-step examples in order to show the influence of different sets of meta-features used to better estimate the competence of the base classifier. The analysis is conducted using the P2 problem~\cite{Valentini05,henniges} which is a two-class non-linear problem with a complex decision boundary. Furthermore, the two-classes of the P2 problem have multiple class means, making it a difficult classification problem. 

The following points of the META-DES framework are studied:

\begin{itemize}
	
	\item The use of weak, linear classifiers in the pool. In this work we consider both Perceptrons and Decision Stumps as base classifiers.
%	\item Understanding how the sample selection mechanism embedded in the META-DES framework works.
	\item The influence of each set of meta-features for estimating the competence of a base classifier.
	
	\item The influence of the dynamic selection set (DSEL)\footnote{DSEL is often called validation data in several dynamic selection techniques.} in the recognition rate. The dynamic selection data is used in order to extract the meta-features.	
	
	\item The influence of the size of the Pool in the classification accuracy of the META-DES framework.
	
\end{itemize}

The contributions of this work are as follows:

\begin{itemize}
	
	\item It shows that using dynamic selection of linear and weak classifiers, such as Perceptrons and Decision stumps, we can solve problems with complex decision boundaries, including classification problems with multiple class centers.
	
	\item It allows an understanding of why the META-DES framework achieves high recognition accuracy using only linear classifiers. In previous, works the META-DES was presented as a black box system. In this work, we use a step-by-step example to illustrate how the framework is able to select the competent classifiers based on the five defined sets of meta-features.
	
	%\item size of the pool dsel influence
	
	\item It compares the dynamic selection of linear classifiers against static combination rules such as AdaBoost, as well as classical single classifier models, such as Multi-Layer Perceptron neural networks, Random Forest, and Support Vector Machnies (SVMs).

\end{itemize}

This document is organized as follows. Theoretical aspects of dynamic selection are introduced in Section~\ref{sec:dynamicSelection}. The META-DES framework is presented in Section~\ref{sec:proposed}. An illustrative example of the META-DES is presented in Section~\ref{sec:whyMETA}. Experiments are carried out in Section~\ref{sec:experiments}. Conclusions are given in the last section.

\section{Why does dynamic selection of linear classifiers work?}
\label{sec:dynamicSelection}

Let $C = \{c_{1}, \ldots, c_{M}\}$ ($M$ is the size of the pool of classifiers) be the pool of classifiers and $c_{i}$ a base classifier belonging to the pool $C$. The goal of dynamic selection is to find an ensemble of classifiers $C' \subset C$ that has the best classifiers to classify a given test sample $\mathbf{x}_{j}$. DES techniques rely on the assumption that each base classifier is an expert in a different local region of the feature space~\cite{zhu}. Only the classifiers that attain a certain competence level, according to a selection criterion, are selected to predict the label of $\mathbf{x}_{j}$. This is a different strategy when compared with static selection, where the ensemble of classifiers $C'$ is selected during the training phase, and considering the global performance of the base classifiers over a validation dataset~\cite{GiacintoR01,Cruz2012,multiga,greedy}. 
 
When dealing with dynamic selection, we aim to select the appropriate classifier(s) for a specific test sample $\mathbf{x}_{j}$, rather than find the best decision border separating the classes. This is a different concept, as compared to classical classification models, such as Support Vector Machines (SVM) or Multi-Layer Perceptrons (MLP) Neural Networks in the sense that these classifiers search for the best separation between the classes during the training stages. This is an important property of dynamic selection techniques, which makes them suitable for solving problems that are ill-defined, i.e., when there is not enough data available to train a strong classifier having a lot of parameters to learn~\cite{paulo2}. In addition, due to insufficient training data, the distribution of the training data may not adequately represent the real distribution of the problem. Consequently, the classifiers cannot learn the separation between the classes. %In our previous work~\cite{CruzPR} we show that DES techniques improve classification performance in an experiments considering 30 classification problems.

\begin{figure}[H]
	\centering
	\subfigure[Two circles representing the XOR problem]{\includegraphics[width=0.7\textwidth,clip=]{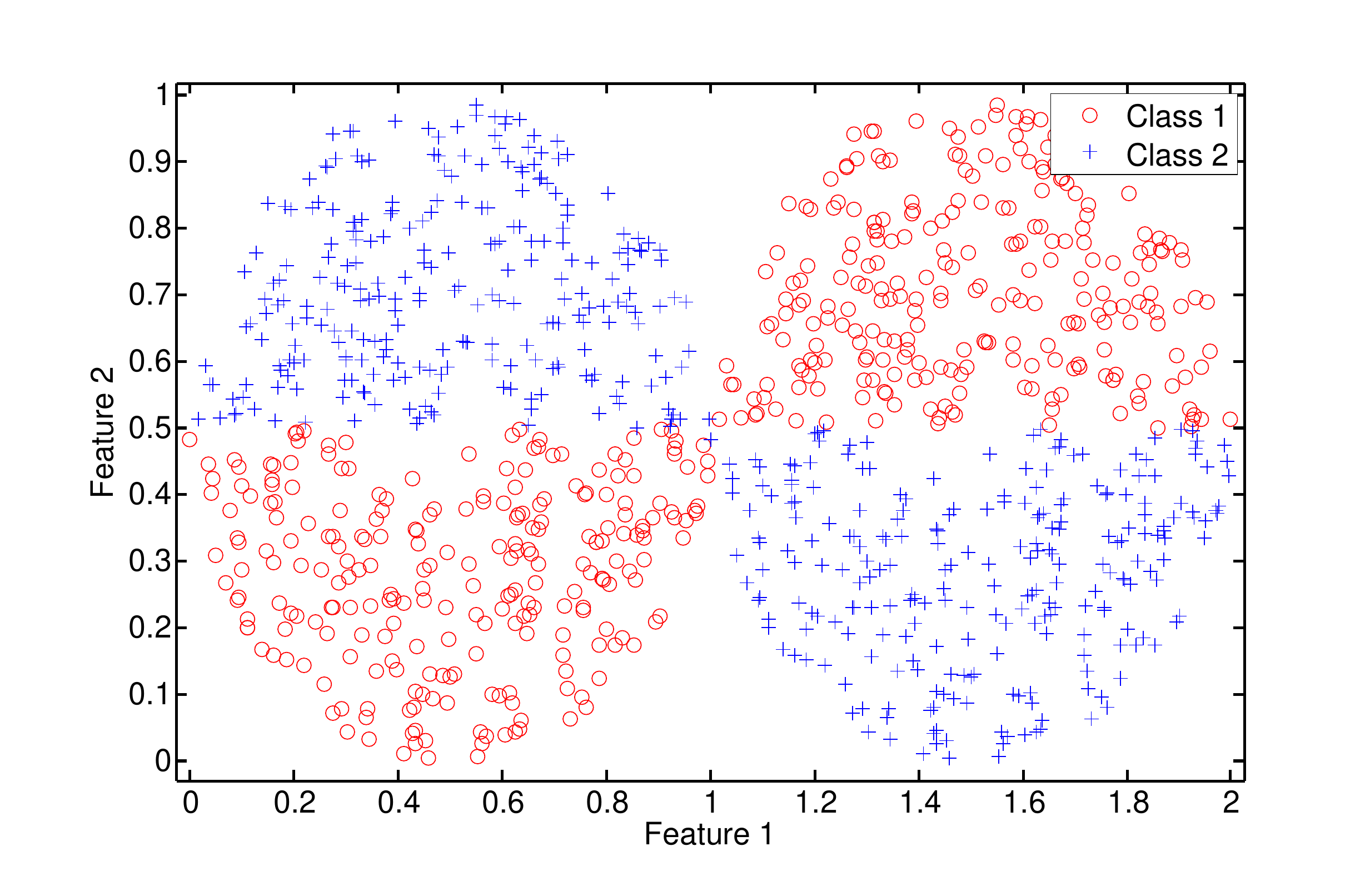}} \\
	\subfigure{\includegraphics[width=3.2in,clip=]{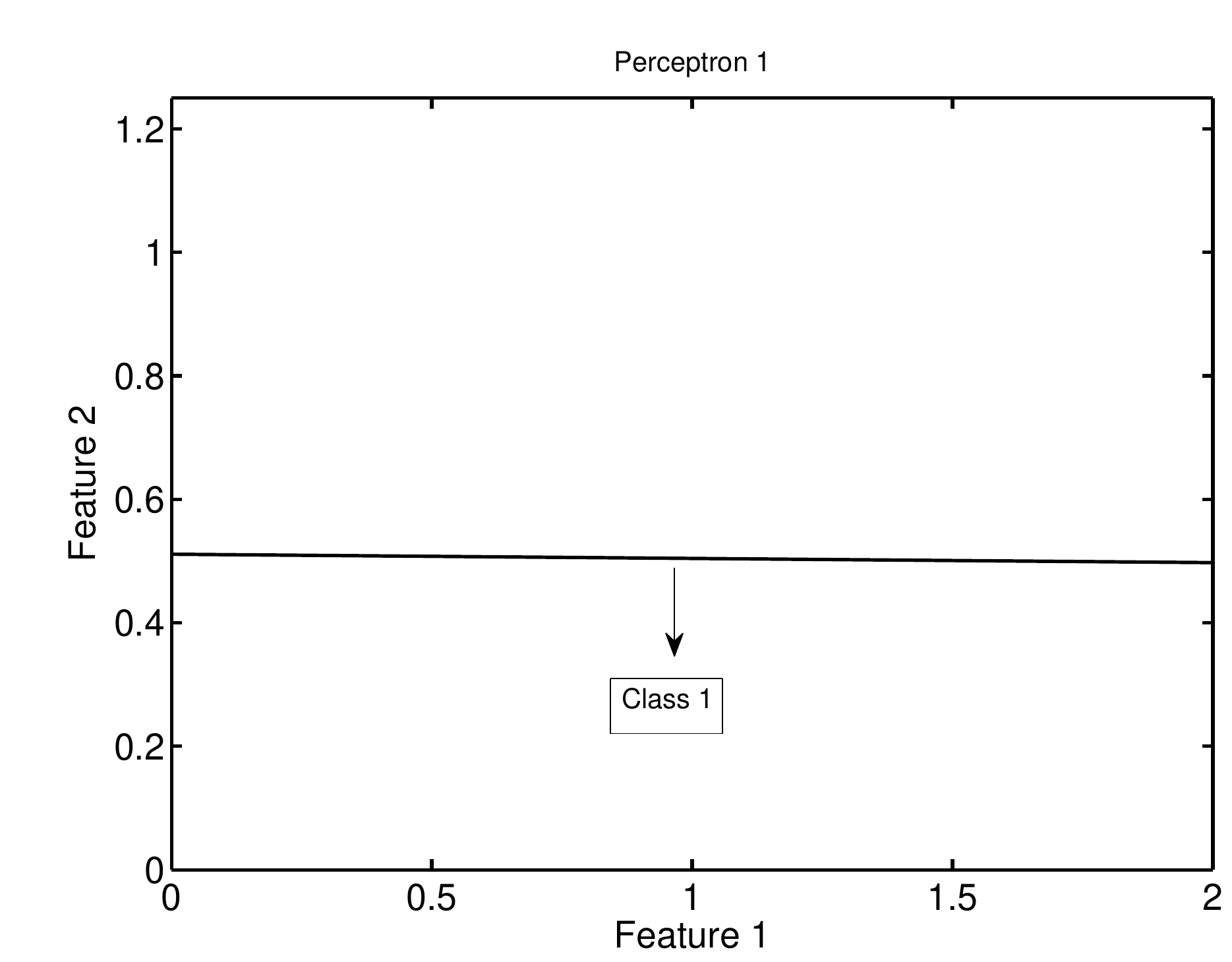}} 
	\subfigure{\includegraphics[width=3.2in,clip=]{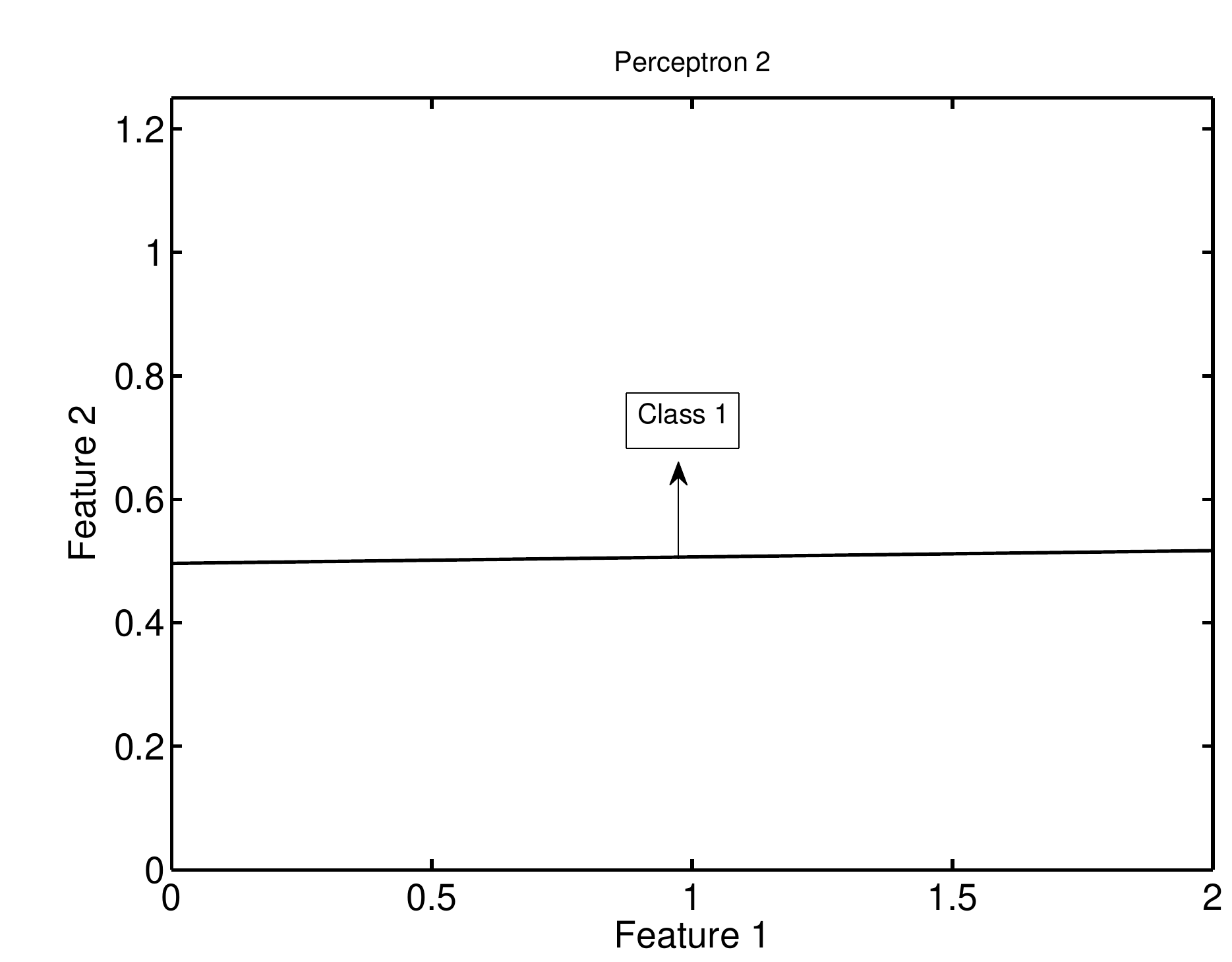}}
	\caption{(a) The two circles data generated with 1000 data points, 500 samples for each class; (b) illustrates the decision made by the Perceptron $c_{1}$; (c) shows the decision made by the Perceptron $c_{2}$.}
	\label{fig:TwoCirclesData}	  
\end{figure}

Let us consider, for instance, two circles representing the exclusive or XOR problem. The problem is generated with 1000 data points, 500 for each class (Figure~\ref{fig:TwoCirclesData} (a)). Two linear classifiers trained for this problem (two Perceptrons) $c_{1}$ and $c_{2}$, both with an individual accuracy of 50\%. The decisions of $c_{1}$ and $c_{2}$ are shown in (Figure~\ref{fig:TwoCirclesData} (b) and (c) respectively). 

Static combination rules, such as majority voting or averaging are useless in this case since the base classifiers always yield opposite decisions, i.e., for any query sample $\mathbf{x}_{j}$, if $c_{1}$ predicts that $\mathbf{x}_{j}$ belongs to class 1, $c_{2}$ will predict that $\mathbf{x}_{j}$ belongs to class 2 and vice versa. There is never a consensus between the decisions obtained by these two classifiers.  

\begin{figure}[!ht]
	\begin{center}  	 
		\includegraphics[clip=, width=0.7\textwidth]{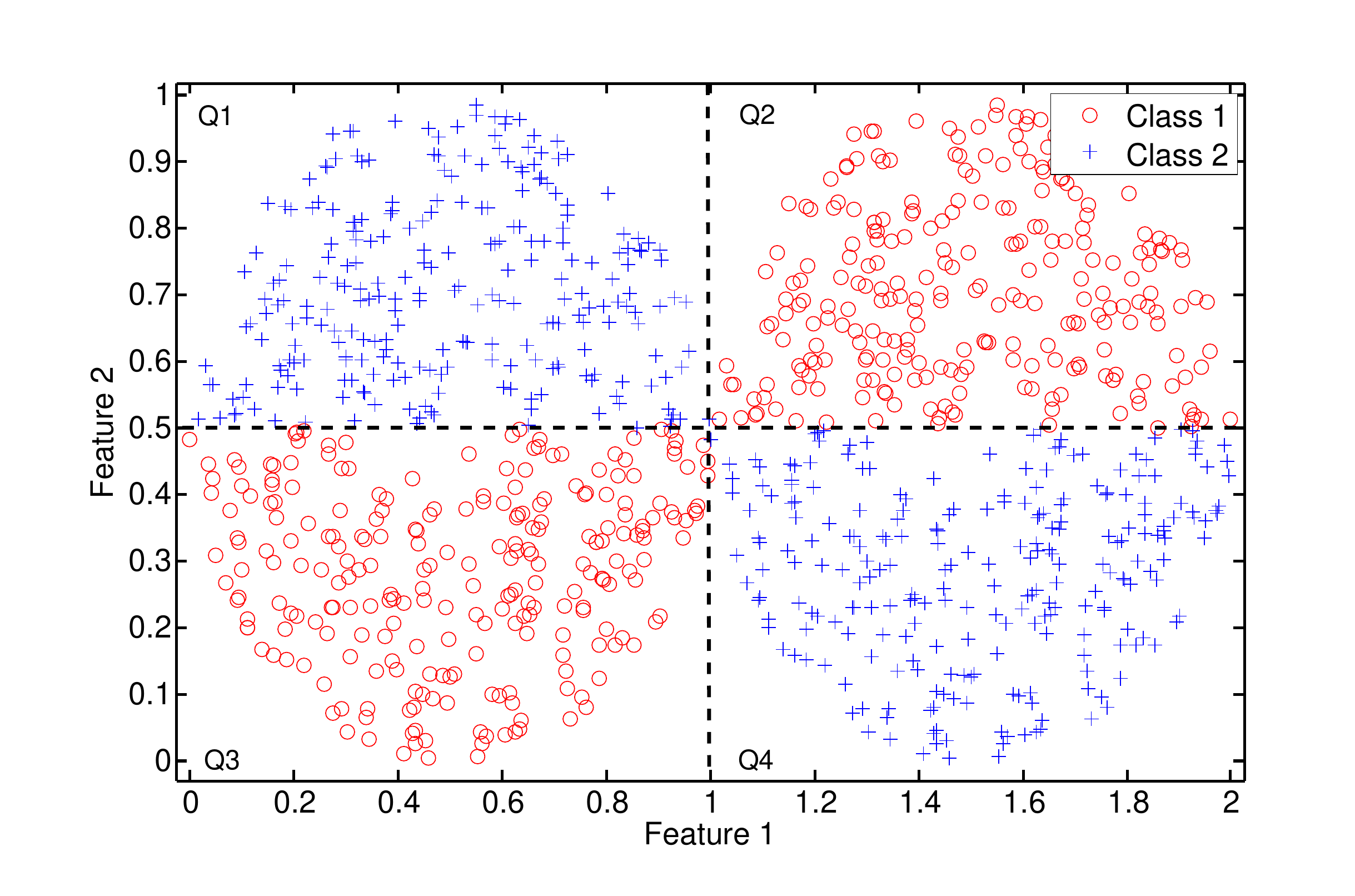}
	\end{center}
	\caption{The two circles data divided into four regions.}
	\label{fig:twoCirclesDiv}	  
\end{figure}

Considering the same data, it is possible to split the feature space into four local regions (Figure~\ref{fig:twoCirclesDiv}): Q1, Q2, Q3 and Q4. 

Using dynamic selection, it is possible to obtain a 100\% accuracy rate using only these two classifiers. Given a query instance $\mathbf{x}_{j}$, the system first checks the competence of each classifier in the pool. Only the classifier(s) with the highest competence are selected. Classifiers that are not experts in the local region will not influence the ensemble decision. 

%Then, using dynamic selection, it is possible to obtain 100\% accuracy rate using only these two classifiers.
Given a query sample $\mathbf{x}_{j}$ to be classified, using dynamic selection, the classification is performed as follows(Equation~\ref{eq:selection}):

\begin{equation}\left\{\begin{matrix}
\text{If} \;  \mathbf{x}_{j} \in \text{Q1} & \text{Select} \;  c_{1} \\ 
\text{If} \;  \mathbf{x}_{j} \in \text{Q2} & \text{Select} \;  c_{2} \\ 
\text{If} \;  \mathbf{x}_{j} \in \text{Q3} & \text{Select} \;  c_{2} \\ 
\text{If} \;  \mathbf{x}_{j} \in \text{Q4} & \text{Select} \;  c_{1} \\ 
\end{matrix}\right.
\label{eq:selection}
\end{equation}

The key issue in DES is to define a criterion to measure the level of competence of a base classifier. Most DES techniques~\cite{knora,lca,mcb,ijcnn2011,singh,Smits_2002,clussel,SoaresSCS06} use estimates of the classifiers' local accuracy in small regions of the feature space surrounding the query instance as a search criterion to perform the ensemble selection. There are other criteria, such as the degree of consensus, in the ensemble~\cite{docs}, probabilistic models applied to the classifier outputs~\cite{Woloszynski} and decision templates~\cite{paulo2,logid}. A recent survey on dynamic selection~\cite{Alceu2014} covers all the DES criteria used by different techniques. 

In~\cite{CruzPR, icpr2014}, we proposed a novel DES framework in which multiple criteria regarding the behavior of a base classifier are used to have a better estimation of its level of competence. The META-DES framework is presented in the following sections.

\section{The META-DES Framework}
\label{sec:proposed}

The META-DES framework is based on the assumption that the dynamic ensemble selection problem can be considered as a meta-problem. This meta-problem uses different criteria regarding the behavior of a base classifier $c_{i}$, in order to decide whether it is competent enough to classify a given test sample $\mathbf{x}_{j}$. The meta-problem is defined as follows~\cite{CruzPR}:

 \begin{itemize}
 
 \item The \textbf{meta-classes} of this meta-problem are either ``competent'' (1) or ``incompetent'' (0) to classify $\mathbf{x}_{j}$.
 
 \item Each set of \textbf{meta-features} $f_{i}$ corresponds to a different criterion for measuring the level of competence of a base classifier.
 
 \item The meta-features are encoded into a \textbf{meta-features vector} $v_{i,j}$.
 
 \item A \textbf{meta-classifier} $\lambda$ is trained based on the meta-features $v_{i,j}$ to predict whether or not $c_{i}$ will achieve the correct prediction for $\mathbf{x}_{j}$, i.e., if it is competent enough to classify $\mathbf{x}_{j}$
 
 \end{itemize}
 
\begin{figure*}[!ht]
  
   \begin{center}  	 
       	  \includegraphics[clip=,  width=0.900\textwidth]{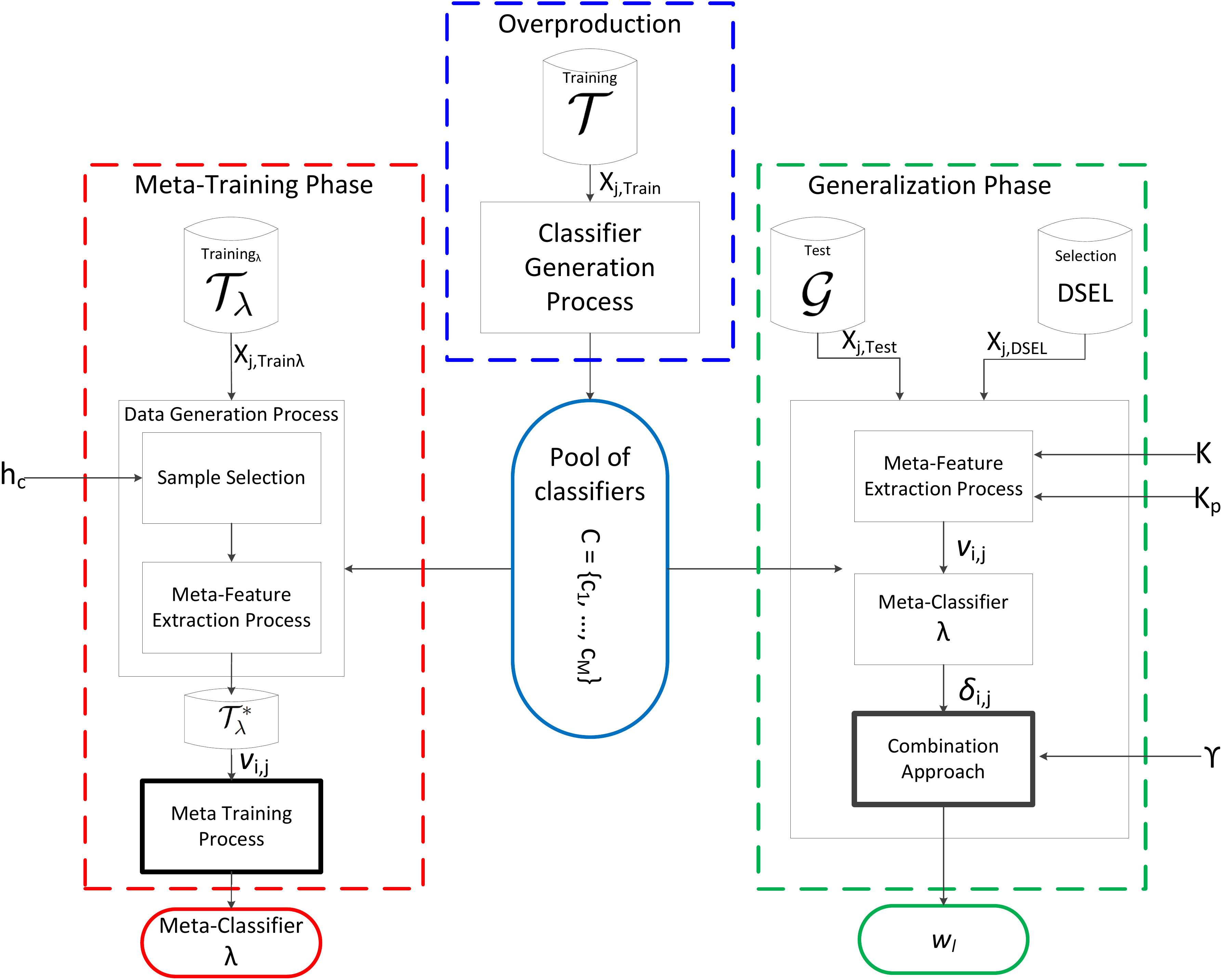}
   \end{center}
\caption{Overview of the proposed framework. It is divided into three steps 1) Overproduction, where the pool of classifiers $C = \{c_{1}, \ldots, c_{M}\}$ is generated, 2) The training of the selector $\lambda$ (meta-classifier), and 3) The generalization phase where the level of competence $\delta_{i,j}$ of each base classifier $c_{i}$ is calculated specifically for each new test sample $\mathbf{x}_{j,test}$. Then, the level of competence $\delta_{i,j}$ is used by the combination approach to predict the label $w_{l}$ of the test sample $\mathbf{x}_{j,test}$. Three combination approaches are considered: Dynamic selection (META-DES.S), Dynamic weighting (META-DES.W) and Hybrid (META-DES.H). $h_{C}$, $K$, $K_{p}$ and $\Upsilon$ are the hyper-parameters required by the proposed system. [Adapted from~\cite{CruzPR}].}
\label{fig:overview}
\end{figure*}

A general overview of the META-DES framework is depicted in Figure~\ref{fig:overview}. It is divided into three phases: Overproduction, Meta-training and Generalization.

\subsection{Overproduction} 

In this step, the pool of classifiers $C = \{c_{1}, \ldots, c_{M}\}$, where $M$ is the pool size, is generated using the training dataset $\mathcal{T}$. The Bagging technique~\cite{bagging} is used in this work in order to build a diverse pool of classifiers. 

\subsection{Meta-Training}

In this phase, the meta-features are computed and used to train the meta-classifier $\lambda$. As shown in Figure~\ref{fig:overview}, the meta-training stage consists of three steps: sample selection, meta-features extraction process and meta-training. A different dataset $\mathcal{T}_{\lambda}$ is used in this phase to prevent overfitting.

\subsubsection{Sample selection}
 
We decided to focus the training of $\lambda$ on cases in which the extent of consensus of the pool is low. This decision was based on the observations made in~\cite{docs,paulo2} the main issues in dynamic ensemble selection occur when classifying testing instances where the degree of consensus among the pool of classifiers is low, i.e., when the number of votes from the winning class is close to or even equal to the number of votes from the second class. We employ a sample selection mechanism based on a threshold $h_{C}$, called the consensus threshold. For each $\mathbf{x}_{j,train_{\lambda}} \in \mathcal{T}_{\lambda}$, the degree of consensus of the pool, denoted by $H \left ( \mathbf{x}_{j,train_{\lambda}}, C \right )$, is computed. If $H \left ( \mathbf{x}_{j,train_{\lambda}}, C \right )$ falls below the threshold $h_{C}$, $\mathbf{x}_{j,train_{\lambda}}$ is passed down to the meta-features extraction process. 

\subsubsection{Meta-feature extraction}
\label{sec:metafeatures}

The first step in extracting the meta-features involves computing the region of competence of $\mathbf{x}_{j,train_{\lambda}}$, denoted by $\theta_{j} = \left \{ \mathbf{x}_{1}, \ldots, \mathbf{x}_{K} \right \}$. The region of competence is defined in the $\mathcal{T_{\lambda}}$ set using the K-Nearest Neighbor algorithm. Then, $\mathbf{x}_{j,train_{\lambda}}$ is transformed into an output profile, $\tilde{\mathbf{x}}_{j,train_{\lambda}}$. The output profile of the instance $\mathbf{x}_{j,train_{\lambda}}$ is denoted by $\tilde{\mathbf{x}}_{j,train_{\lambda}} = \left\lbrace \tilde{\mathbf{x}}_{j,train_{\lambda},1}, \tilde{\mathbf{x}}_{j,train_{\lambda},2}, \ldots, \tilde{\mathbf{x}}_{j,train_{\lambda},M} \right\rbrace $, where each $\tilde{\mathbf{x}}_{j,train_{\lambda},i}$ is the decision yielded by the base classifier $c_{i}$ for the sample $\mathbf{x}_{j,train_{\lambda}}$~\cite{paulo2}.

The similarity between $\tilde{\mathbf{x}}_{j,train_{\lambda}}$ and the output profiles of the instances in $\mathcal{T}_{\lambda}$ is obtained through the Euclidean distance. The most similar output profiles are selected to form the set $\phi_{j} = \left \{ \tilde{\mathbf{x}}_{1}, \ldots, \tilde{\mathbf{x}}_{K_{p}} \right \}$, where each output profile $\tilde{\mathbf{x}}_{k}$ is associated with a label $w_{l,k}$. Next, for each base classifier $c_{i} \in C$, five sets of meta-features are calculated:

\begin{itemize}

\item  \emph{\boldsymbol{$f_{1}$} \textbf{- Neighbors' hard classification:}} First, a vector with $K$ elements is created. For each sample $\mathbf{x}_{k}$, belonging to the region of competence $\theta_{j}$, if $c_{i}$ correctly classifies $\mathbf{x}_{k}$, the $k$-th position of the vector is set to 1, otherwise it is 0. Thus, $K$ meta-features are computed. 

\item \emph{\boldsymbol{$f_{2}$} \textbf{- Posterior Probability:}} First, a vector with $K$ elements is created. Then, for each sample $\mathbf{x}_{k}$, belonging to the region of competence $\theta_{j}$, the posterior probability of $c_{i}$, $P(w_{l}\mid \mathbf{x}_{k})$ is computed and inserted into the $k$-th position of the vector. Consequently, $K$ meta-features are computed. 

\item \emph{\boldsymbol{$f_{3}$} \textbf{- Overall Local Accuracy:}} The accuracy of $c_{i}$ over the whole region of competence $\theta_{j}$ is computed and encoded as $f_{3}$. 

\item \emph{\boldsymbol{$f_{4}$} \textbf{- Outputs' profile classification:}} First, a vector with $K_{p}$ elements is generated. Then, for each member $\tilde{\mathbf{x}}_{k}$ belonging to the set of output profiles $\phi_{j}$, if the label produced by $c_{i}$ for $\mathbf{x}_{k}$ is equal to the label $w_{l,k}$ of $\tilde{\mathbf{x}}_{k}$, the $k$-th position of the vector is set to 1, otherwise it is set to 0. A total of $K_{p}$ meta-features are extracted using output profiles.  %correctly classifies $\mathbf{x}_{k}$

\item \emph{\boldsymbol{$f_{5}$} \textbf{- Classifier's confidence:}} The perpendicular distance between the reference sample $\mathbf{x}_{j}$ and the decision boundary of the base classifier $c_{i}$ is calculated and encoded as $f_{5}$. 

\end{itemize}

\begin{figure}[htbp]
	\begin{center}
		\includegraphics[clip=, width=0.80\textwidth]{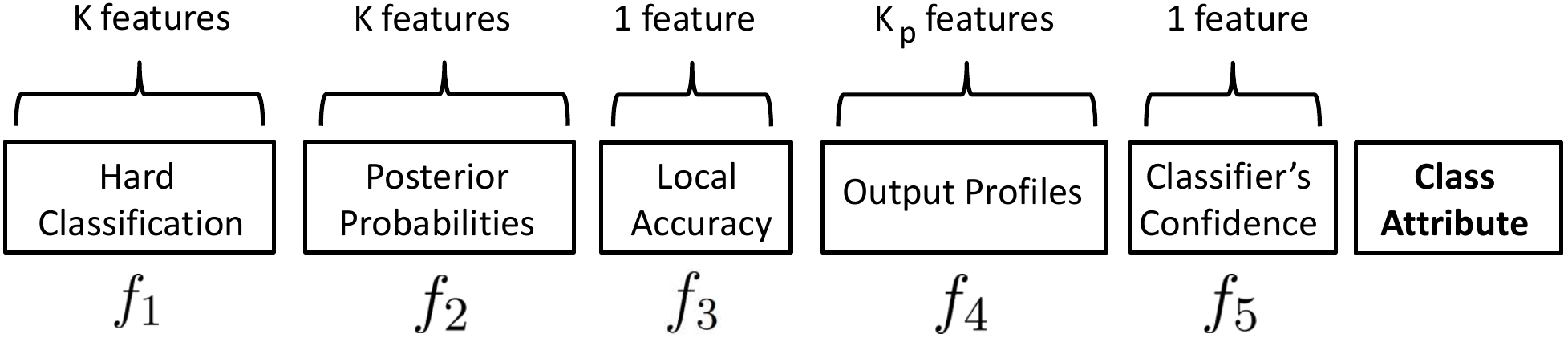}
	\end{center}
	\caption{Feature Vector containing the meta-information about the behavior of a base classifier. A total of 5 different meta-features are considered. The size of the feature vector is $(2 \times K) + K_{p} + 2$. The class attribute indicates whether or not $c_{i}$ correctly classified the input sample.}
	\label{fig:featVector}
\end{figure}

A vector $v_{i,j} = \left\lbrace f_{1} \cup f_{2} \cup f_{3} \cup f_{4} \cup f_{5} \right\rbrace$ (Figure~\ref{fig:featVector}) is obtained at the end of the process. If $c_{i}$ correctly classifies $\mathbf{x}_{j}$, the class attribute of $v_{i,j}$, $\alpha_{i,j} = 1$ (i.e., $v_{i,j}$ belongs to the meta-class ``competent''), otherwise $\alpha_{i,j} = 0$. $v_{i,j}$ is stored in the meta-features dataset $\mathcal{T}_{\lambda}^{*}$ that is used to train the meta-classifier $\lambda$. Figure~\ref{fig:featVector} illustrates the format of the meta-features vector $v_{i,j}$.

\subsubsection{Training}

The last step of the meta-training phase is the training of the meta-classifier $\lambda$. In this work, we considered a Naive Bayes for the meta-classifier $\lambda$, since this classifier model presented the best classification results for the META-DES framework when compared against different classifier models, such as a Multi-Layer Perceptron Neural Network and a Random Forest classifier~\cite{ijcnn2015}. %In addition, as the outputs of the Naive Bayes classifier represents the posterior probabilities that the base classifier belongs to the ``competent'' meta-class. % so that they can be used to weight the decision of the selected base classifier according to the probability that it belongs to the ``competent'' meta-class.

\subsection{Generalization}
\label{sec:generalization}

%The generalization procedure is formalized by Algorithm~\ref{alg:generalization}.

Given the query sample $\mathbf{x}_{j,test}$, the region of competence $\theta_{j}$ is computed using the samples from the dynamic selection dataset $DSEL$. Following that, the output profiles $\tilde{\mathbf{x}}_{j,test}$ of the test sample, $\mathbf{x}_{j,test}$, are calculated. The set with $K_{p}$ similar output profiles $\phi_{j}$, of the query sample $\mathbf{x}_{j,test}$, is obtained through the Euclidean distance applied over the output profiles of the dynamic selection dataset, $\tilde{DSEL}$.

For each base classifier, $c_{i}$, belonging to the pool of classifiers, $C$, the five sets of meta-features are computed, returning the meta-features vector $v_{i,j}$. Then, $v_{i,j}$ is used as input to the meta-classifier $\lambda$. The support obtained by $\lambda$ for the ``competent'' meta-class is computed as the level of competence, $\delta_{i,j}$, of the base classifier $c_{i}$ for the classification of the test sample $\mathbf{x}_{j,test}$. As in~\cite{ijcnn2015}, we consider a hybrid combination approach called META-DES.H. First, the base classifiers that achieve a level of competence $\delta_{i,j} > \Upsilon = 0.5$ are selected to compose the ensemble $C'$. Next, the decision of each selected base classifier is weighted by its level of competence.A weighted majority voting approach is used to predict the label $w_{l}$ of the sample $\mathbf{x}_{j,test}$.  Thus, the decisions obtained by the base classifiers that attained a higher level of competence $\delta_{i,j}$ have a greater influence in the final decision.

\section{Why does the META-DES work: A Step-by-step example}
\label{sec:whyMETA}

In this section, we present a step-by-step example of the training and test phases of the META-DES framework in order to understand the mechanisms behind the META-DES, and why it achieves good generalization performance using only linear classifiers. For this example, we use the P2 problem.

\subsection{The P2 Problem}

\begin{figure}[H]
   \begin{center}  	 
       	  \includegraphics[clip=, width=0.7\textwidth]{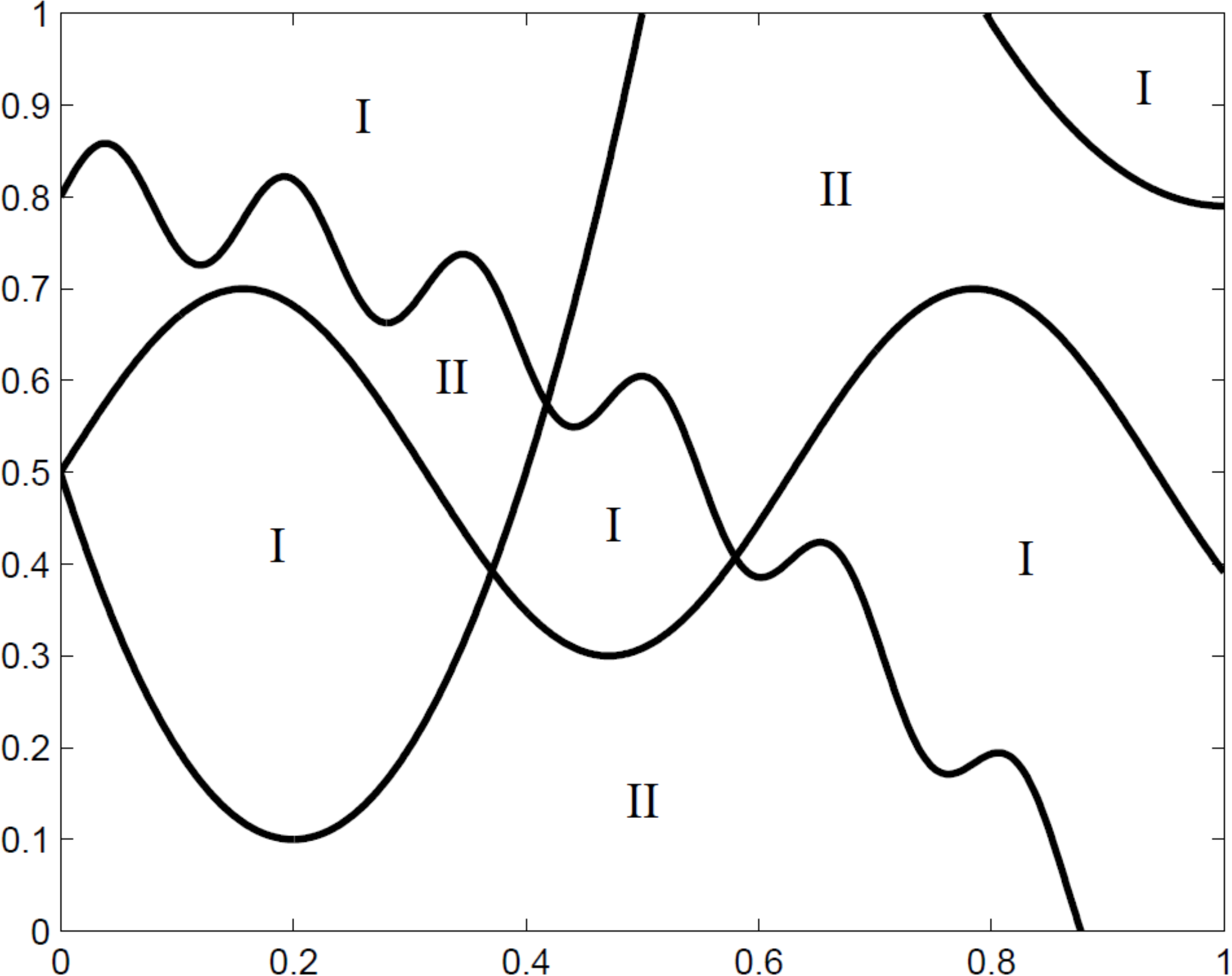}
   \end{center}
   \caption{The P2 Problem. The symbols I and II represents the area of the classes 1 and 2 respectively}
\label{fig:P2Problem}	  
\end{figure}

The P2 is a two-class problem, presented by Valentini~\cite{Valentini05}, in which each class is defined in multiple decision regions delimited by polynomial and trigonometric functions (Equation~\ref{eq:problem2}). As in~\cite{henniges}, $E4$ was modified such that the area of each class was approximately equal. The P2 problem is illustrated in Figure~\ref{fig:P2Problem}. One can clearly see that it is impossible to solve this problem using linear classifiers. The performance of the best possible linear classifier is around 50\%.

%The decision of the P2 problem is shown in dashed lines.

\begin{eqnarray} 
\label{eq:problem2}
E1(x) = sin(x) + 5 \\
E2(x) = (x - 2)^{2} + 1 \\
E3(x) = -0.1 \cdot x^{2} + 0.6sin(4x) + 8 \\
E4(x) = \frac{(x - 10)^{2}}{2} + 7.902 
\end{eqnarray} 

For this illustrative example, the P2 problem is generated as follows: 500 samples for training ($\mathcal{T}$), 500 instances for the meta-training dataset ($\mathcal{T}_{\lambda}$), 500 instances for the dynamic selection dataset $DSEL$, and 2000 samples for the test dataset, $\mathcal{G}$. For the sake of simplicity, we use a pool composed of 5 Perceptrons. We demonstrate that using only 5 Perceptrons it is possible to approximate the complex decision boundary of the P2 problem using the META-DES framework.

\subsection{Overproduction}

\begin{figure}[H]
   \begin{center}  	 
       	  \includegraphics[clip=, width=1.0\textwidth]{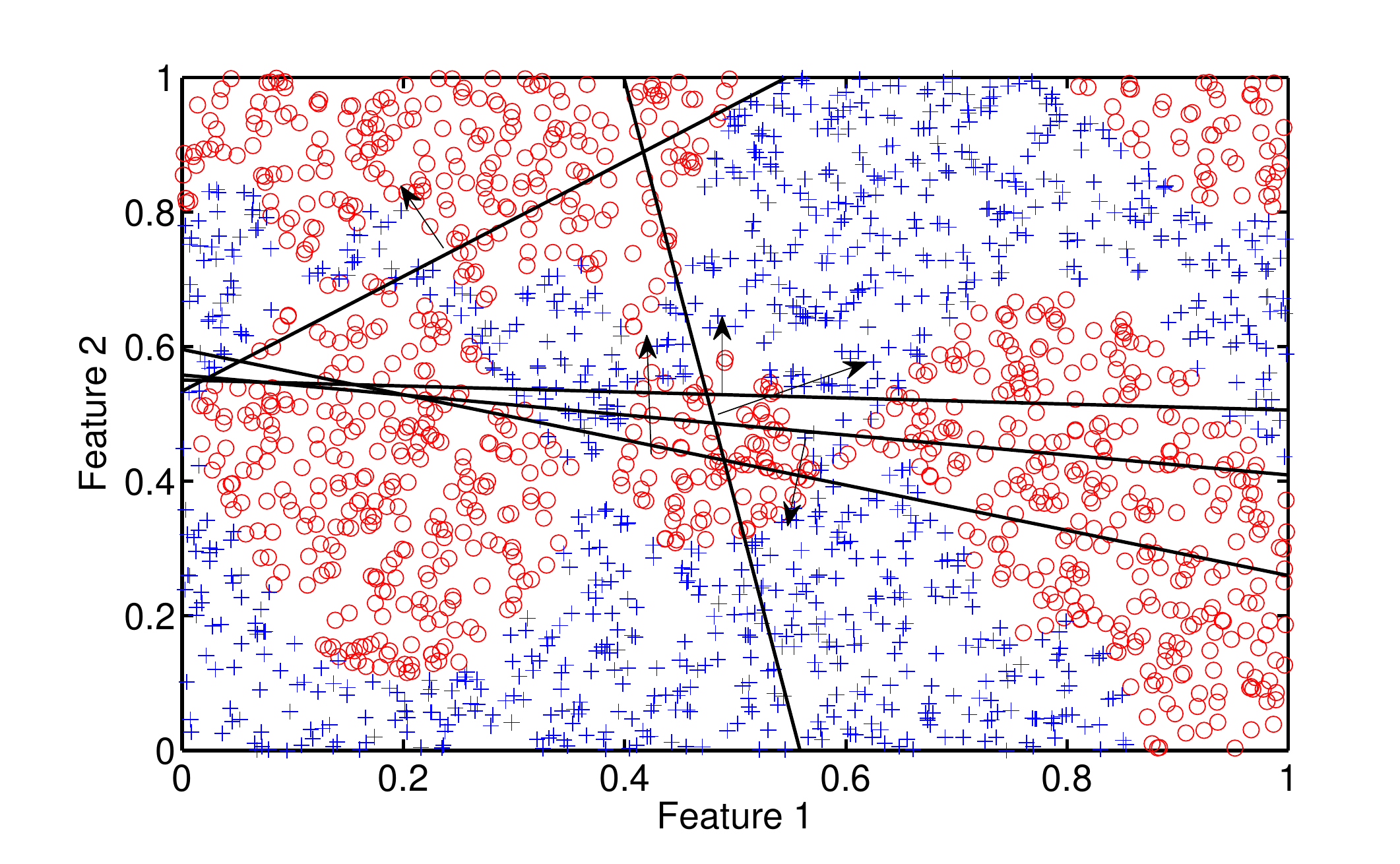}
   \end{center}
\caption{Five Perceptrons trained for the P2 Problem. The bagging technique was used to generate the pool. The arrows in each Perceptron points to the region where the classifier output is class 1 (red circle).}
\label{fig:FivePerceptrons}	  
\end{figure}

\begin{figure}[H]
	\centering
	\subfigure[Perceptron $c_{1}$]{\includegraphics[width=3.2in,clip=]{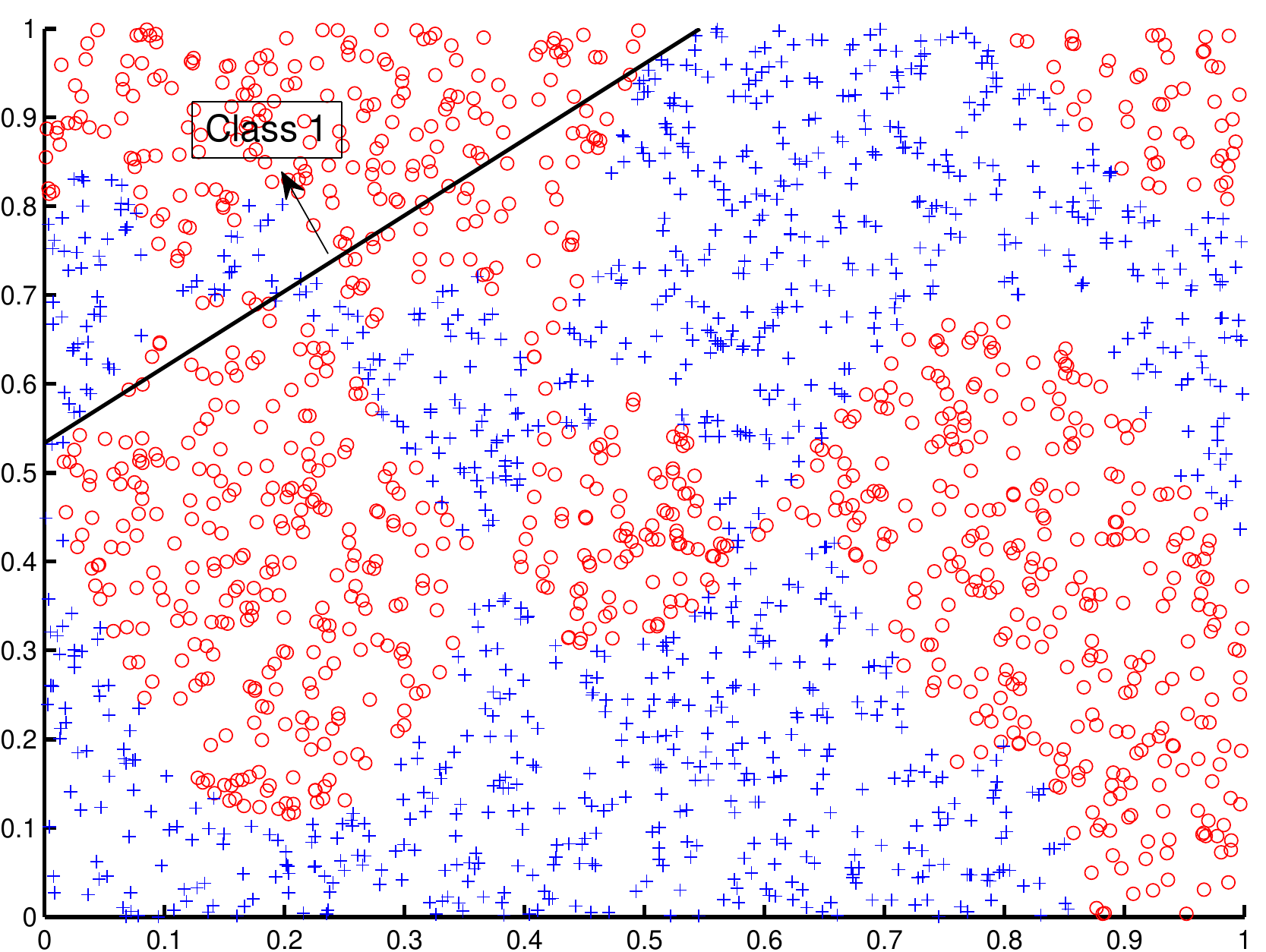}} 
	\subfigure[Perceptron $c_{2}$]{\includegraphics[width=3.2in,clip=]{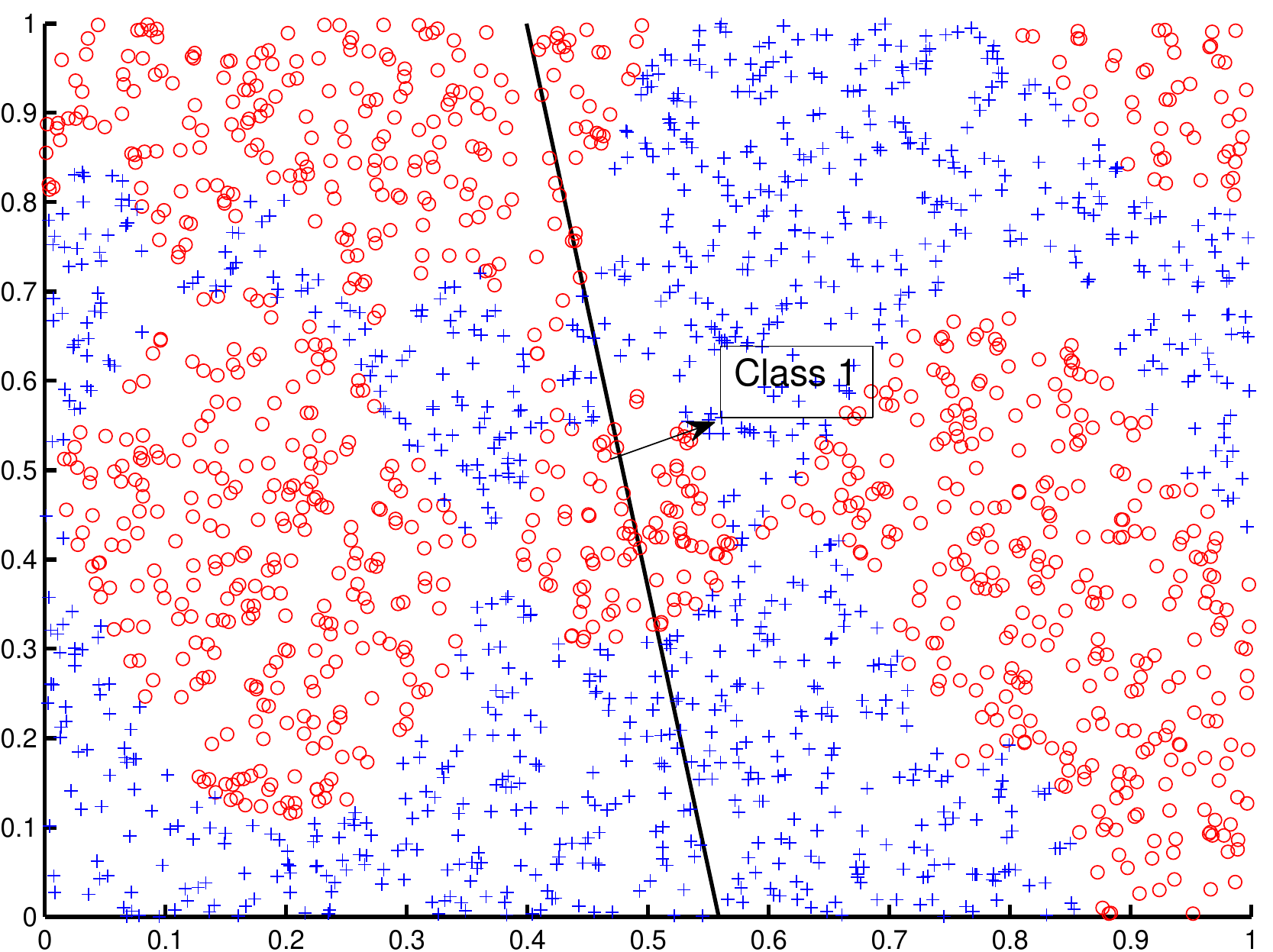}} 
	\subfigure[Perceptron $c_{3}$]{\includegraphics[width=3.2in,clip=]{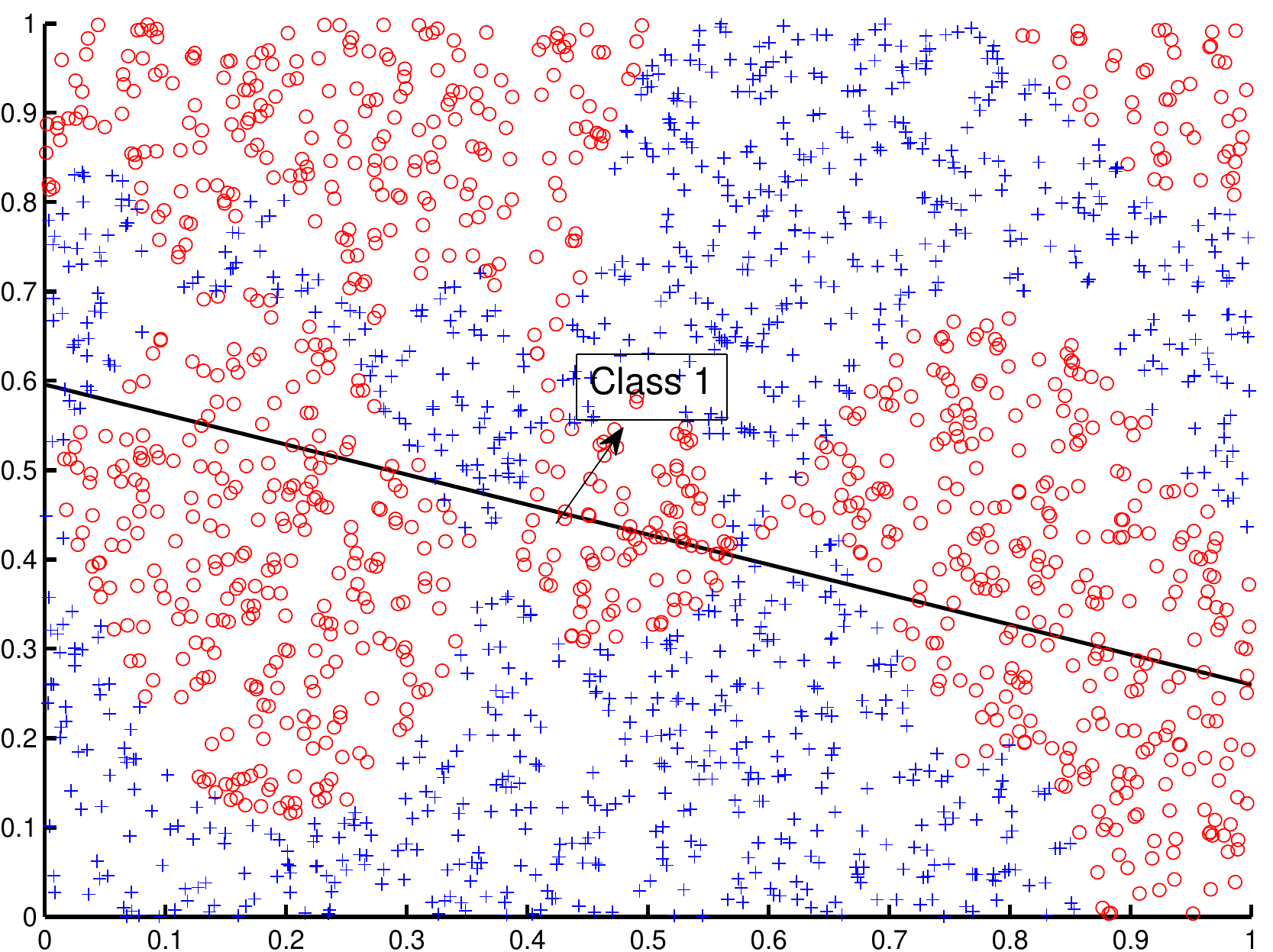}} 
	\subfigure[Perceptron $c_{4}$]{\includegraphics[width=3.2in,clip=]{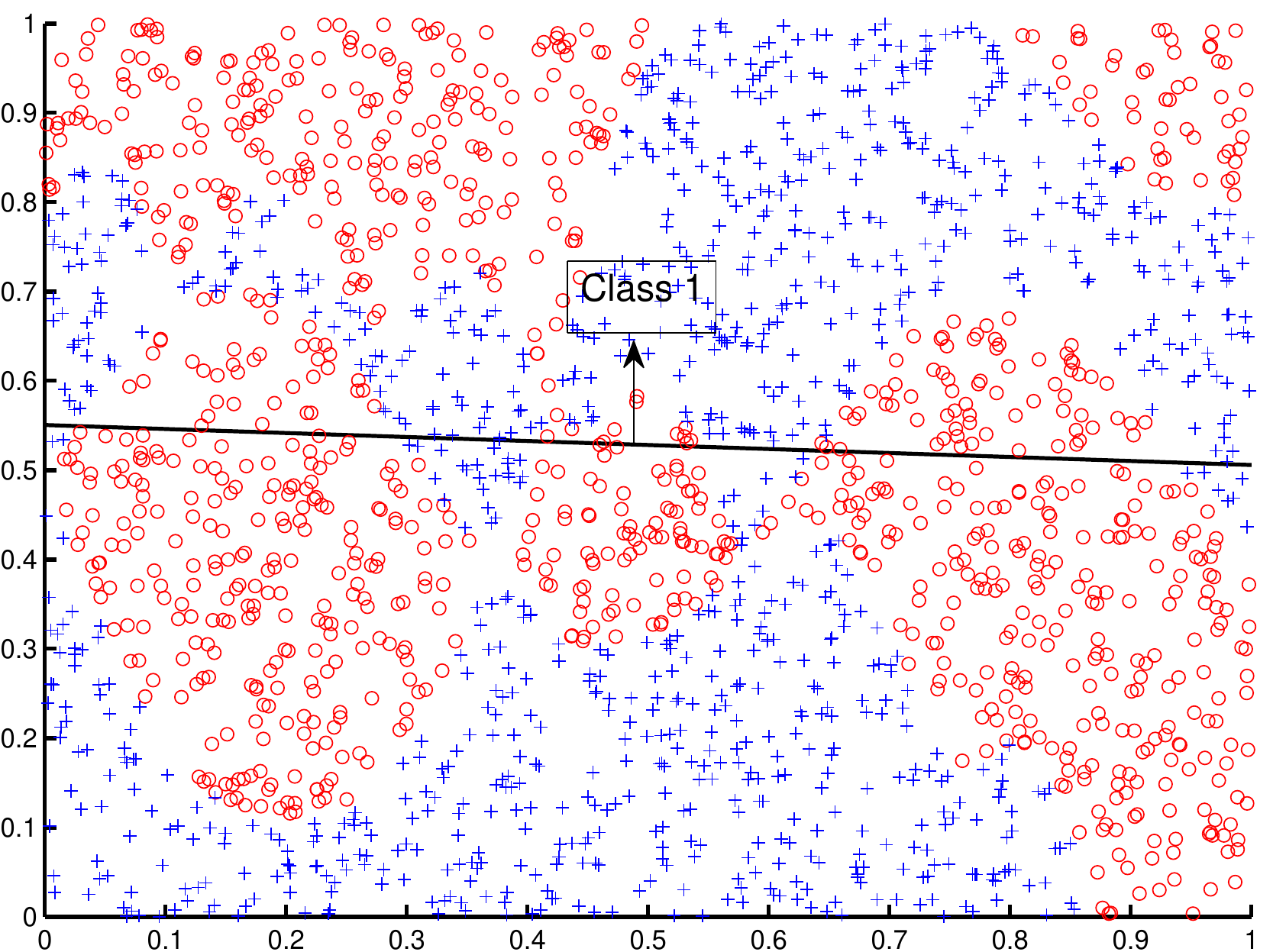}} 
	\subfigure[Perceptron $c_{5}$]{\includegraphics[width=3.2in,clip=]{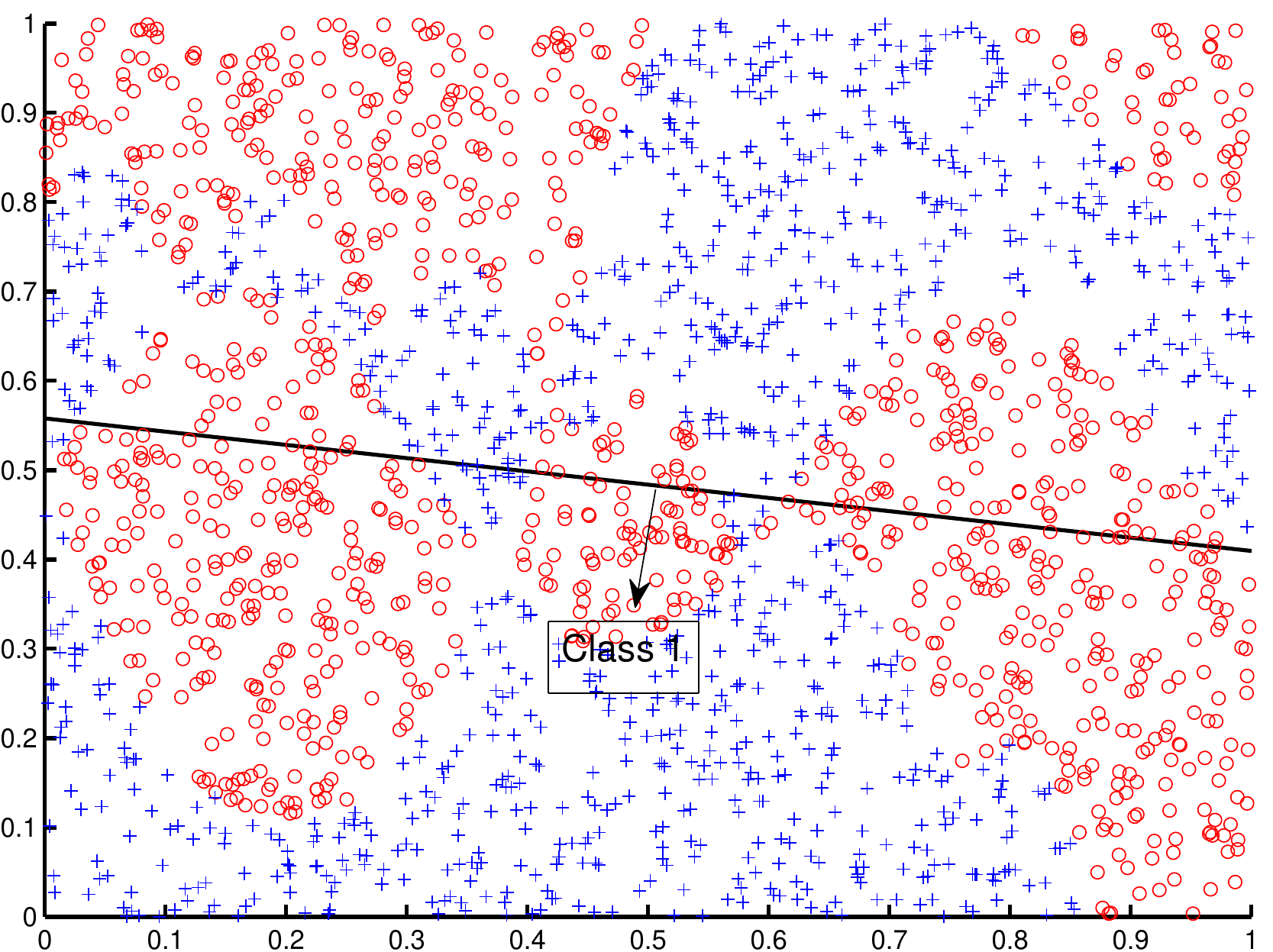}} 
	\caption{Decision of each of the five Perceptrons shown separately. The arrow in each Perceptron points to the region where the classifier output is class 1 (red circle).}
	\label{fig:PerceptronsIndividual}	  
\end{figure}

Figure~\ref{fig:FivePerceptrons} shows five Perceptrons generated using the bagging technique for the P2 problem. The arrow in each Perceptron points to the region where the classifier output is class 1 (red circle). Figure~\ref{fig:PerceptronsIndividual} presents the decision of each Perceptron individually.

The best classifier of the pool (Single Best) achieves an accuracy rate of 53.5\% ($c_{1}$). The performance of all other base classifiers is around the 50\% mark. The Oracle result of this pool obtained a recognition rate of 99.5\%. The Oracle is an abstract model defined in~\cite{Kuncheva:2002}, which always selects the classifier that predicted the correct label, for the given query sample, if such a classifier exists. In other words, it represents the ideal classifier selection scheme. There is at least one base classifier that predicts the correct label for 99.5\% of the test instances. The key issue is finding the right criteria to estimate the competence of the base classifiers in order to select only the competent ones. 

\subsection{Meta-training: Sample Selection}

\begin{figure}[H]
	\centering
	\subfigure[The original $\mathcal{T}_{\lambda}$ set ]{\includegraphics[width=3.2in,clip=]{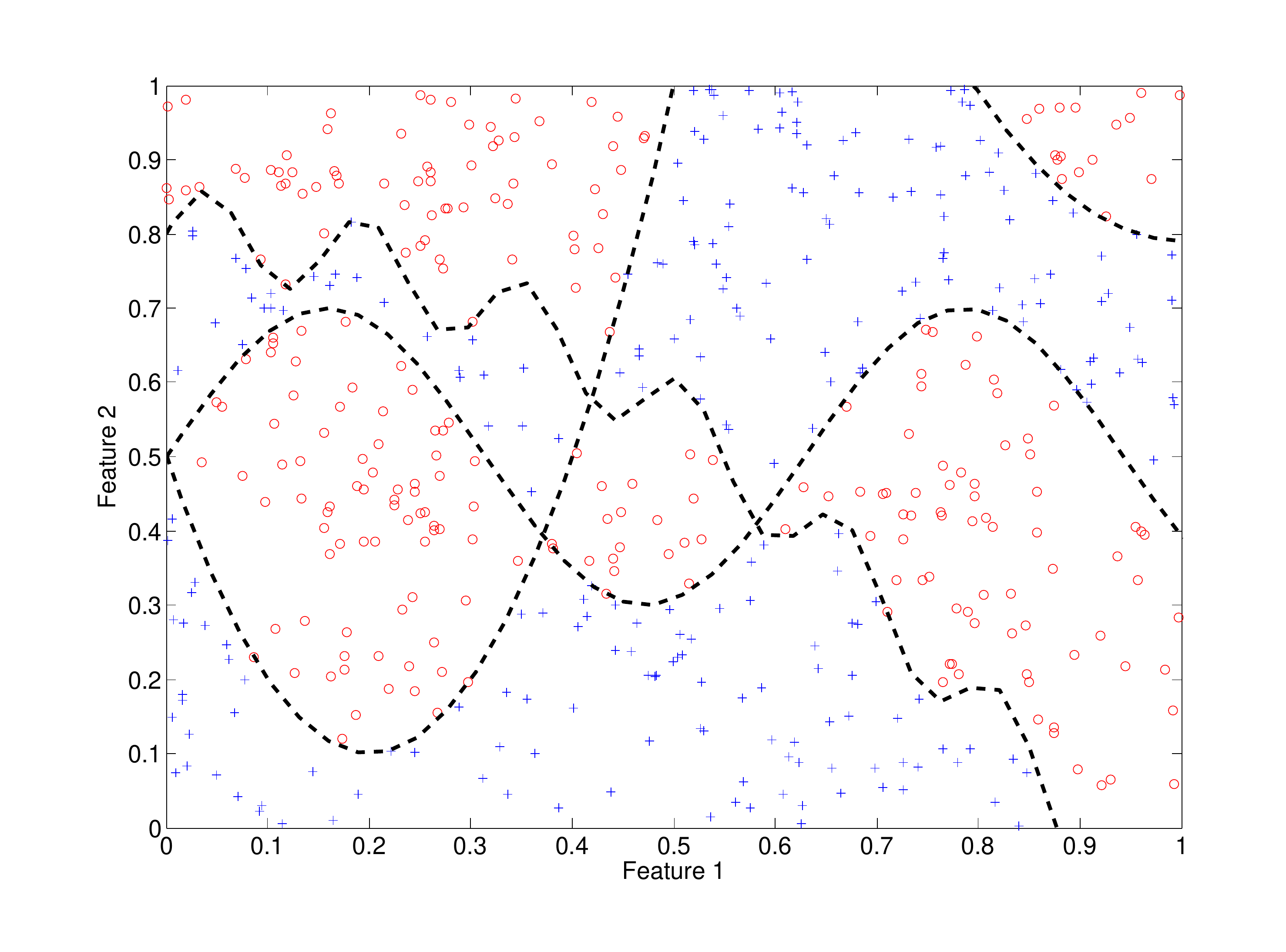}} 
	\subfigure[$\mathcal{T}_{\lambda}$ after the sample selection mechanism]{\includegraphics[width=3.2in,clip=]{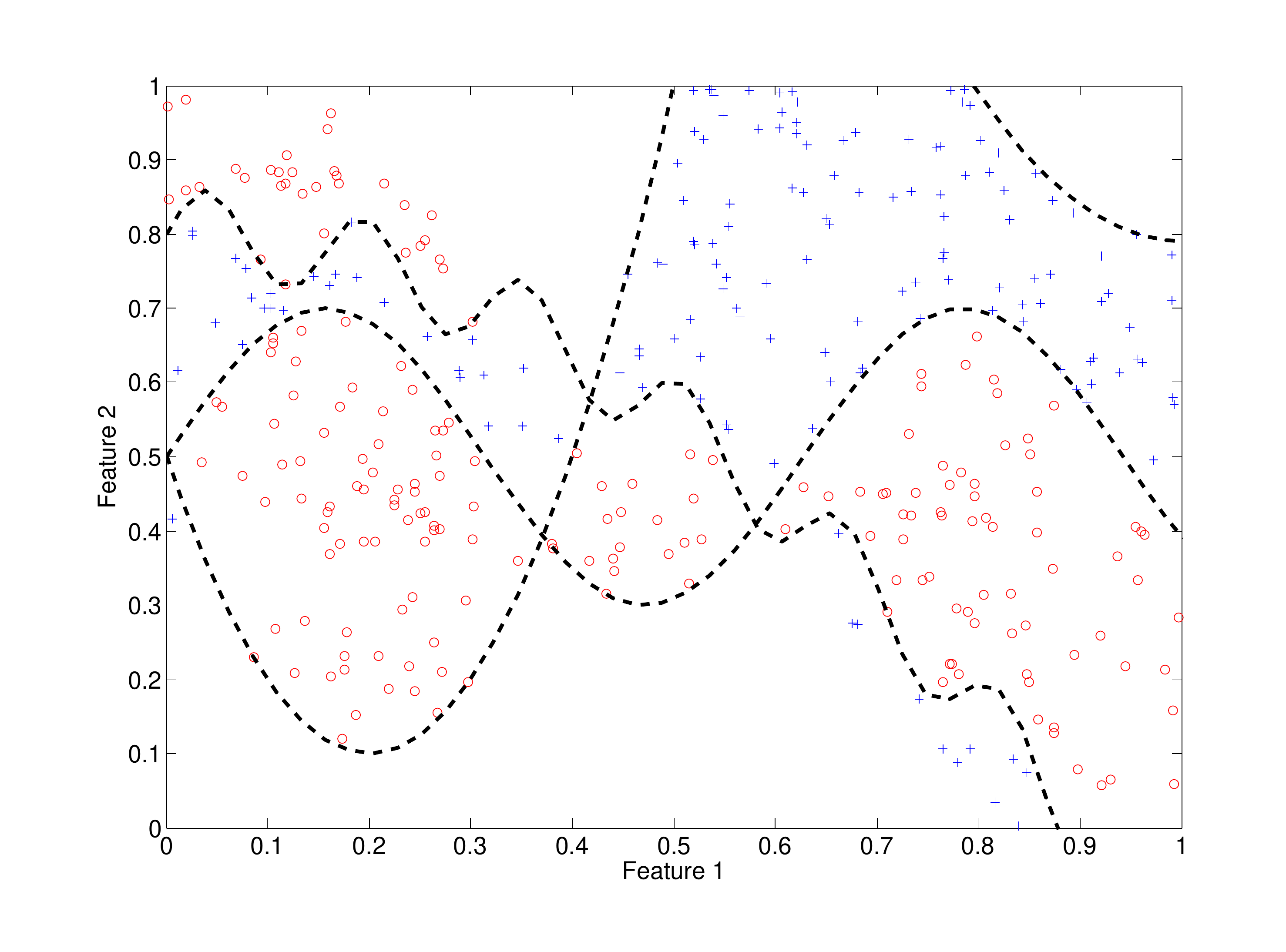}} 
	\caption{(a) The original $\mathcal{T}_{\lambda}$ dataset generated with 500 samples (250 for each class). (b) $\mathcal{T}_{\lambda}$ after the sample selection mechanism was applied. 349 samples were selected}
	\label{fig:SampleSelectionMechanism}	  
\end{figure}

After generating the pool of classifiers $C$, the next step is the sample selection mechanism for training the meta-classifier. Figure~\ref{fig:SampleSelectionMechanism} illustrates the effect of the sample selection mechanism. As in~\cite{CruzPR,ijcnn2015} the consensus threshold $h_{c}$ is set at 70\%. (Figure~\ref{fig:SampleSelectionMechanism} (a)) shows the original $\mathcal{T}_{\lambda}$ set before the sample selection. Figure~\ref{fig:SampleSelectionMechanism} (b) shows the samples that were selected for training the meta-classifier. 

The sample selection mechanism focuses on samples whose correct labels are harder to predict, i.e., when there is no consensus between the classifiers in the pool. Samples close to the decision boundary are the ones more likely to be selected for the training of the meta-classifier. This principle is similar to the support vectors in the SVM technique, in which samples close to the decision boundary are used as support vectors to achieve a better separation between classes. In the META-DES framework, the samples close to the decision boundary are used to train the meta-classifier, while samples that are closer to the class mean are not used for training since the majority of base classifiers can correctly classify those samples. Only the samples shown in Figure~\ref{fig:SampleSelectionMechanism} (b) are passed down to the meta-features extraction process and are used for the training of the meta-classifier $\lambda$.

\subsection{Classification}
\label{sec:classification}

\begin{figure}[H]
	\begin{center}  	 
	\includegraphics[clip=,  width=0.7\textwidth]{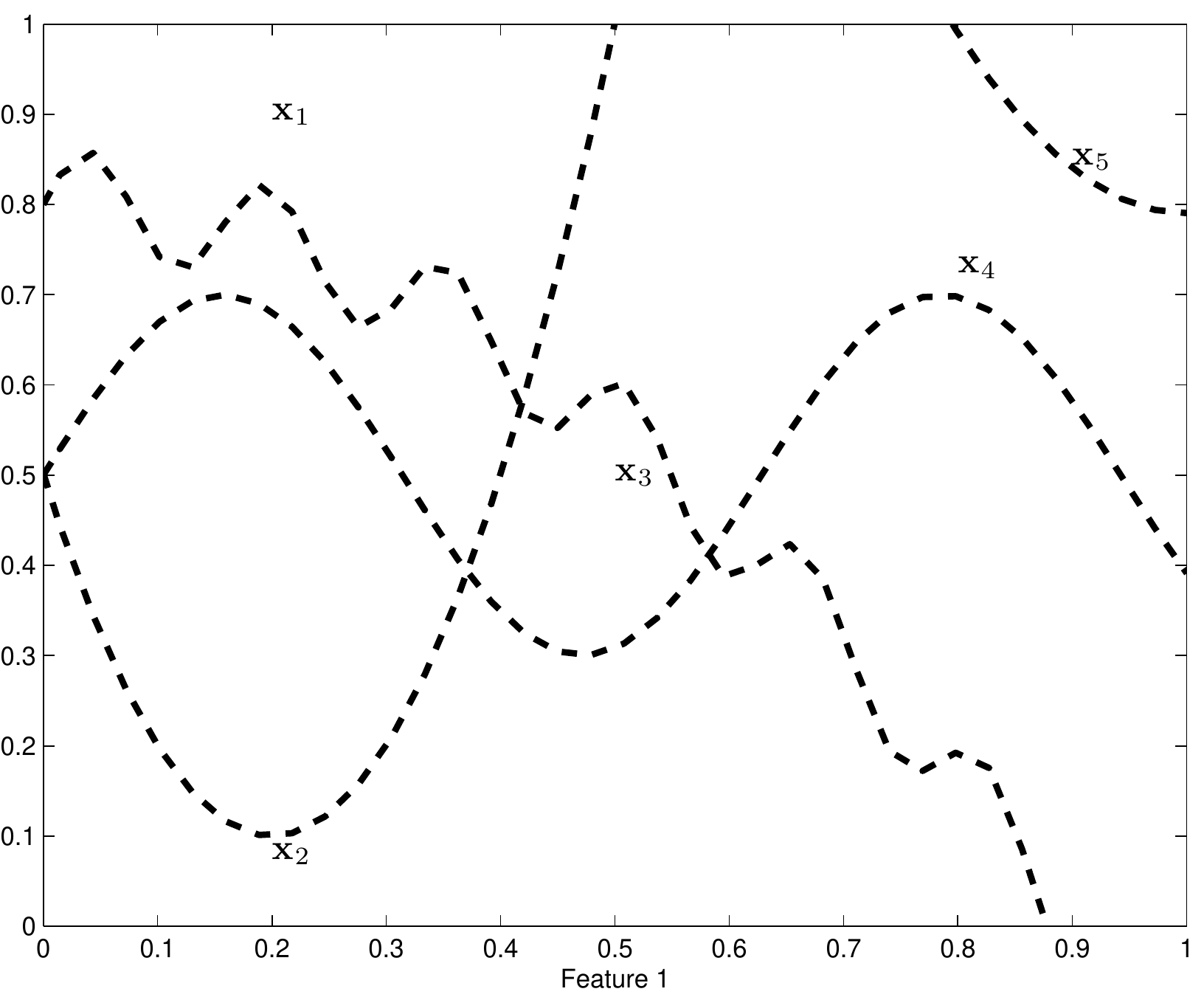}
	\end{center}
	\caption{Five samples to be classified. $\mathbf{x}_{1}$  $\mathbf{x}_{3}$ and $\mathbf{x}_{5}$ belonging to class 1,  $\mathbf{x}_{2}$ and $\mathbf{x}_{4}$ belonging to class 2.}
	\label{fig:P2Example}	  
\end{figure}

To illustrate the classification steps of the system we consider five testing samples in different parts of the feature space. The coordinates of the each query instance are: $\mathbf{x}_{1} = [0.2, \: 0.9] $, $\mathbf{x}_{2} = [0.2, \: 0.1] $, $\mathbf{x}_{3} = [0.5, \: 0.5] $, $\mathbf{x}_{4} = [0.8, \: 0.7]$ and $\mathbf{x}_{5} = [0.9, \: 0.85]$. Figure~\ref{fig:P2Example} illustrates the positions of the five testing samples. $\mathbf{x}_{1}$  $\mathbf{x}_{3}$ and $\mathbf{x}_{5}$ belongs to class 1,  $\mathbf{x}_{2}$ and $\mathbf{x}_{4}$ belongs to class 2.

\begin{figure}[H]
	\begin{center}  	 
		\includegraphics[ clip=,  width=0.825\textwidth]{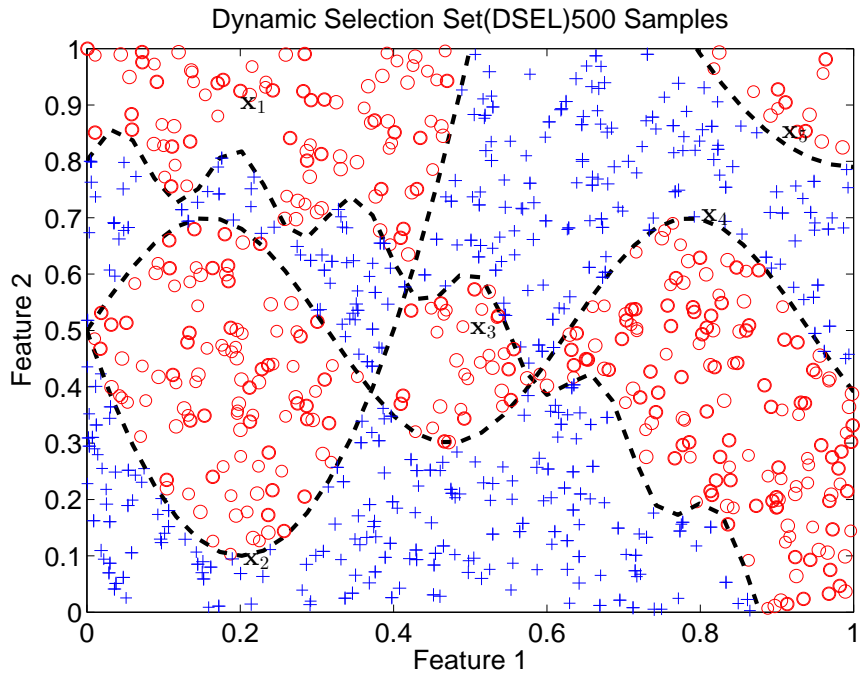}
	\end{center}
	\caption{The dynamic selection dataset (DSEL) that is used to extract the meta-features. The set DSEL was generated with 500 samples, 250 for each class.}
	\label{fig:DynamicSelectioDataset}	  
\end{figure}

In order to compute the region of competence and extract the meta-features for the given query sample, the dynamic selection dataset (DSEL) is used in the generalization phase. The dynamic selection dataset is shown in Figure~\ref{fig:DynamicSelectioDataset}.

\begin{figure}[H]
	\centering
	\subfigure[All samples]{\includegraphics[width=3.2in,clip=]{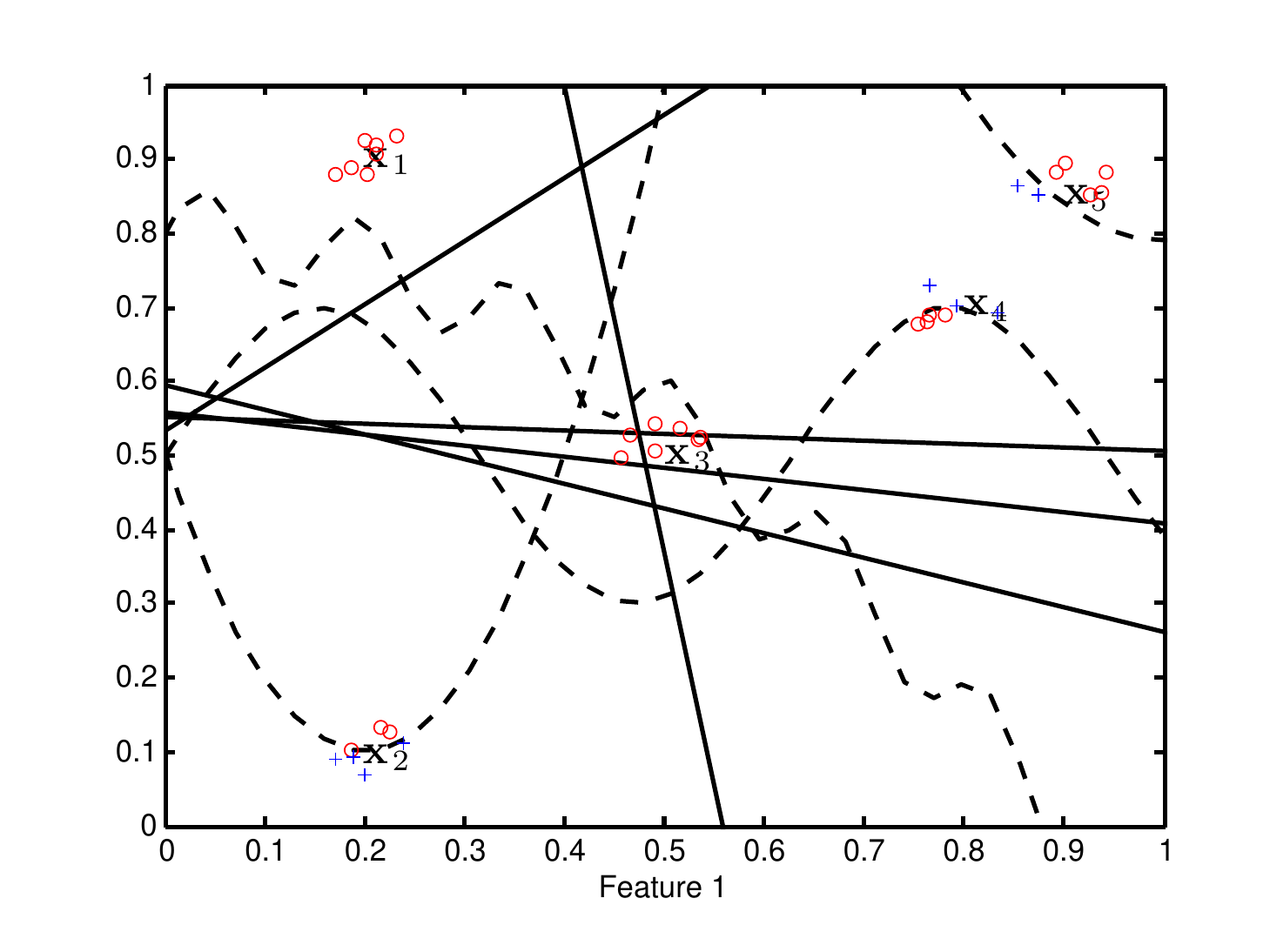}} 
	\subfigure[Neighborhood of $\mathbf{x}_{1}$ in DSEL]{\includegraphics[width=3.2in,clip=]{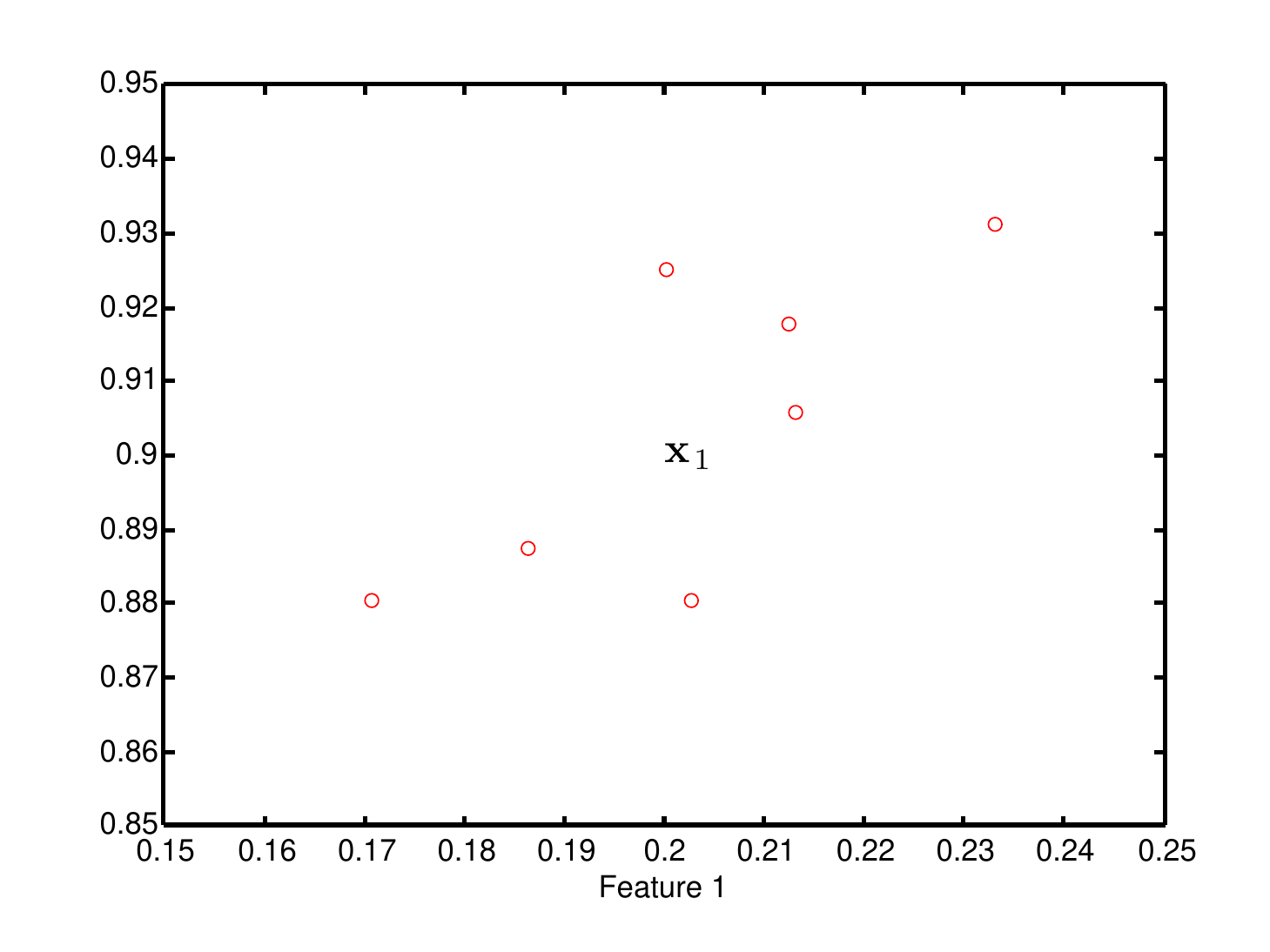}} 
	\subfigure[Neighborhood of $\mathbf{x}_{2}$ in DSEL]{\includegraphics[width=3.2in,clip=]{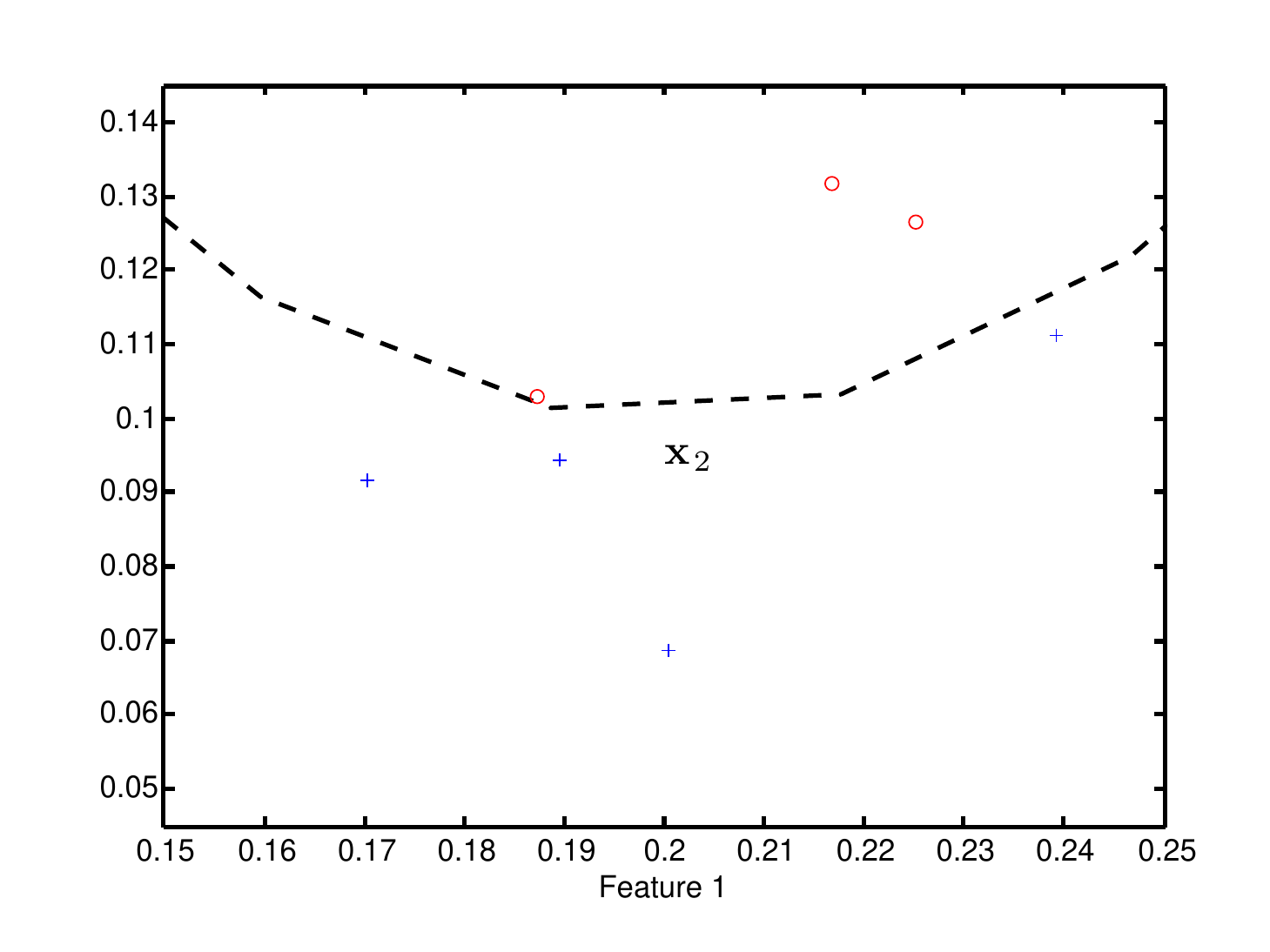}} 
	\subfigure[Neighborhood of $\mathbf{x}_{3}$ in DSEL]{\includegraphics[width=3.2in,clip=]{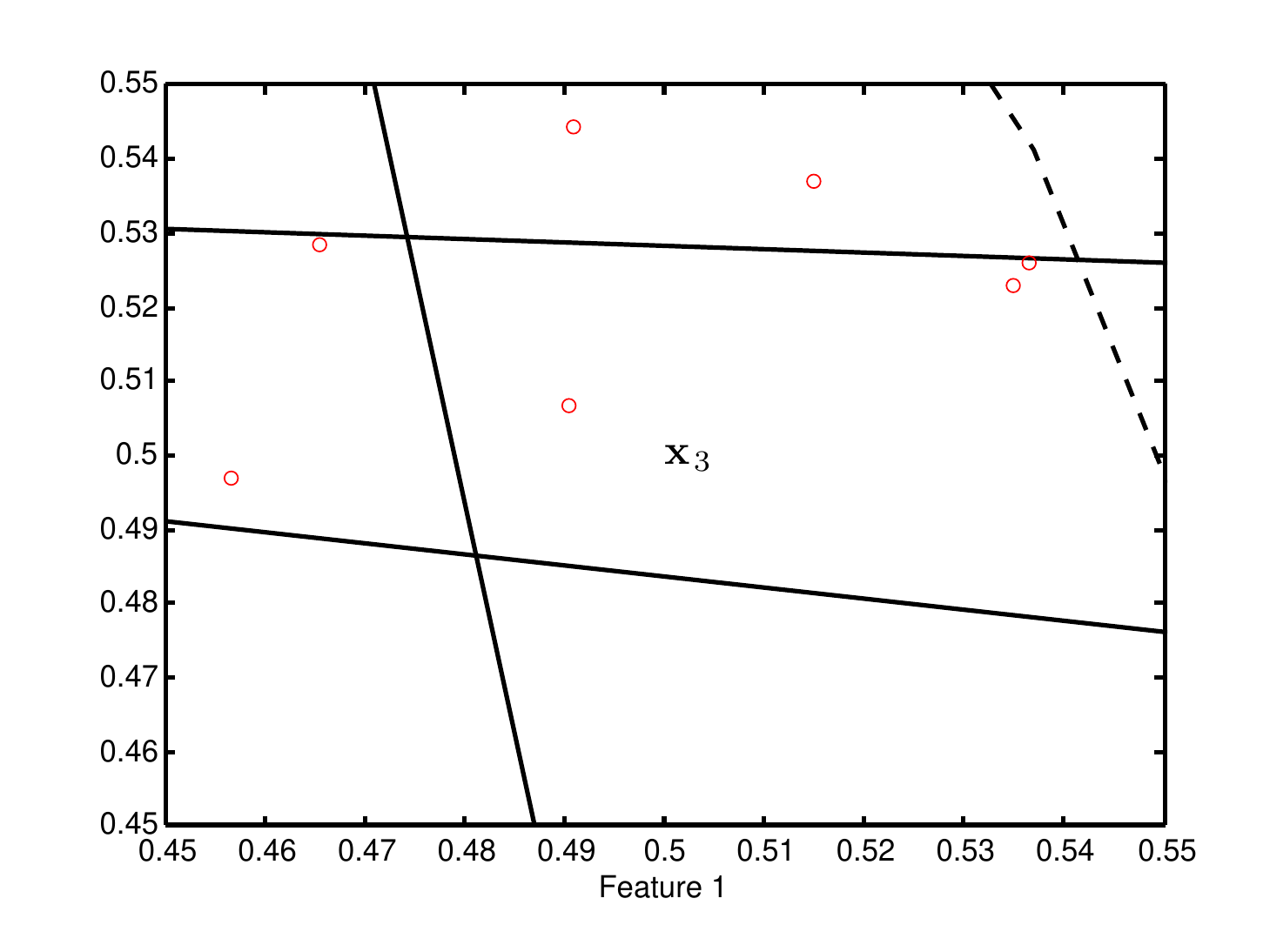}} 
	\subfigure[Neighborhood of $\mathbf{x}_{4}$ in DSEL]{\includegraphics[width=3.2in,clip=]{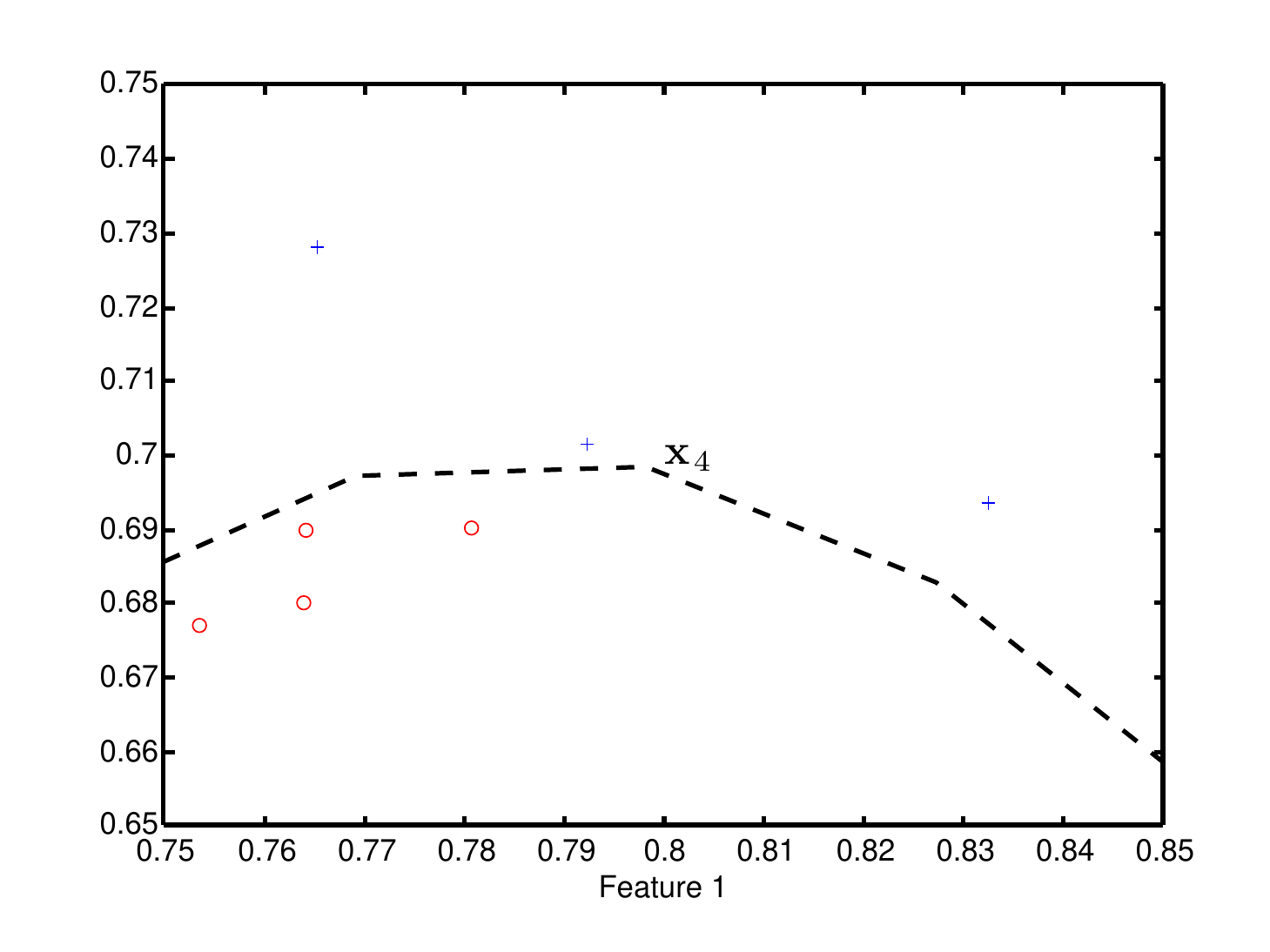}} 
	\subfigure[Neighborhood of $\mathbf{x}_{5}$ in DSEL]{\includegraphics[width=3.2in,clip=]{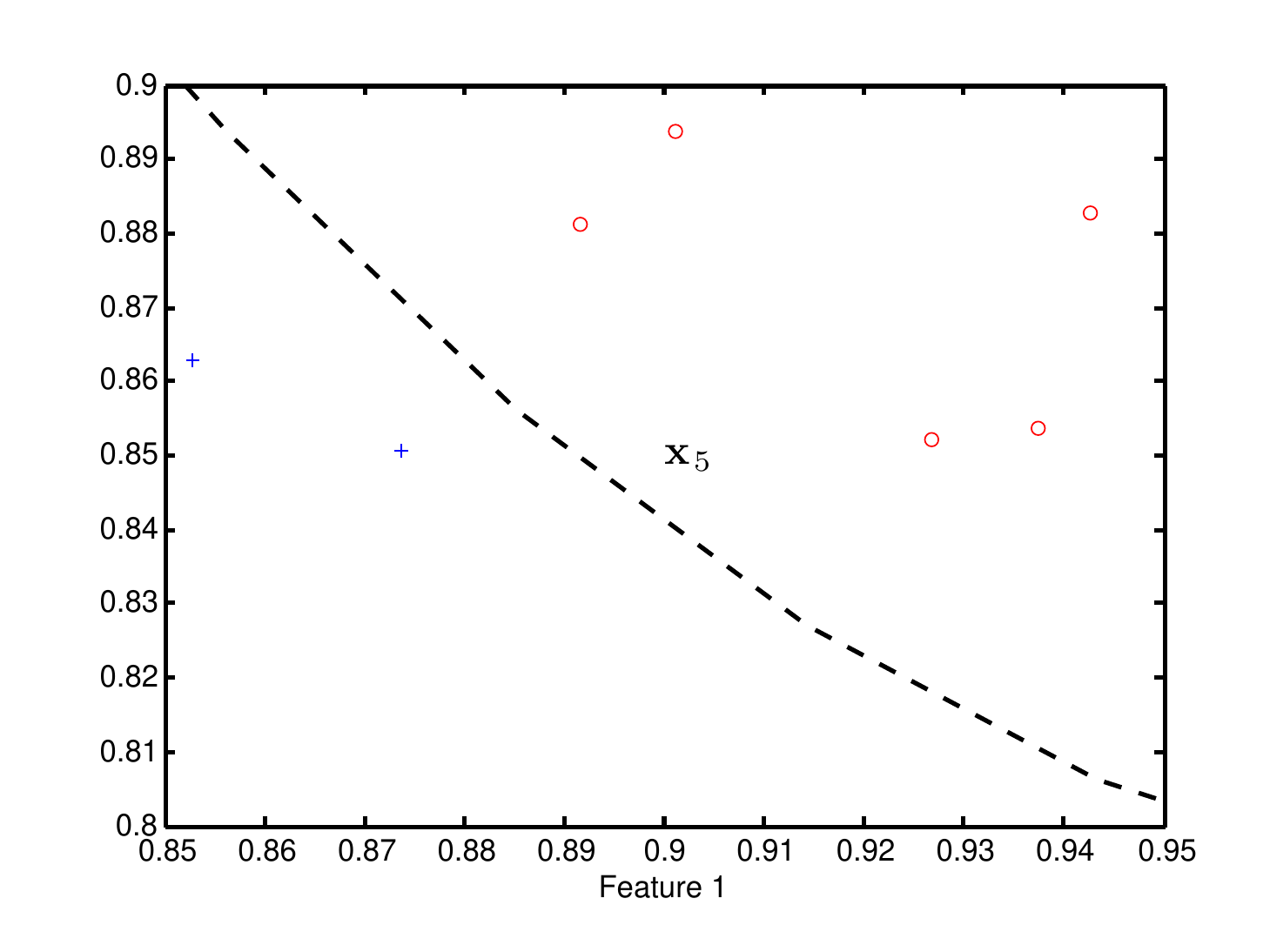}} 
	\caption{Local regions computed using the K-Nearest Neighbor algorithm in the feature space. The region of competence of each testing sample is shown in one sub-figure}
	\label{fig:P2IndividualExample}	  
\end{figure}

As in our previous papers~\cite{CruzPR,icpr2014}, we consider the size of the region of competence $K = 7$, i.e., the seven nearest neighbors of the query sample, and the size of the output profiles set $K_{p} = 5$. Figure~\ref{fig:P2IndividualExample} shows the regions of competence of each training sample. The samples belonging to the region of competence $\theta_{j}$, defined using DSEL, are shown for each testing sample separately (Figures~\ref{fig:P2IndividualExample} (b) to Figure~\ref{fig:P2IndividualExample} (f)).
 
%Based on the five query samples and the five base classifiers, the meta-feature vectors $v_{i,j}$ are computed. for each pair (base classifier $c_{i}$, sample $\mathbf{x}_{j}$) are extracted
For each test sample $\mathbf{x}_{j}$, five meta-feature vectors are extracted, each one corresponding to the behavior of one base classifier ($c_{1}$ to $c_{5}$) for the classification of $\mathbf{x}_{j}$. Tables~\ref{tabx1} to~\ref{tabx5} present the meta-feature vectors obtained for each test sample and base classifier. For each instance $\mathbf{x}_{j}$, we present the meta-feature vectors computed for each of the 5 base classifiers as well as the decision obtained by the meta-classifier, denoted by $\delta_{i,j}$. $\delta_{i,j} = 1$ means that the base classifier was considered competent, and was thus used to predict the label of the query sample.

For the sample $\mathbf{x}_{1}$ (Table~\ref{tabx1}), it is an easier classification case since it is located close to the mean of one of the class centers ($w_{1}$). We can see in Figure~\ref{fig:P2IndividualExample} (b) that all instances in the region of competence of $\mathbf{x}_{1}$ belong to the same class. The classifiers $c_{1}, c_{3} $ and $c_{4}$ achieve a 100\% recognition rate in the local region (as can be seen in Figure~\ref{fig:PerceptronsIndividual}). This also holds true for the decision space, where those base classifiers present the correct label for the most similar output profiles as well. Thus it is clear that they are competent for the classification of $\mathbf{x}_{1}$.

\begin{table}[H]
	\centering
	\caption{Meta-Features extracted for the sample $\mathbf{x}_{1}$}
	\label{tabx1}
	\resizebox{1.00\textwidth}{!}{
		\begin{tabular}{l|l|l|l|l|l|l|l|l|l|l|l|l|l|l|l|l|l|l|l|l|l|c|}
			\cline{2-23}
			& \multicolumn{7}{ |c| }{$f_{1}$} & \multicolumn{7}{ |c| }{$f_{2}$} & $f_{3}$ & \multicolumn{5}{ |c| }{$f_{4}$} & $f_{5}$ & $\delta_{i,j}$ \\
			\cline{2-23}
			$c_{1}$ & 1  & 1 & 1 & 1 & 1 & 1 & 1 & 0.65 & 0.66 & 0.59 & 0.62 & 0.66 & 0.76 & 0.61 & 1.00 & 1  & 1 & 1 & 1 & 1 & 0.97 &  1          \\
			$c_{2}$ & 0  & 0 & 0 & 0 & 0 & 0 & 0 & 0.39 & 0.38 & 0.31 & 0.34 & 0.38 & 0.35 & 0.06 & 0.00 & 0  & 0 & 0 & 0 & 0 & 0.87 &   0          \\
			$c_{3}$ & 1  & 1 & 1 & 1 & 1 & 1 & 1 & 0.84 & 0.81 & 0.77 & 0.81 & 0.82 & 0.91 & 0.82 & 1.00 & 1  & 1 & 1 & 1 & 1 & 0.99 &   1          \\
			$c_{4}$ & 1  & 1 & 1 & 1 & 1 & 1 & 1 & 0.79 & 0.78 & 0.73 & 0.76 & 0.79 & 0.88 & 0.77 & 1.00 & 1  & 1 & 1 & 1 & 1 & 0.98 &  1          \\
			$c_{5}$ & 0  & 0 & 0 & 0 & 0 & 0 & 0 & 0.30 & 0.32 & 0.24 & 0.24 & 0.29 & 0.37 & 0.23 & 0.00 & 0  & 0 & 0 & 0 & 0 & 0.87 &  0          \\
			\cline{2-23}
		\end{tabular} }
	\end{table}
	
For the classification of the instance $\mathbf{x}_{2}$, we can see that it is located closer to the border separating the two classes. We can see that there are samples in the region of competence of $\mathbf{x}_{2}$ belonging to both classes. The base classifiers that achieve a good performance considering both the validation samples in the region of competence $\theta_{j}$ and the most similar output profiles, meta-feature $f_{4}$, are considered competent.
	
	\begin{table}[H]
		\centering
		\caption{Meta-Features extracted for the sample $\mathbf{x}_{2}$}
		\label{tabx2}
		\resizebox{1.00\textwidth}{!}{
			\begin{tabular}{l|l|l|l|l|l|l|l|l|l|l|l|l|l|l|l|l|l|l|l|l|l|c|}
				\cline{2-23}
				& \multicolumn{7}{ |c| }{F1} & \multicolumn{7}{ |c| }{F2} & F3 & \multicolumn{5}{ |c| }{F4} & F5 & $\delta_{i,j}$ \\
				\cline{2-23}
				$c_{1}$ & 1  & 0 & 1 & 1 & 0 & 0 & 1 & 1.00 & 0.00 & 0.97 & 1.00 & 0.00 & 0.00 & 0.89 & 0.57 & 1  & 0 & 1 & 1 & 1 & 1.00 & 1          \\
				$c_{2}$ & 1  & 0 & 1 & 1 & 0 & 0 & 1 & 0.67 & 0.32 & 0.63 & 0.62 & 0.33 & 0.39 & 0.59 & 0.57 & 1  & 0 & 1 & 1 & 1 & 0.97 &  1          \\
				$c_{3}$ & 1  & 0 & 1 & 1 & 0 & 0 & 1 & 0.89 & 0.11 & 0.87 & 0.87 & 0.14 & 0.17 & 0.81 & 0.57 & 1  & 0 & 1 & 1 & 1 & 0.99 &  1          \\
				$c_{4}$ & 1  & 0 & 1 & 1 & 0 & 0 & 1 & 0.86 & 0.13 & 0.83 & 0.85 & 0.15 & 0.18 & 0.81 & 0.57 & 1  & 0 & 1 & 1 & 1 & 0.99 &  1          \\
				$c_{5}$ & 0  & 1 & 0 & 0 & 1 & 1 & 0 & 0.28 & 0.70 & 0.19 & 0.20 & 0.67 & 0.79 & 0.23 & 0.43 & 0  & 1 & 0 & 0 & 0 & 0.87 &  0          \\
				\cline{2-23}
			\end{tabular} }
		\end{table}

The sample $\mathbf{x}_{3}$ is located in a region close to the lines generated by the Perceptrons $c_{2}$, $c_{3}$, $c_{4}$ and $c_{5}$. However, all neighbor samples of $\mathbf{x}_{3}$ belong to the same class. Thus, the classifiers that achieve a good performance in the region of competence $\theta_{j}$, and also for the set $\phi_{j}$ with the most the similar outputs profiles of $\tilde{\mathbf{x}}_{3}$, are selected. It is important to note that, in contrast to the testing instances $\mathbf{x}_{1}$ and $\mathbf{x}_{2}$, we can see that both the posterior probability meta-feature, meta-feature $f_{2}$, and the classifier's confidence, meta-feature $f_{5}$, produce lower results than the ones presented in Tables~\ref{tabx1} and~\ref{tabx2} since the samples are closer to the decision boundary of the base classifiers.
		
		\begin{table}[H]
			\centering
			\caption{Meta-Features extracted for the sample $\mathbf{x}_{3}$}
			\label{tabx3}
			\resizebox{1.00\textwidth}{!}{
				\begin{tabular}{l|l|l|l|l|l|l|l|l|l|l|l|l|l|l|l|l|l|l|l|l|l|c|}
					\cline{2-23}
					& \multicolumn{7}{ |c| }{F1} & \multicolumn{7}{ |c| }{F2} & F3 & \multicolumn{5}{ |c| }{F4} & F5 & $\delta_{i,j}$ \\
					\cline{2-23}
					$c_{1}$ & 0  & 0 & 0 & 0 & 0 & 0 & 0 & 0.12 & 0.13 & 0.00 & 0.06 & 0.15 & 0.13 & 0.08 & 0.00 & 0  & 0 & 0 & 0 & 1 & 0.39 &   0          \\
					$c_{2}$ & 1  & 1 & 1 & 0 & 0 & 1 & 1 & 0.60 & 0.51 & 0.46 & 0.43 & 0.47 & 0.60 & 0.45 & 0.71 & 1  & 0 & 1 & 1 & 0 & 0.66 &   1          \\
					$c_{3}$ & 1  & 1 & 1 & 1 & 1 & 1 & 1 & 0.56 & 0.60 & 0.53 & 0.49 & 0.56 & 0.68 & 0.54 & 1.00 & 1  & 1 & 1 & 1 & 0 & 0.66 & 1          \\
					$c_{4}$ & 0  & 1 & 0 & 0 & 0 & 0 & 1 & 0.47 & 0.52 & 0.43 & 0.41 & 0.48 & 0.57 & 0.45 & 0.29 & 0  & 0 & 0 & 0 & 0 & 0.36 &  0          \\
					$c_{5}$ & 0  & 0 & 0 & 0 & 0 & 0 & 0 & 0.48 & 0.47 & 0.40 & 0.43 & 0.46 & 0.44 & 0.41 & 0.00 & 0  & 0 & 0 & 0 & 1 & 0.36 &  0          \\
					\cline{2-23}
				\end{tabular} }
			\end{table}

For the sample $\mathbf{x}_{4}$ (Table~\ref{tabx4}), we can see that the majority of its neighbor samples come from a different class (Figure~\ref{fig:P2IndividualExample} (d)). If we consider dynamic selection techniques that are based solely on accuracy information, such as local classifier accuracy (LCA)~\cite{lca} or overall classifier accuracy (OLA)~\cite{lca}, as well as the \textit{a priori} and a posteriori methods~\cite{DidaciGRM05}, the base classifiers $c_{2}$, $c_{3}$ and $c_{4}$ are considered the most competent. So, using only the accuracy information in the local regions (region of competence) may not be sufficient to select the competent classifiers. However, these three classifiers predict the wrong label for $\mathbf{x}_{4}$; as shown in Figure~\ref{fig:PerceptronsIndividual}, they would predict that $\mathbf{x}_{4}$ belongs to class 1 (red circle).

			\begin{table}[H]
				\centering
				\caption{Meta-Features extracted for the sample $\mathbf{x}_{4}$}
				\label{tabx4}
				\resizebox{1.00\textwidth}{!}{
					\begin{tabular}{l|l|l|l|l|l|l|l|l|l|l|l|l|l|l|l|l|l|l|l|l|l|c|}
						\cline{2-23}
						& \multicolumn{7}{ |c| }{F1} & \multicolumn{7}{ |c| }{F2} & F3 & \multicolumn{5}{ |c| }{F4} & F5 & $\delta_{i,j}$ \\
						\cline{2-23}
						$c_{1}$ & 1  & 0 & 1 & 0 & 0 & 1 & 0 & 0.92 & 0.37 & 0.93 & 0.00 & 0.37 & 1.00 & 0.00 & 0.43 & 1 & 1 & 1 & 0 & 0 & 0.99 &  1          \\
						$c_{2}$ & 0  & 1 & 0 & 1 & 1 & 0 & 1 & 0.42 & 0.66 & 0.22 & 0.60 & 0.63 & 0.39 & 0.59 & 0.57 & 0 & 0 & 0 & 1 & 1 & 0.90 & 0          \\
						$c_{3}$ & 0  & 1 & 0 & 1 & 1 & 0 & 1 & 0.16 & 0.81 & 0.06 & 0.77 & 0.77 & 0.17 & 0.75 & 0.57 & 0 & 0 & 0 & 1 & 1 & 0.90 &  0          \\
						$c_{4}$ & 0  & 1 & 0 & 1 & 1 & 0 & 1 & 0.34 & 0.64 & 0.27 & 0.59 & 0.61 & 0.37 & 0.58 & 0.57 & 0 & 0 & 0 & 1 & 1 & 0.90 &   0          \\
						$c_{5}$ & 1  & 0 & 1 & 0 & 0 & 1 & 0 & 0.62 & 0.37 & 0.57 & 0.31 & 0.37 & 0.72 & 0.32 & 0.43 & 1 & 1 & 1 & 0 & 0 & 0.89 &  0          \\
						\cline{2-23}
					\end{tabular} }
				\end{table}

Through the use of different meta-features, the META-DES is able to select a competent classifier ($c_{1}$) for the sample $\mathbf{x}_{4}$. The base classifier $c_{1}$ achieves a better performance in the decision space, (meta-feature $f_{4}$) (it is able to predict the correct class label for the closest samples in the decision space). Since each output profile $\tilde{\mathbf{x}}_{k}$ in the decision space is associated with a sample $\mathbf{x}_{k}$ in the feature space, we present the most similar output profiles of the sample $\tilde{\mathbf{x}}_{4}$. We can see that computing the similarity using the decision space yields distinct results, i.e., different validation samples are selected for extracting the meta-features. In this case, the closest output profiles, selected in the decision space, are from samples that belong to the same class of $\mathbf{x}_{4}$. So, the meta-features extracted using those samples are more likely to reflect the behavior of the base classifier $c_{1}$ for the classification of the sample $\mathbf{x}_{4}$. In addition, the base classifier $c_{1}$ also presents a higher posterior probability for the correct class label (meta-feature $f_{2}$), and a higher confidence in its answer for the classification of the query sample $\mathbf{x}_{4}$ (meta-feature $f_{5}$) when compared to the other base classifiers. Thus, it is considered as a competent classifier for the classification of the sample $\mathbf{x}_{4}$.
			
It is important to mention that the base classifier $c_{5}$ also predicts the correct label for the sample $\mathbf{x}_{4}$. However, it was not considered as a competent classifier since it presented lower confidence in its prediction (meta-feature $f_{5}$) as well as lower results for $f_{2}$ when compared to $c_{1}$. 

				\begin{table}[H]
					\centering
					\caption{Meta-Features extracted for the sample $\mathbf{x}_{5}$}
					\label{tabx5}
					\resizebox{1.00\textwidth}{!}{
						\begin{tabular}{l|l|l|l|l|l|l|l|l|l|l|l|l|l|l|l|l|l|l|l|l|l|c|}
							\cline{2-23}
							& \multicolumn{7}{ |c| }{F1} & \multicolumn{7}{ |c| }{F2} & F3 & \multicolumn{5}{ |c| }{F4} & F5 & $\delta_{i,j}$ \\
							\cline{2-23}
							$c_{1}$ & 1  & 0 & 0 & 0 & 0 & 1 & 0 & 0.85 & 0.10 & 0.05 & 0.01 & 0.14 & 0.94 & 0.04 & 0.29 & 0  & 1 & 0 & 1 & 0 & 0.99 &  0          \\
							$c_{2}$ & 0  & 1 & 1 & 1 & 1 & 0 & 1 & 0.27 & 0.74 & 0.68 & 0.70 & 0.71 & 0.33 & 0.71 & 0.71 & 1  & 0 & 1 & 0 & 1 & 0.98 &  1          \\
							$c_{3}$ & 0  & 1 & 1 & 1 & 1 & 0 & 1 & 0.00 & 1.00 & 1.00 & 0.98 & 1.00 & 0.06 & 1.00 & 0.71 & 1  & 0 & 1 & 0 & 1 & 1.00 &  1          \\
							$c_{4}$ & 0  & 1 & 1 & 1 & 1 & 0 & 1 & 0.21 & 0.78 & 0.76 & 0.74 & 0.79 & 0.24 & 0.76 & 0.71 & 1  & 0 & 1 & 0 & 1 & 0.98 &  1          \\
							$c_{5}$ & 1  & 0 & 0 & 0 & 0 & 1 & 0 & 0.70 & 0.28 & 0.18 & 0.21 & 0.25 & 0.81 & 0.20 & 0.29 & 0  & 1 & 0 & 1 & 0 & 0.97 &  0 \\
							\cline{2-23}
						\end{tabular} }
					\end{table}		

Considering these five testing samples, an interesting fact we can obtain from this example is the influence of using the decision space for estimating the competence of the base classifiers, especially considering the closest output profile (which holds the first position in the vector $f_{4}$). Based on Tables~\ref{tabx1} to~\ref{tabx5}, when the base classifier predicts the correct label for the closest (first) output profile of the query sample, the probability of the base classifier being selected as competent is high.

\begin{figure}[H]
   \begin{center}  	 
       	  \includegraphics[clip=,  width=0.8\textwidth]{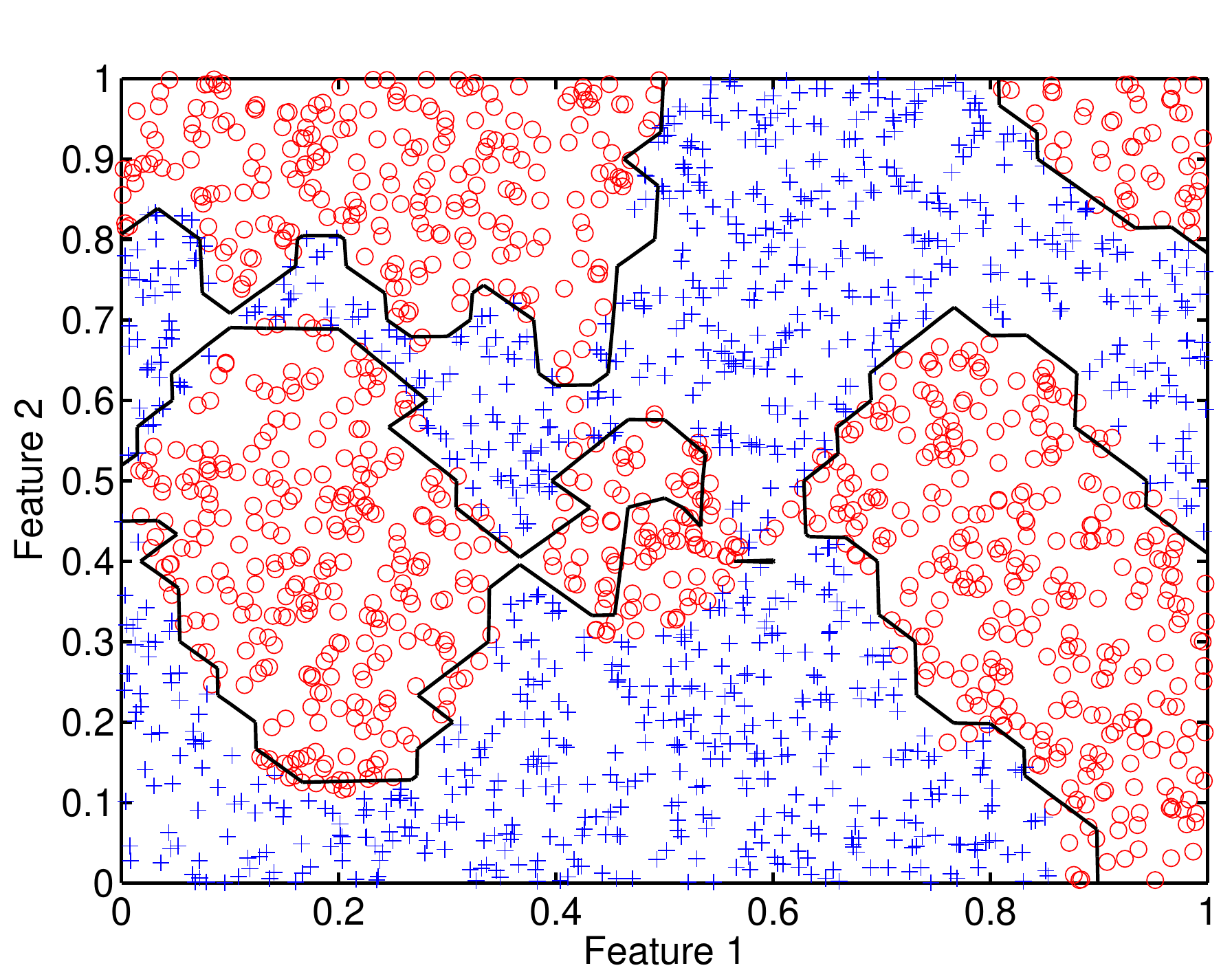}
   \end{center}
\caption{Decision Boundary obtained by the META-DES system using a pool of 5 Perceptrons. The META-DES achieves a recognition rate of 95.50\% using 5 Perceptrons.}
\label{fig:DecisionMETA5}	  
\end{figure}

Figure~\ref{fig:DecisionMETA5} illustrates the decision boundary obtained by the META-DES framework. Using only five linear weak classifiers and dynamic selection, we can approximate the complex decision boundary of the P2 problem. The methodology used to define the decision boundary obtained by the technique is presented in~\ref{sec:decisionboundary}.

%------------------------------------------------------------------------------------------------------------

\begin{figure}[!ht]
	\centering
	\subfigure[Voting decision]{\includegraphics[width=3.15in,clip=]{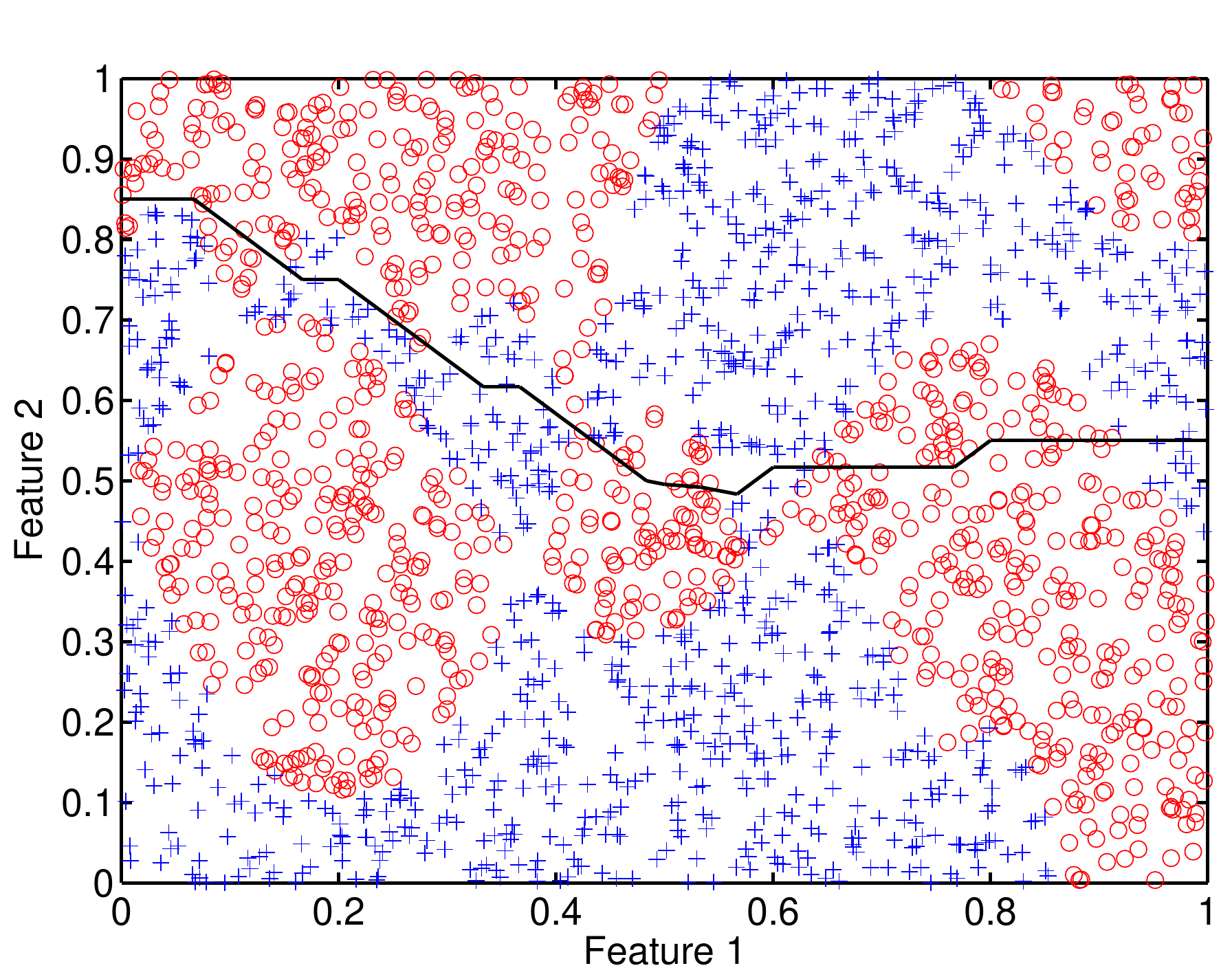}} 
	\subfigure[Averaging decision]{\includegraphics[width=3.15in,clip=]{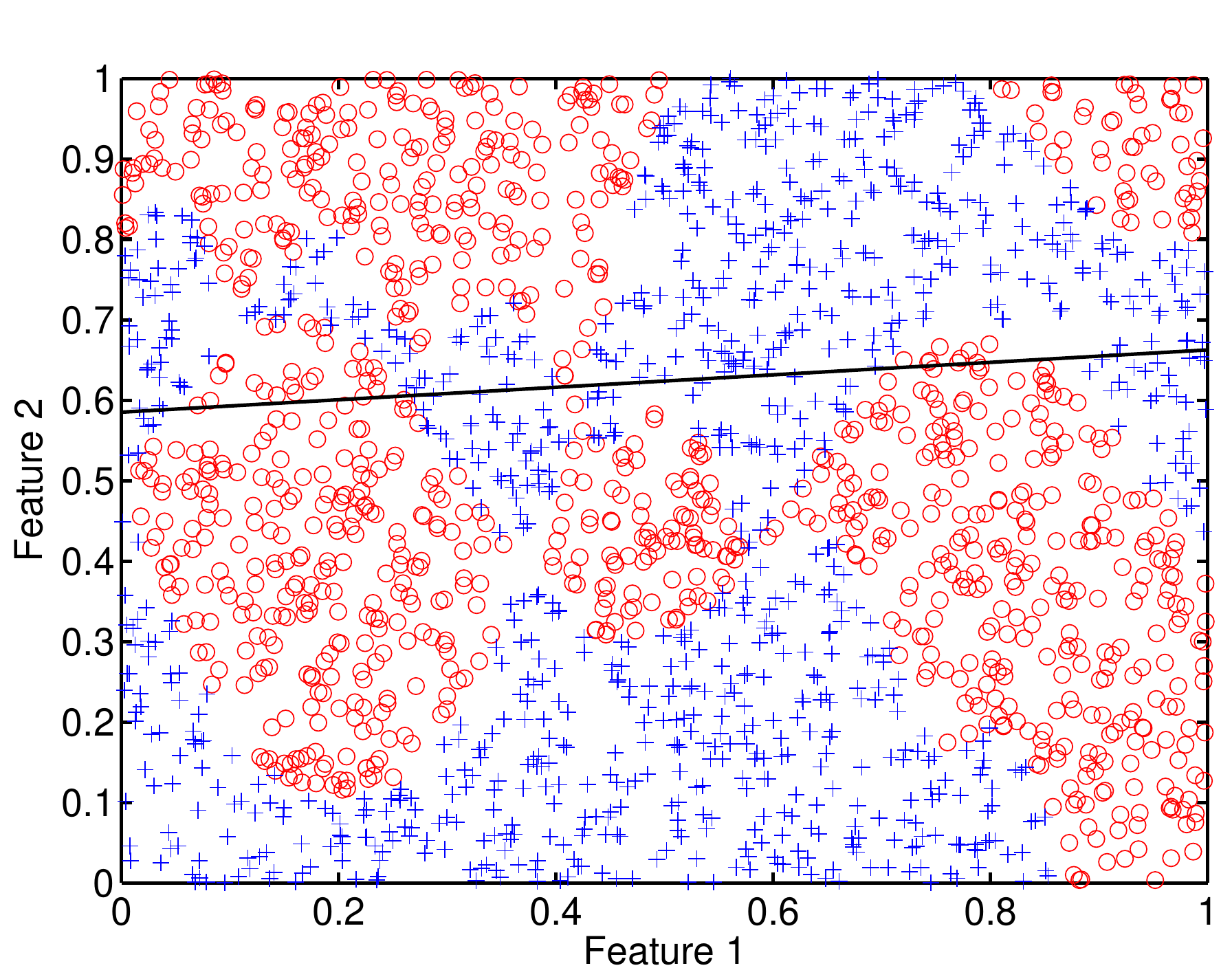}} 
	\subfigure[Maximum decision]{\includegraphics[width=3.15in,clip=]{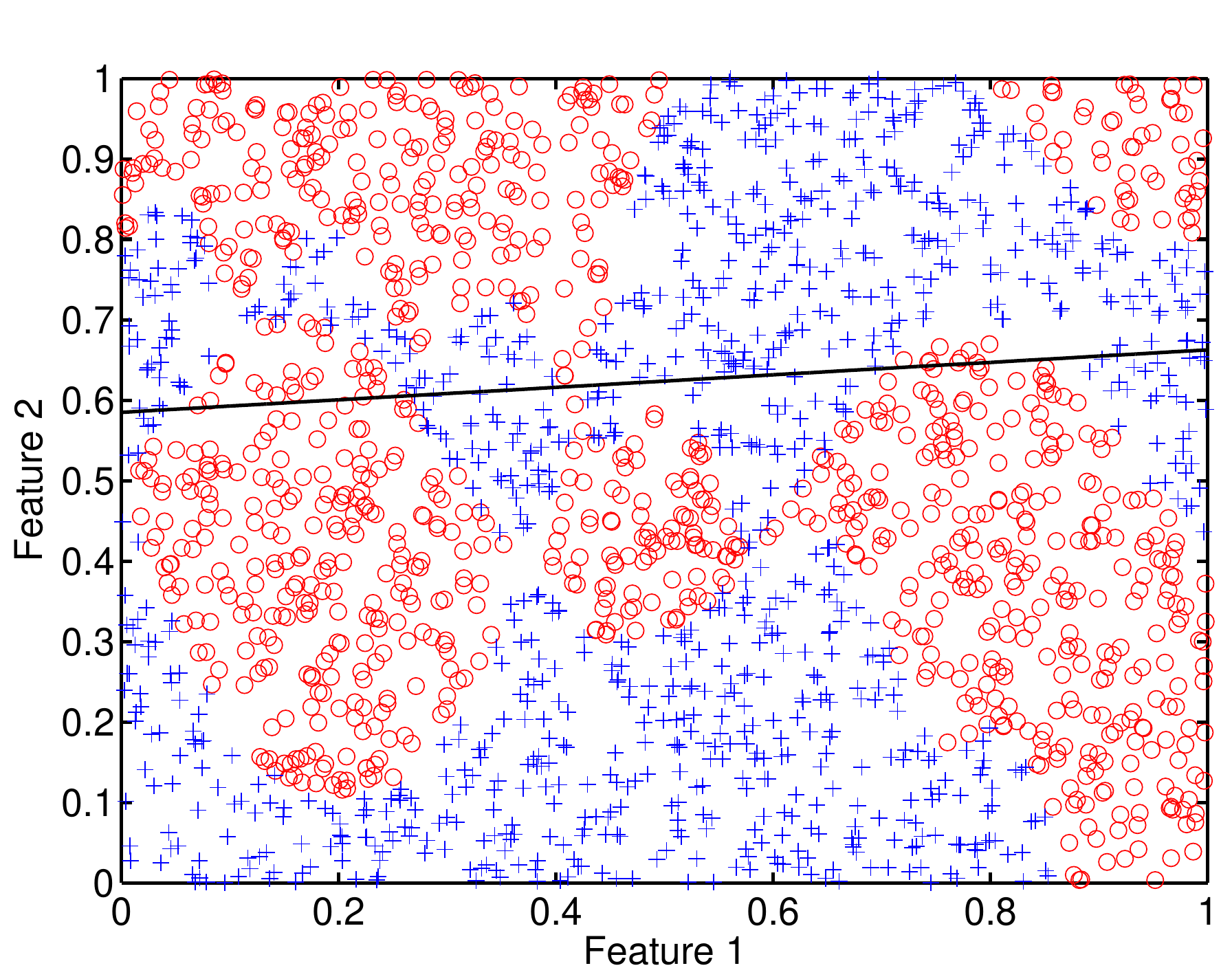}} 
	\subfigure[Product decision]{\includegraphics[width=3.15in,clip=]{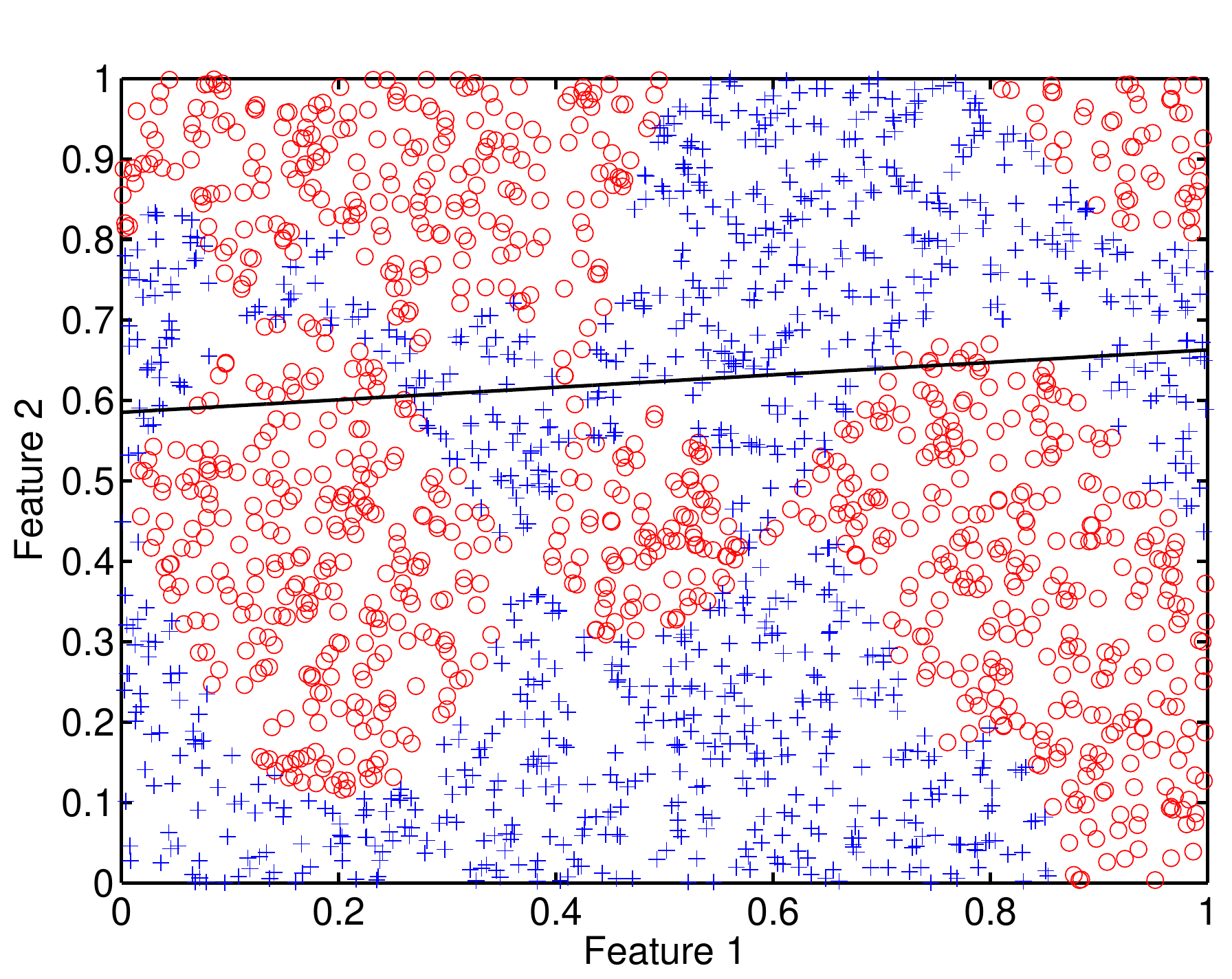}} 
	\subfigure[Adaboost decision]{\includegraphics[width=3.15in,clip=]{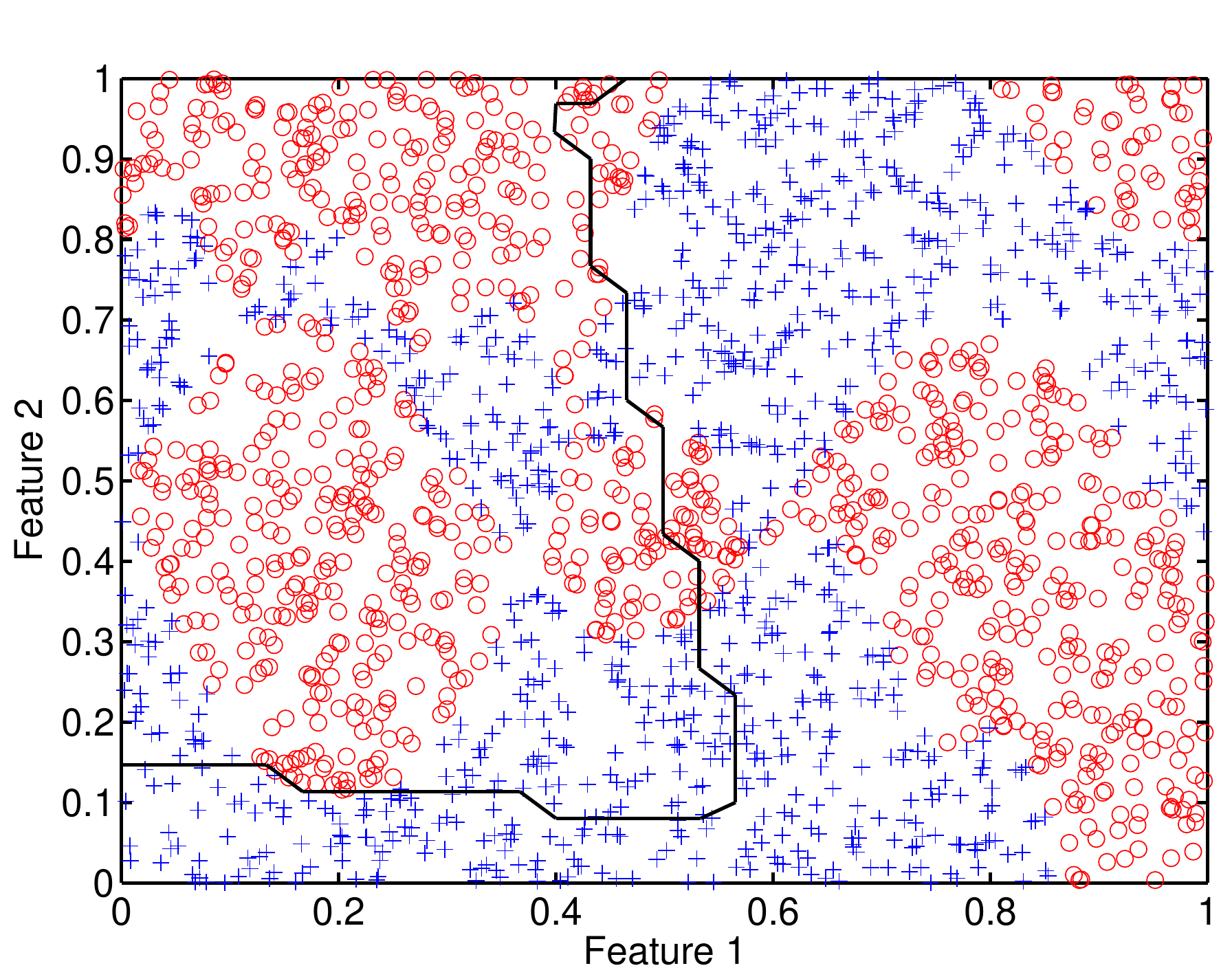}}
	\caption{Decision boundaries generated by each static combination method. The pool of classifiers is composed of the 5 Perceptrons presented in Figure~\ref{fig:FivePerceptrons}.}
	\label{fig:resultsP2StaticPerceptron5}	  
\end{figure}

When we apply static combination rules such as majority voting or Adaboost, the classification accuracy is much lower. Figure~\ref{fig:resultsP2StaticPerceptron5} illustrates the decision boundary obtained by static ensemble techniques using five Perceptron classifiers. We show the decisions obtained using the Average, Majority voting, Product, Maximum, as well as the Adaboost techniques. The average and product rules achieve a recognition rate of 47.5\%, while the maximum and majority voting rules obtain an accuracy of 50\%, and AdaBoost 56\%. This can be explained by the fact that all classifiers in the pool are used to predict the label. However, due to the complexity of the problem, the degree of disagreement between the classifiers is very high. For the majority of test samples, half of the base classifiers disagree with the other half (predicts a different class label). The decisions of classifiers that are not experts for the local region end up negatively influencing the final decision. Thus, the static combination rule yields results that are close to random guessing. Even using techniques that assign weights to the base classifiers, such as Adaboost, we cannot approximate the complex decision of the P2 problem using only five linear classifiers.

\section{Further Analysis}
\label{sec:experiments}

In this section, we evaluate the following aspects of the META-DES framework using the P2 problem:

\begin{enumerate}

\item The effect of the pool size on the classification accuracy.
\item The effect of the size of the dynamic selection dataset (DSEL) on the classification performance of the system.
\item The results of static the combination techniques for the P2 problem. This analysis is performed in order to provide an insight into why dynamic selection should be preferred for solving complex classification problems.
\item The results of classical pattern recognition techniques such as Support Vector Machines and Random Forest for the P2 problem.
\end{enumerate}

For the sake of simplicity, we use the same methodology used in the previous section: 500 samples for training ($\mathcal{T}$), 500 instances for the meta-training dataset ($\mathcal{T}_{\lambda}$), 500 instances for the dynamic selection dataset $DSEL$, and 2000 samples for testing, $\mathcal{G}$. For each set, the prior probabilities of both classes are equal. Moreover, since the objective of this work is to study whether dynamic selection of linear classifier can solve complex non-linear classification problems, we also consider Decision Stumps~\cite{Ai92inductionof} as base classifiers. We show that the META-DES framework works equally well using a pool of Decision Stumps. 

\subsection{The Effect of the Pool Size}

\begin{figure}[!ht]
   \begin{center}  	 
       	  \epsfig{file=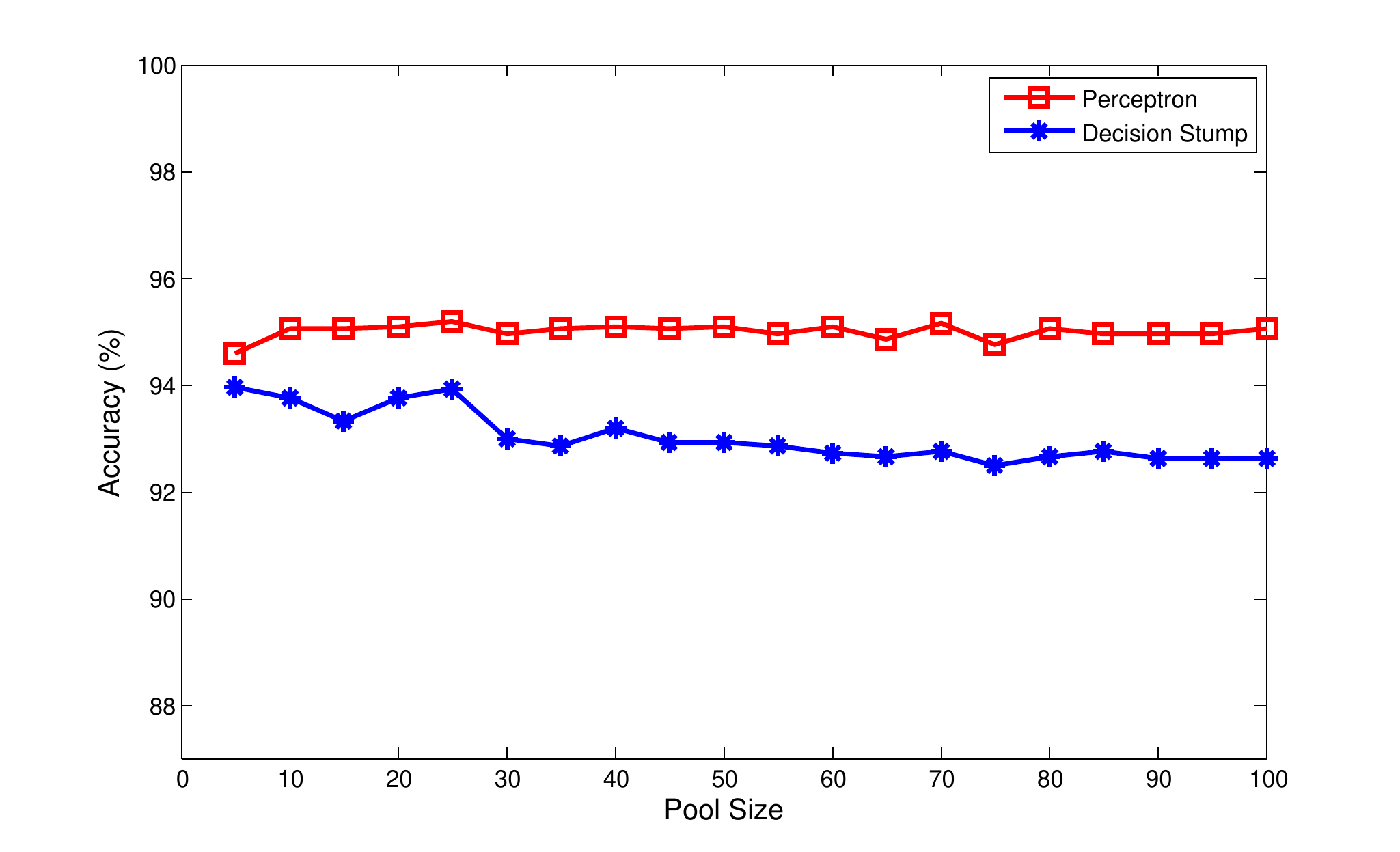, clip=,  width=0.70\textwidth}
   \end{center}
\caption{The effect of the pool size, $M$ in the classification accuracy. Perceptron and Decision Stumps are considered as base classifiers.}
\label{fig:2DPool}	  
\end{figure}

For this experiment, we varied the size of the pool from 5 to 100 at 5 point intervals (20 results are obtained). The size of the dynamic selection dataset (DSEL) was set at 500 (as shown in Figure~\ref{fig:DynamicSelectioDataset}). The effect of the size of the pool of classifiers, $M$, is shown in Figure~\ref{fig:2DPool}. We can see that the size of the pool does not have a significant impact on the classification accuracy of the META-DES, especially when the Perceptron is considered as the base classifier. This finding can be explained by the fact that using only 5 base classifiers, the Oracle (ideal selection scheme) achieves a classification accuracy of 99.5\% and 100\% using Perceptrons and Decision Stumps, respectively. In other words, using five base classifiers, it is possible to represent the whole feature space. The key to having good classification performance lies in defining a criterion to select the best classifier(s) for any given test sample. An interesting point is that the performance using decision stumps decreases as more classifiers are added to the pool, with the recognition performance decreasing when more than 25 base classifiers are used. Therefore, adding more classifiers does not always lead to higher classification accuracy. 

\begin{figure}[H]
	\centering
	\subfigure[5 Perceptrons]{\includegraphics[width=3.2in,clip=]{Images/Perceptron/METADES/DECISIONP2METADES5.pdf}} 
	\subfigure[10 Perceptrons]{\includegraphics[width=3.2in,clip=]{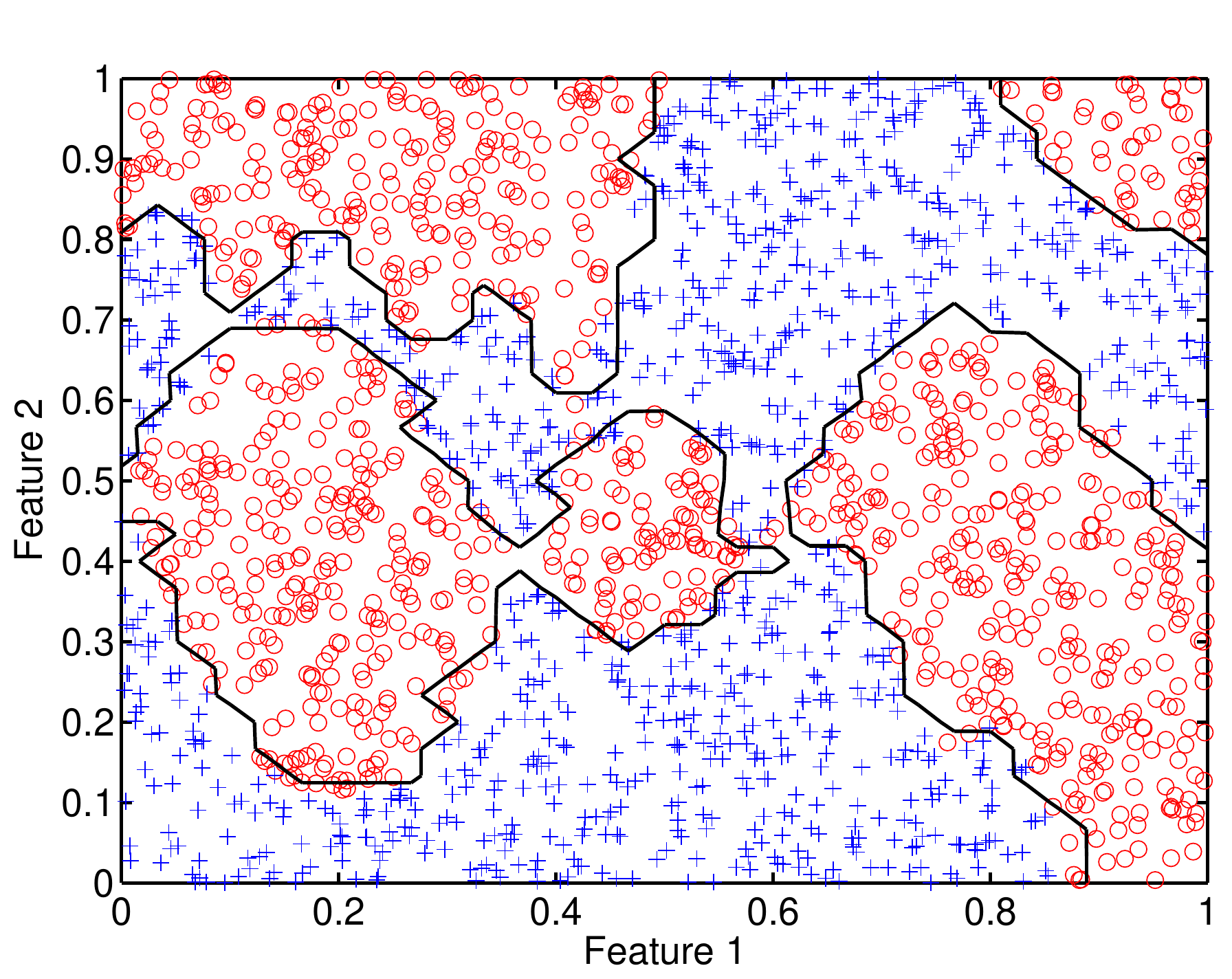}} 
	\subfigure[25 Perceptrons]{\includegraphics[width=3.2in,clip=]{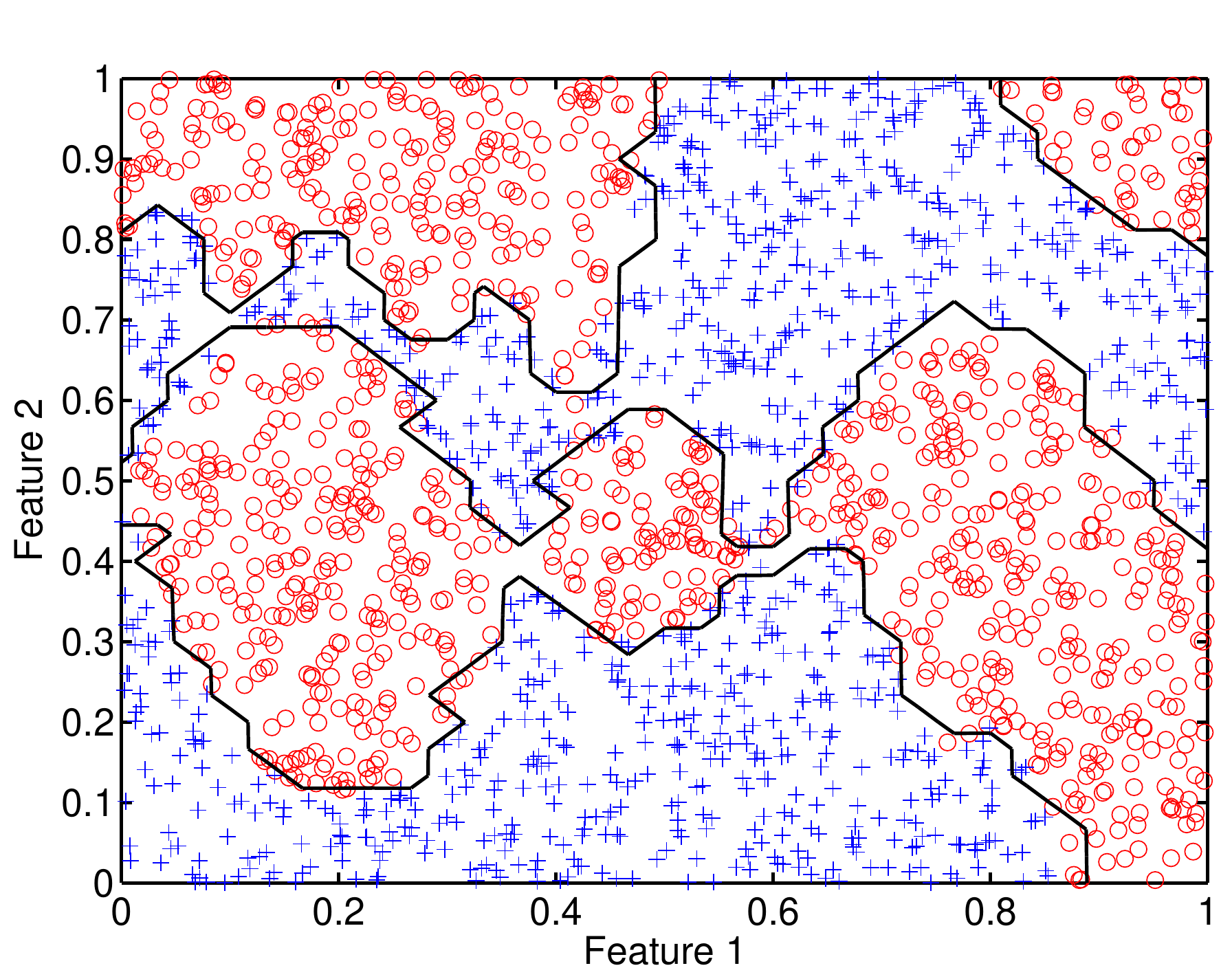}} 
	\subfigure[50 Perceptrons]{\includegraphics[width=3.2in,clip=]{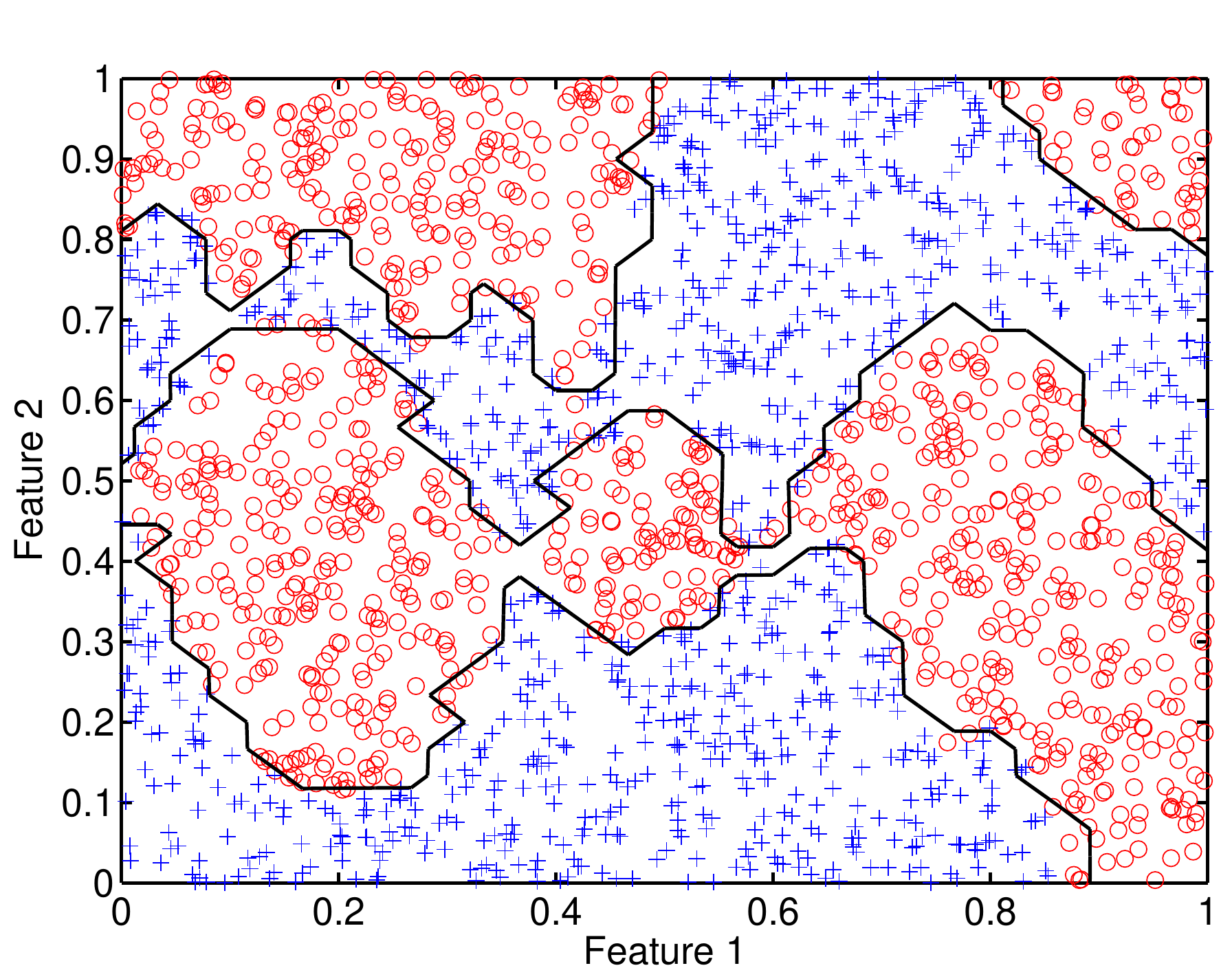}} 
	\subfigure[75 Perceptrons]{\includegraphics[width=3.2in,clip=]{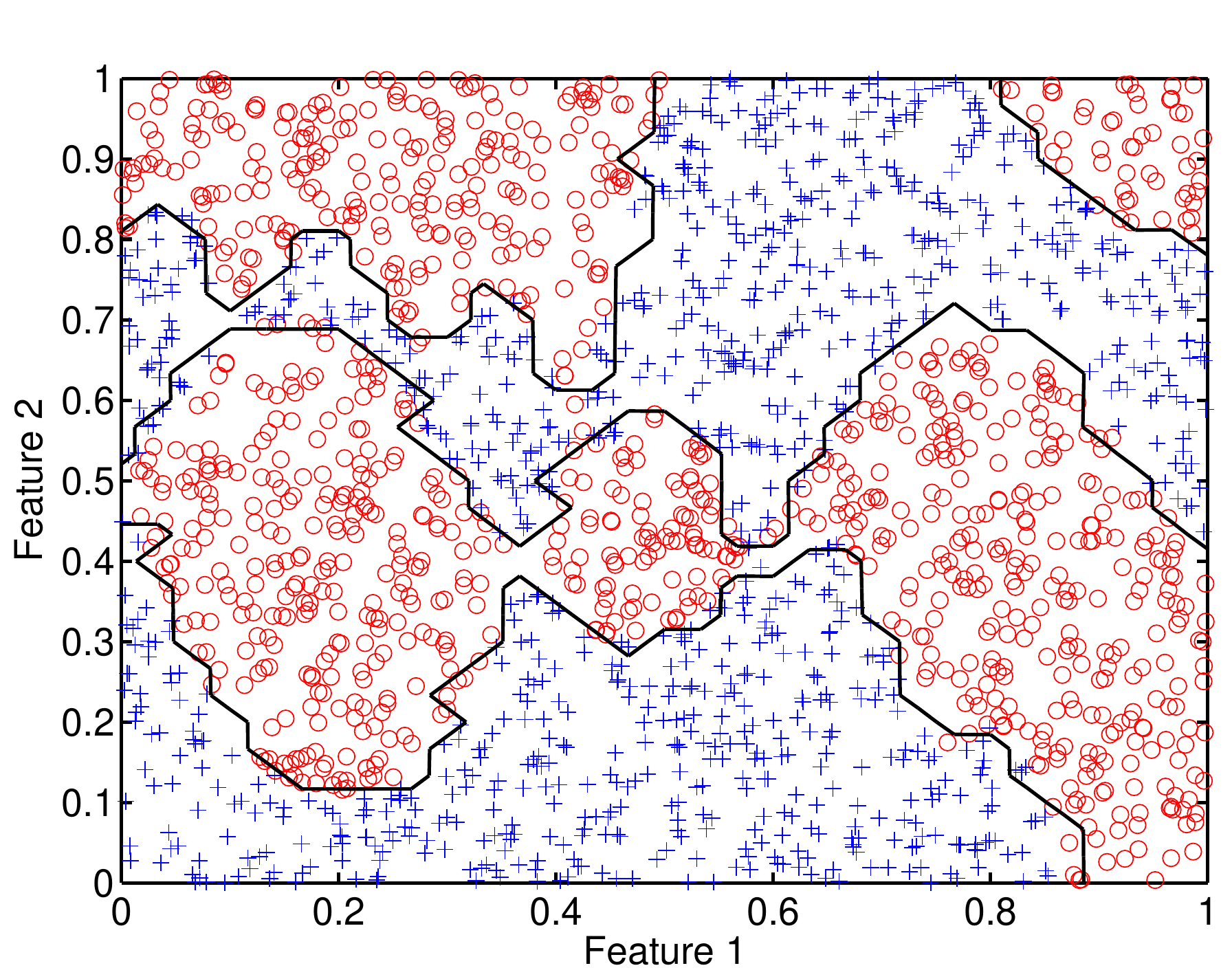}} 
	\subfigure[100 Perceptrons]{\includegraphics[width=3.2in,clip=]{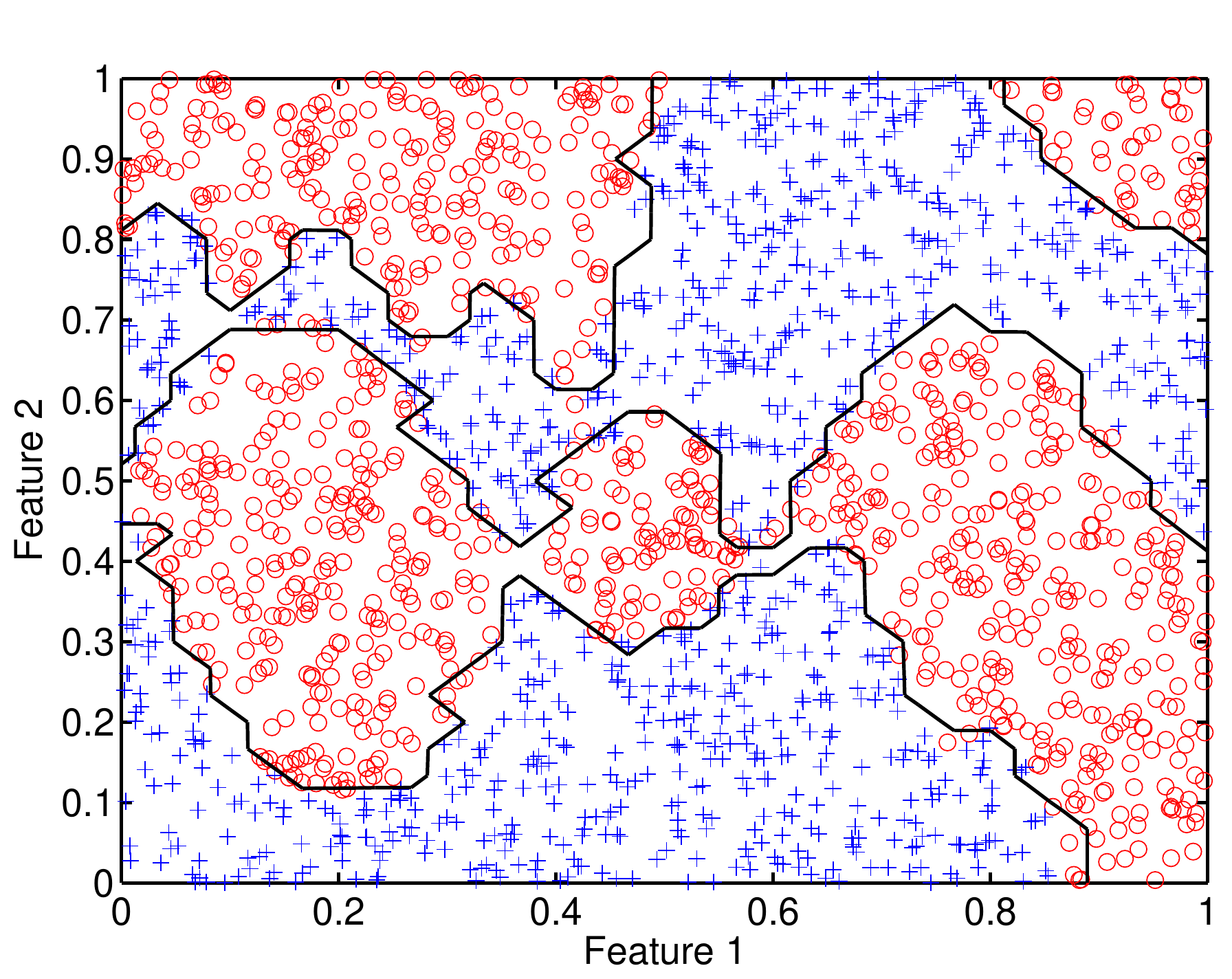}} 
	
	\caption{Decision boundaries generated by the META-DES framework for different pool size. Perceptrons are used as base classifiers.}
	\label{fig:resultsMETADES}	  
\end{figure}

\begin{figure}[H]
	\centering
	\subfigure[5 Stumps]{\includegraphics[width=3.2in,clip=]{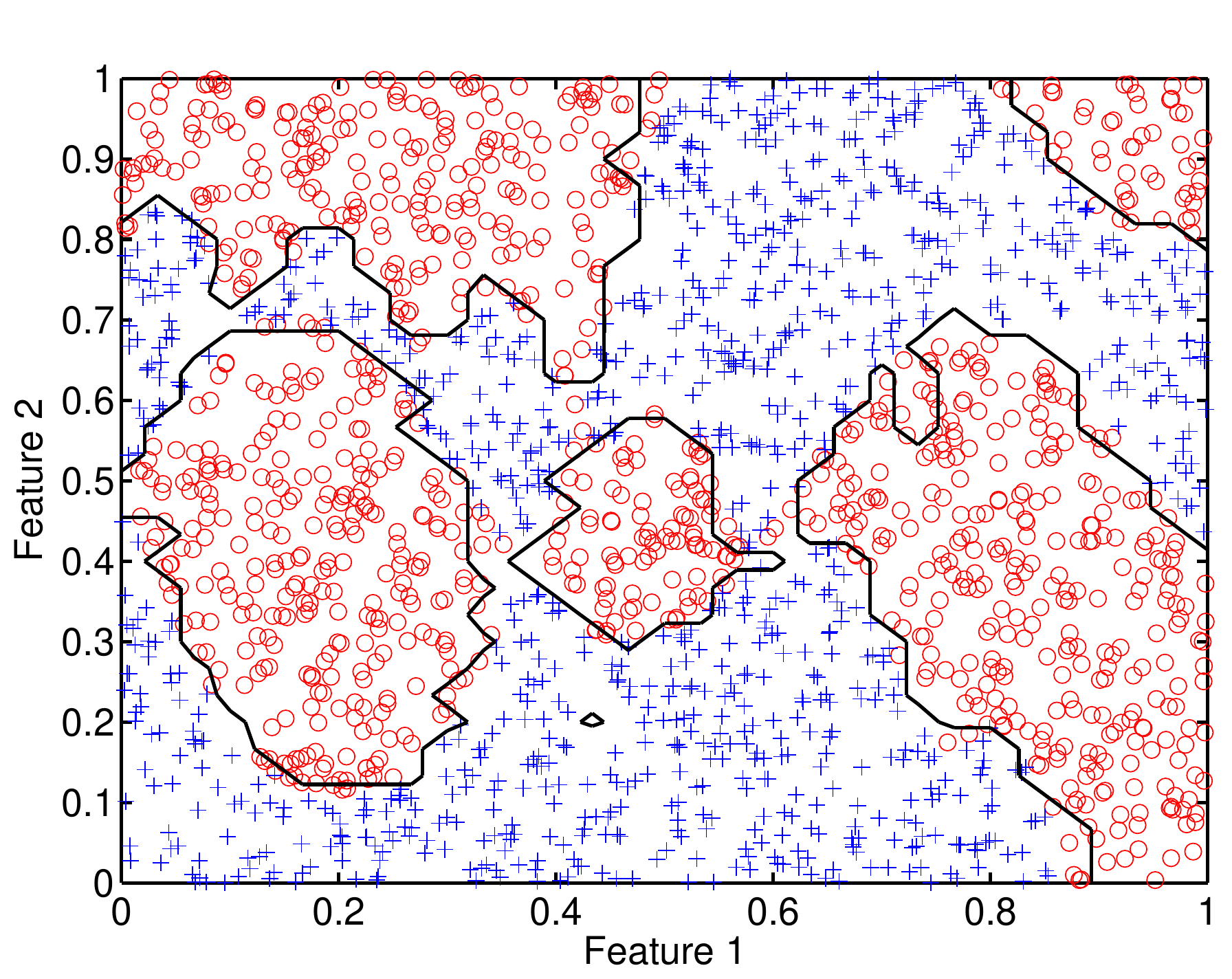}} 
	\subfigure[10 Stumps]{\includegraphics[width=3.2in,clip=]{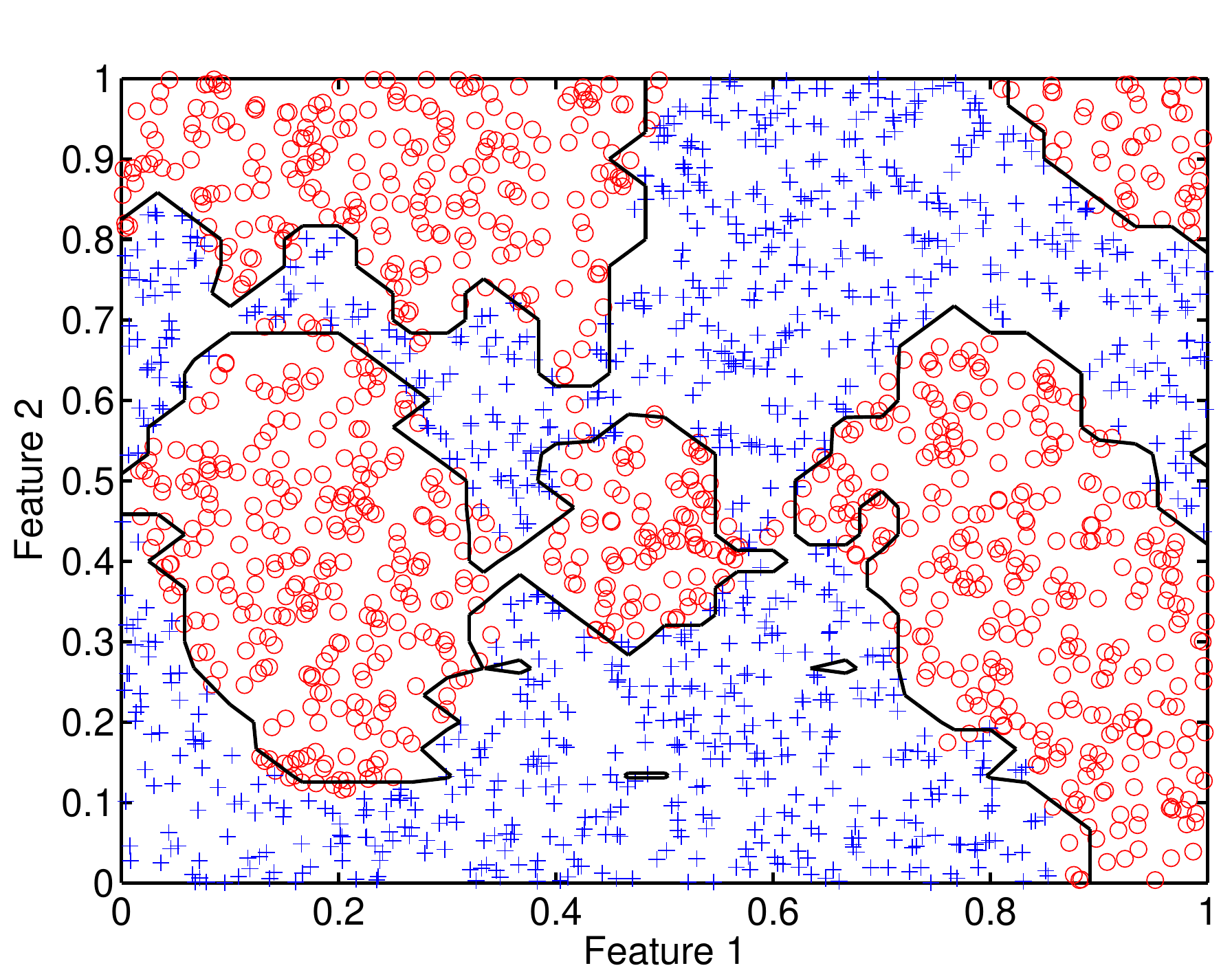}} 
	\subfigure[25 Stumps]{\includegraphics[width=3.2in,clip=]{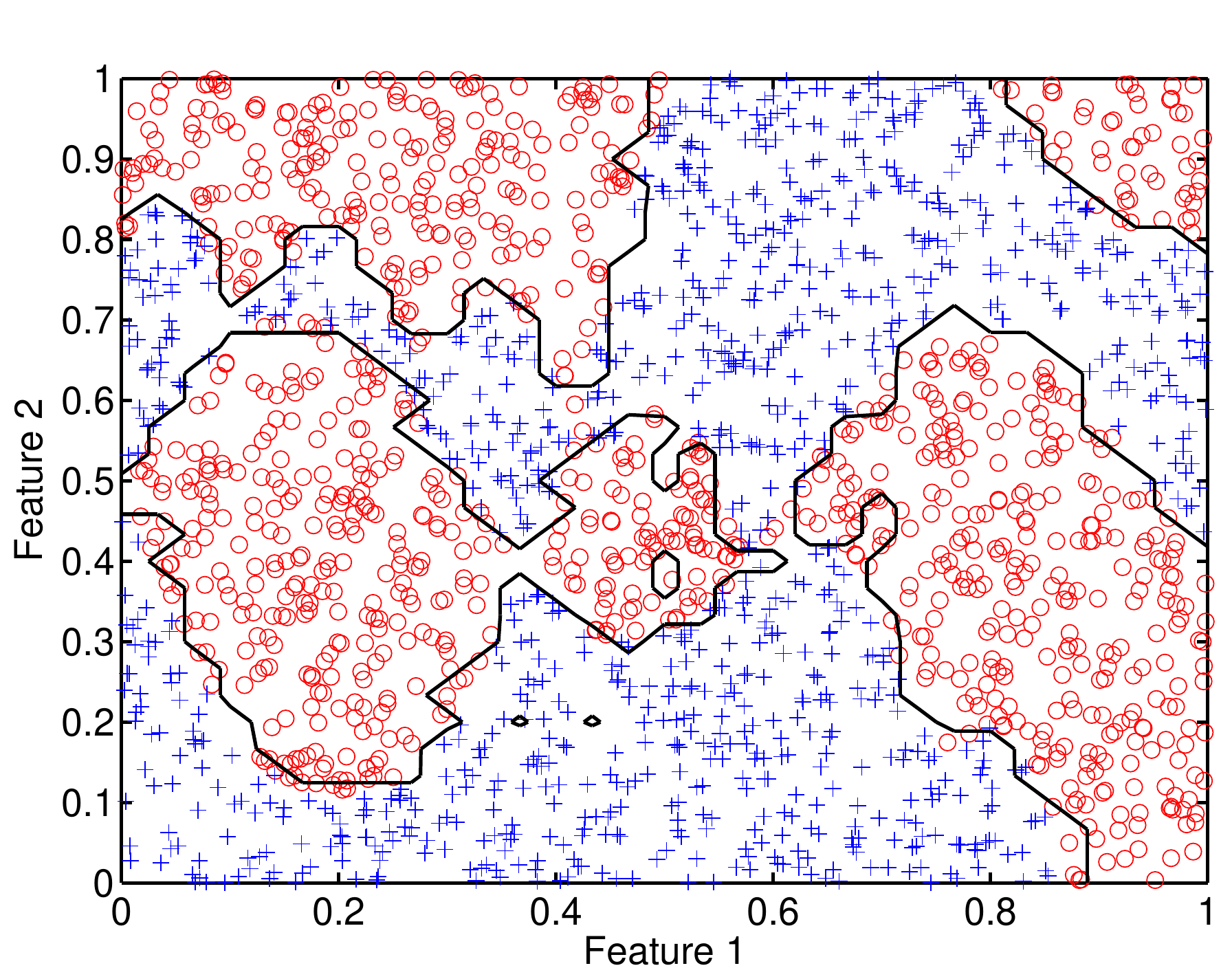}} 
	\subfigure[50 Stumps]{\includegraphics[width=3.2in,clip=]{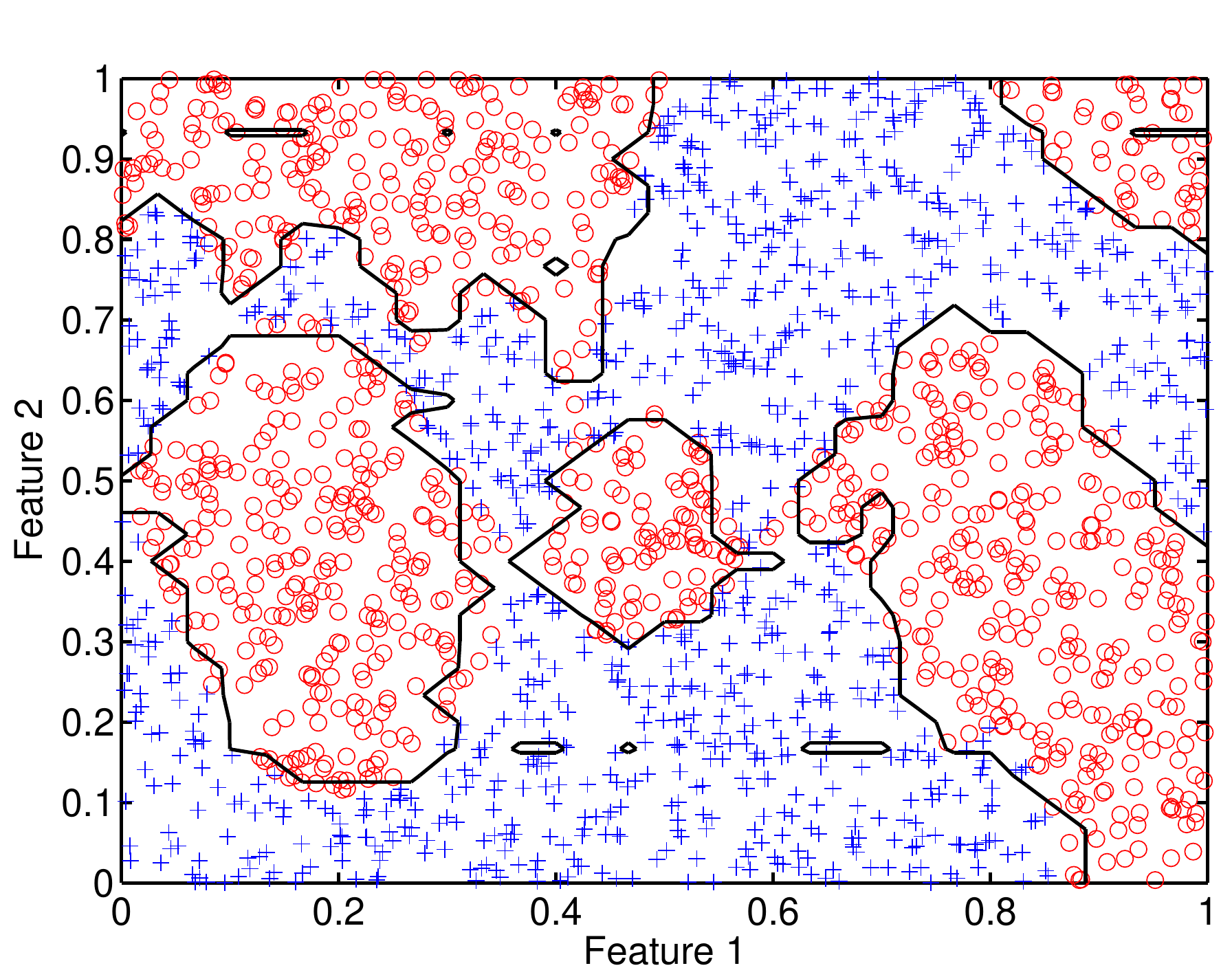}} 
	\subfigure[75 Stumps]{\includegraphics[width=3.2in,clip=]{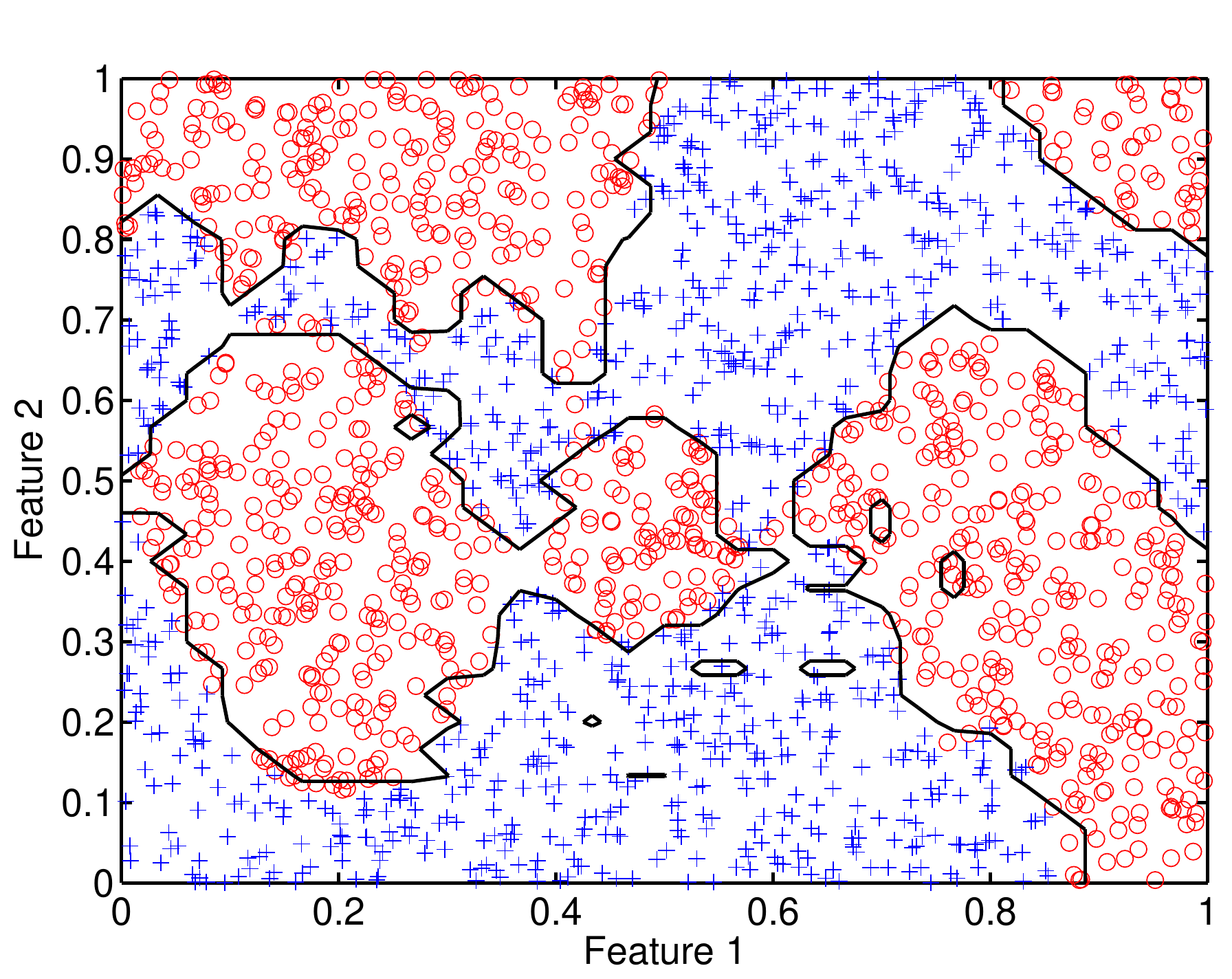}} 
	\subfigure[100 Stumps]{\includegraphics[width=3.2in,clip=]{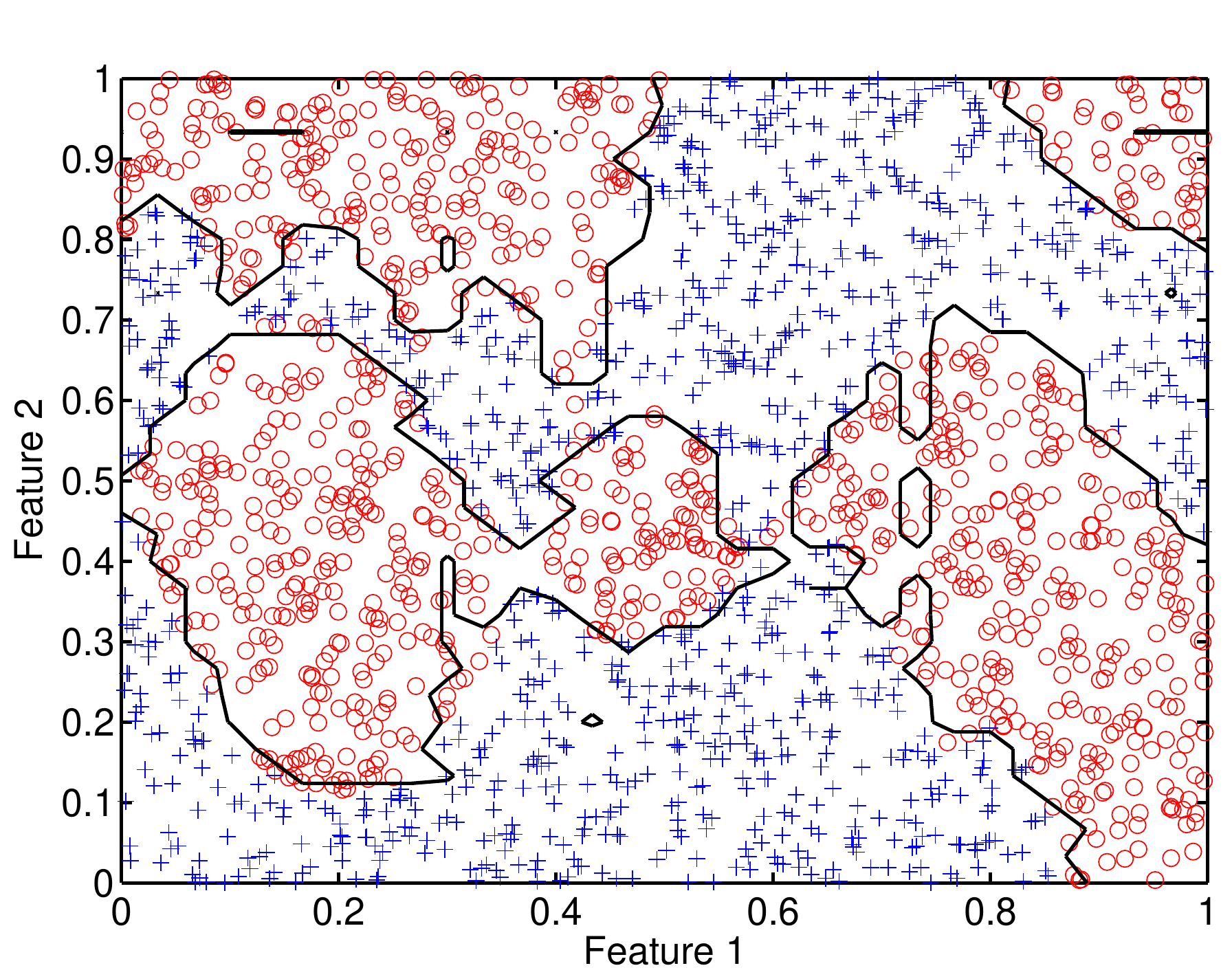}} 
	
	\caption{Decision boundaries generated by the META-DES framework for different pool size. Decision Stumps are used as base classifiers}
	\label{fig:resultsMETADESStumps}	  
\end{figure}

Figures~\ref{fig:resultsMETADES} and~\ref{fig:resultsMETADESStumps} illustrate the decision boundary obtained by the META-DES framework using Perceptron and Decision, respectively, stump as base classifier. We can see that when only 5 base classifiers are used, the decision boundary of the META-DES is close to the real decisions of the problem.

\subsection{The effect of the size of the dynamic selection dataset (DSEL)}
\label{sec:effectDSEL}

\begin{figure}[H]
   \begin{center}  	 
       	  \includegraphics[clip=, width=0.650\textwidth]{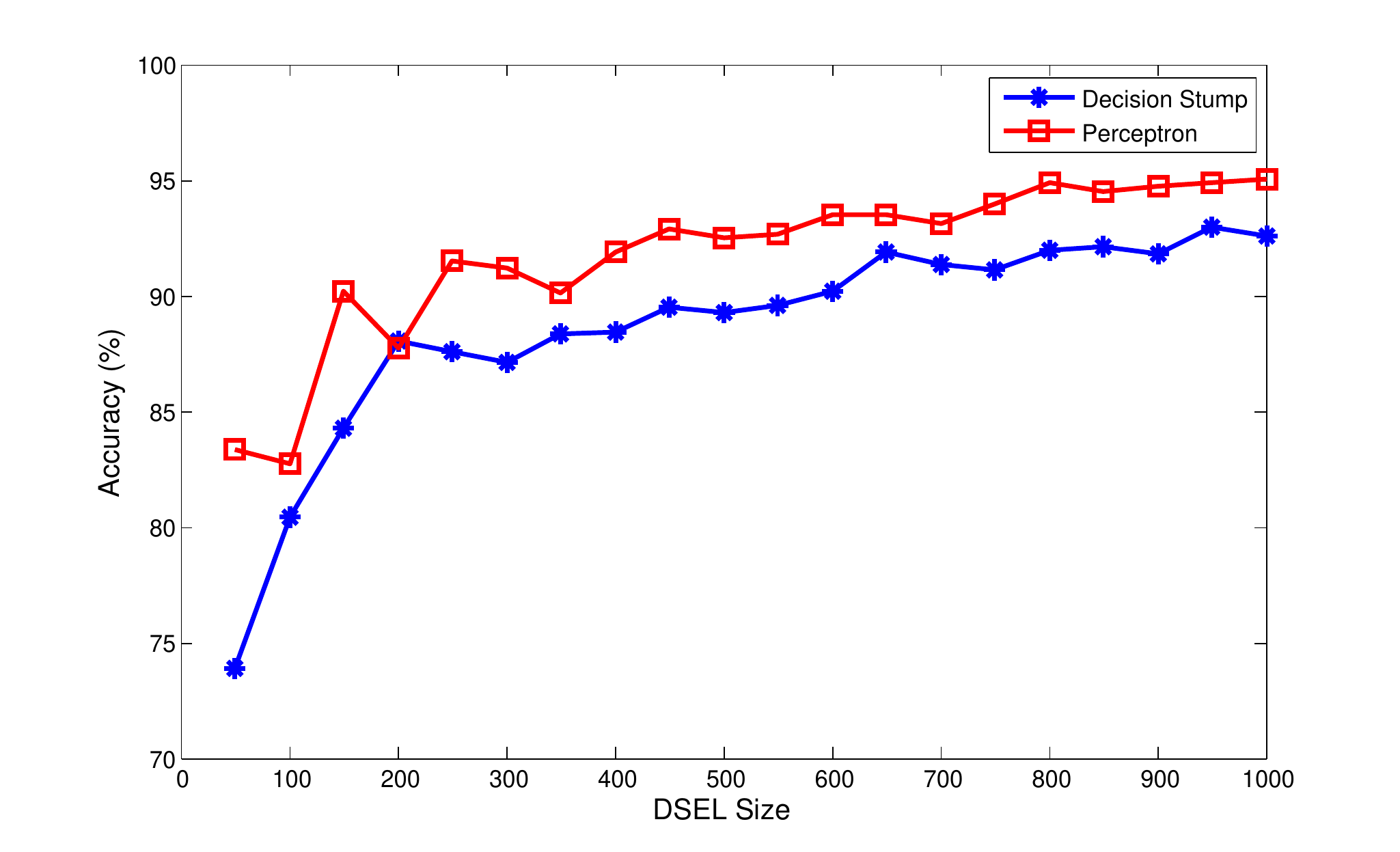}
   \end{center}
\caption{The effect of the DSEL size in the classification accuracy. Perceptron and Decision Stumps are considered as base classifiers. The results are obtained using a pool with 100 base classifiers, $M = 100$.}
\label{fig:2DValidation}	  
\end{figure}

Figure~\ref{fig:2DValidation} shows the performance of the META-DES using both Perceptron and Decision stumps according to the DSEL size. We varied the size of the dynamic selection dataset from 50 to 1000 at 50 point intervals (20 configurations were tested). The distribution varying the size of DSEL is presented in~\ref{sec:dselsize}. For this experiment, the size of the pool was set at $100$. We can observe that the size of the dynamic selection dataset has a greater influence on the classification result. This can be explained by the fact that the dynamic selection dataset, DSEL, is used in estimating the competence of the base classifiers, as shown in the classification example (Section~\ref{sec:classification}). With more samples in DSEL, the probability of selecting samples that are similar to the query sample both in the feature space or in the decision space for extracting the meta-features is higher. Hence, a better estimation of the competence of the base classifiers is achieved.

\begin{figure}[H]
   \begin{center}  	 
       	  \includegraphics[clip=, width=0.720\textwidth]{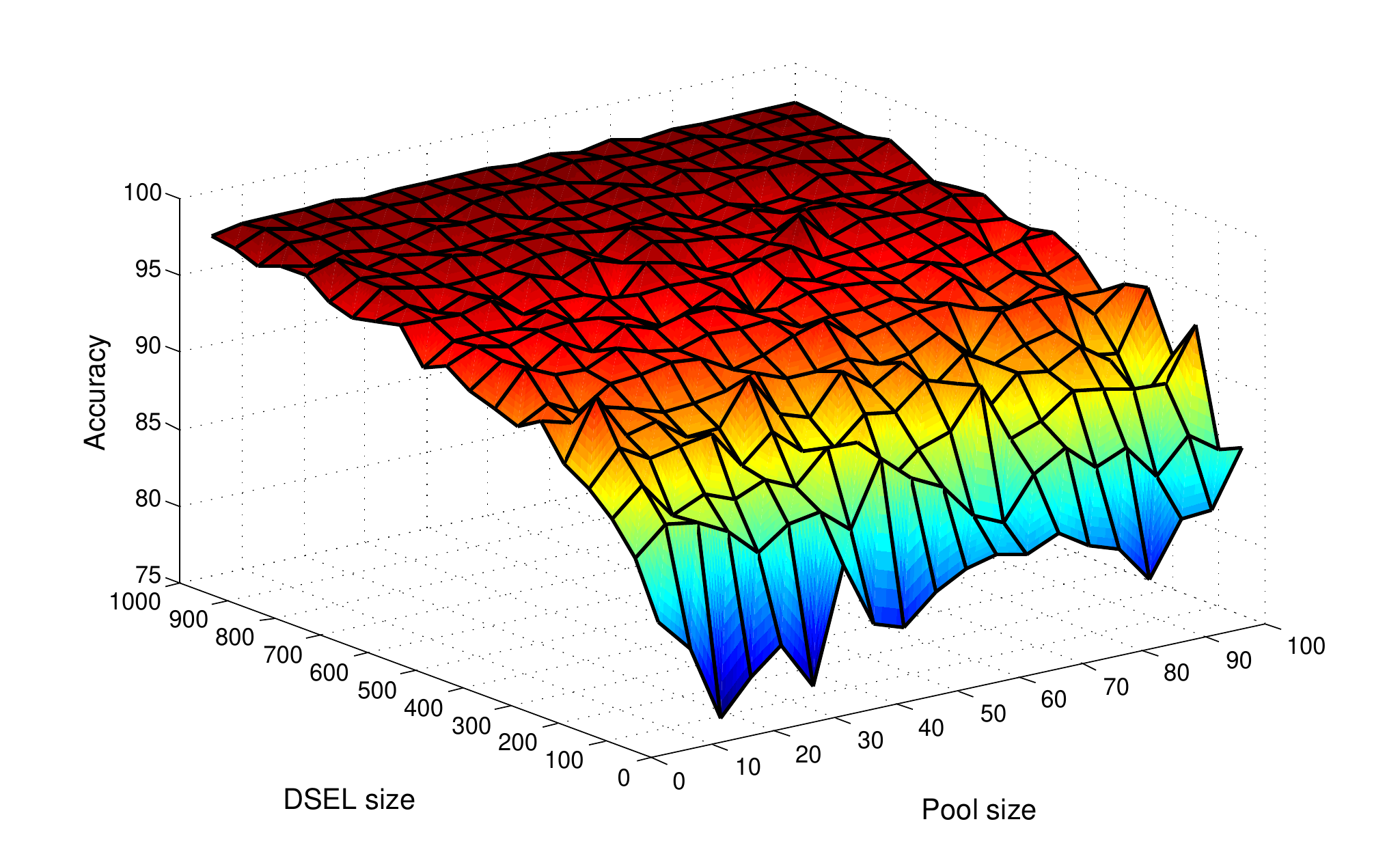}
   \end{center}
\caption{The effect of the pool size and the validation set size (DSEL) in the accuracy of the system. Perceptrons are used as base classifier.}
\label{fig:3DPerceptron}	  
\end{figure}

\begin{figure}[H]
   \begin{center}  	 
       	  \includegraphics[clip=, width=0.72\textwidth]{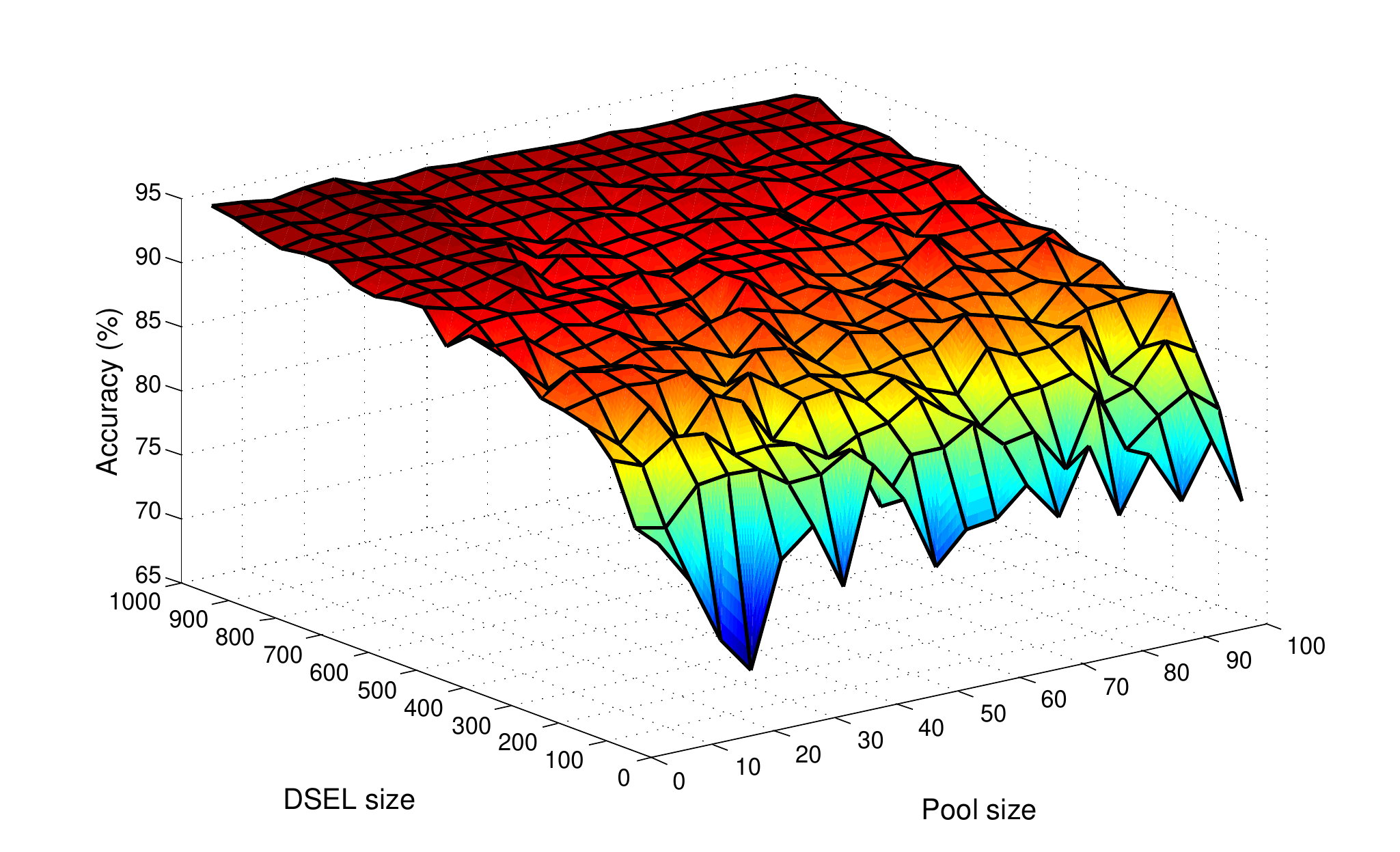}
   \end{center}
\caption{The effect of the pool size and the validation set size (DSEL) in the accuracy of the system. Decision Stumps are used as base classifier.}
\label{fig:3DStumps}	  
\end{figure}

Moreover, to better understand the influence of both the size of the pool and the size of the dynamic selection dataset together, we constructed a 3D mesh plot showing the accuracy of the system according to both parameters (Figures~\ref{fig:3DPerceptron} and~\ref{fig:3DStumps}).

\subsection{Results of static combination techniques}

\begin{figure}[H]
   \begin{center}  	 
       	  \includegraphics[clip=, width=0.70\textheight]{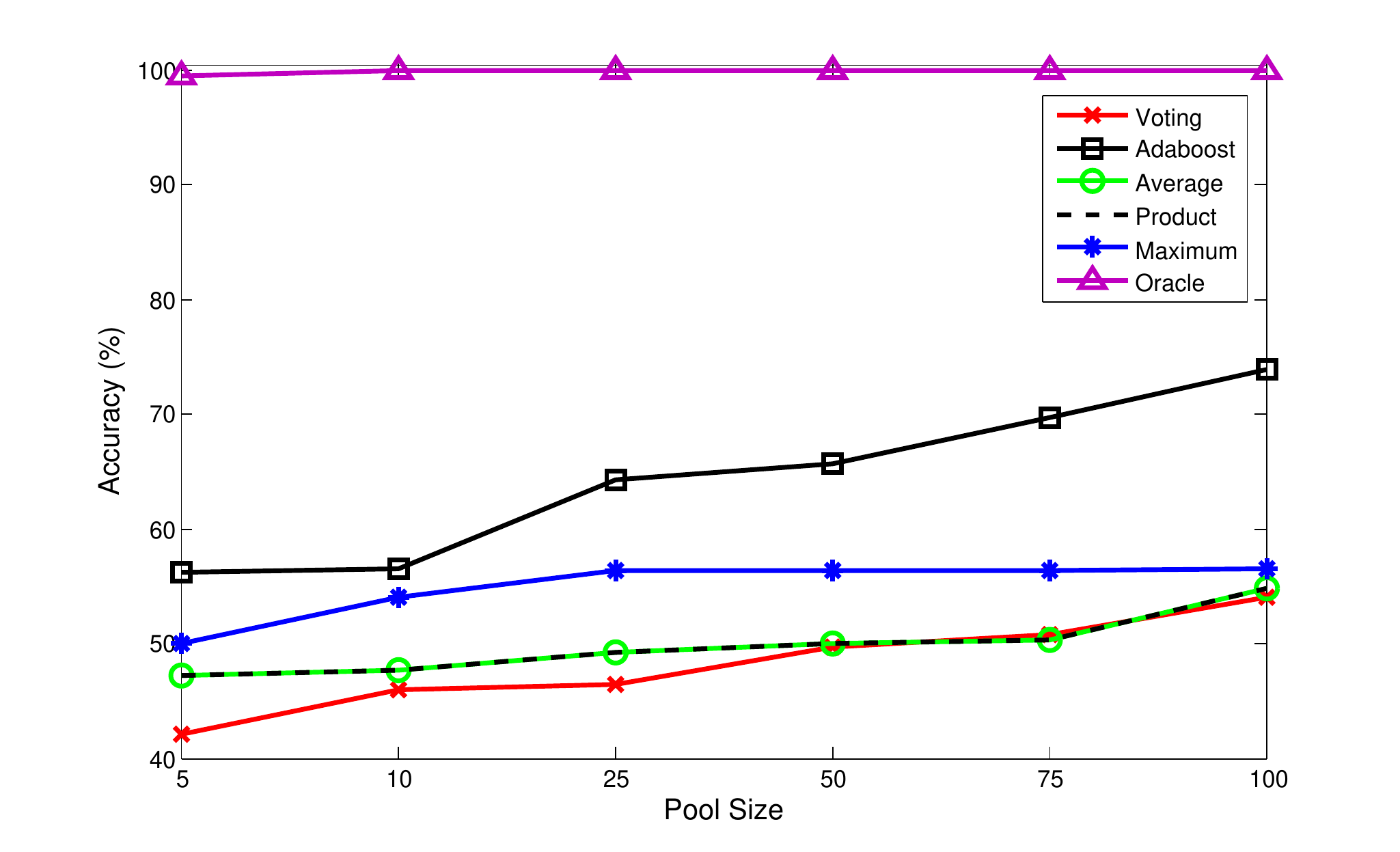}
   \end{center}
\caption{Results of static combination techniques using Perceptron as base classifier.}
\label{fig:StaticPerceptron}	  
\end{figure}

\begin{figure}[H]
   \begin{center}  	 
       	  \includegraphics[clip=,  width=0.70\textheight]{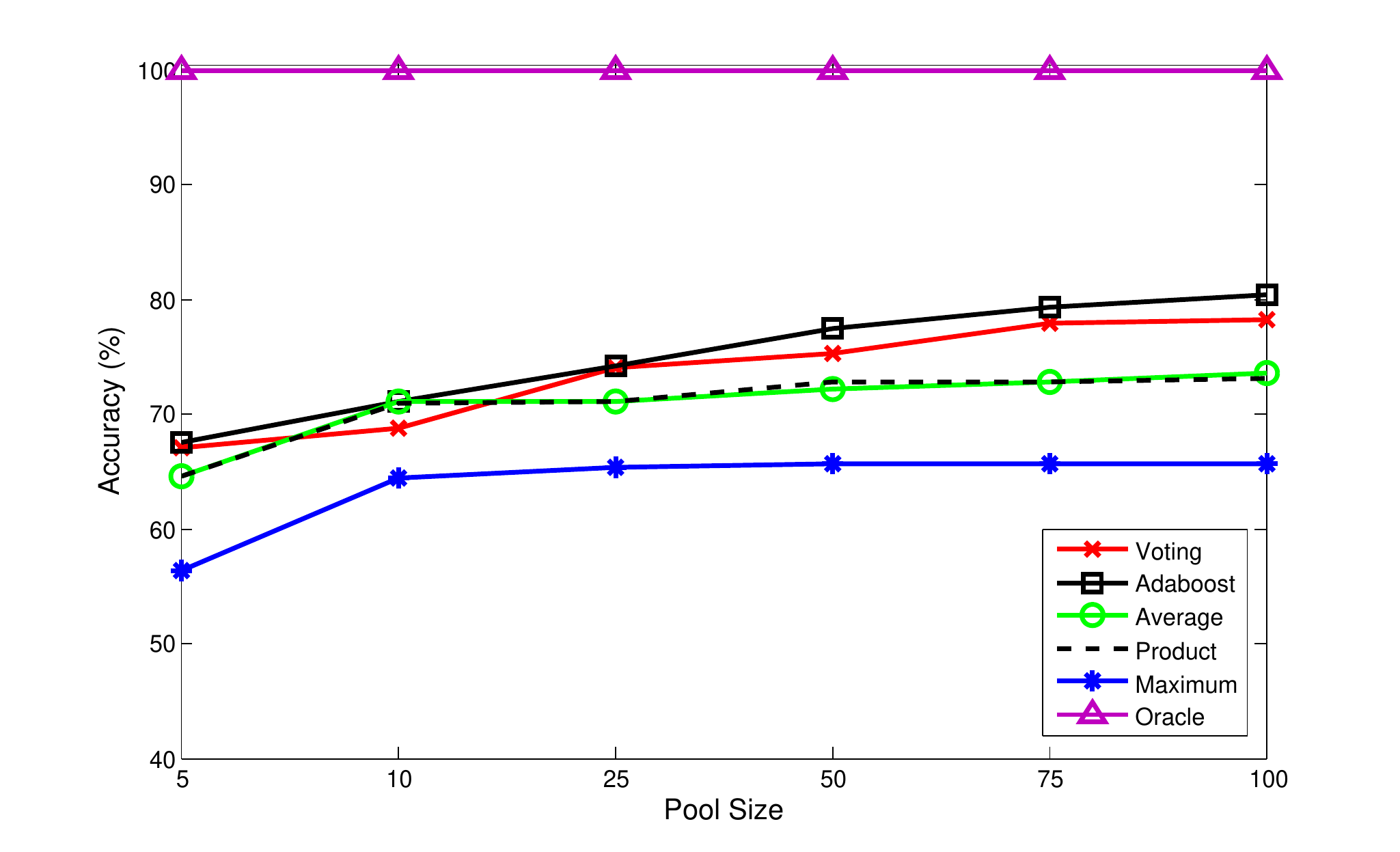}
   \end{center}
\caption{Results of static combination techniques using Decision Stumps as base classifier.}
\label{fig:StaticStumps}	  
\end{figure}

\begin{figure}[H]
	\centering
	\subfigure[Voting decision]{\includegraphics[width=3.2in,clip=]{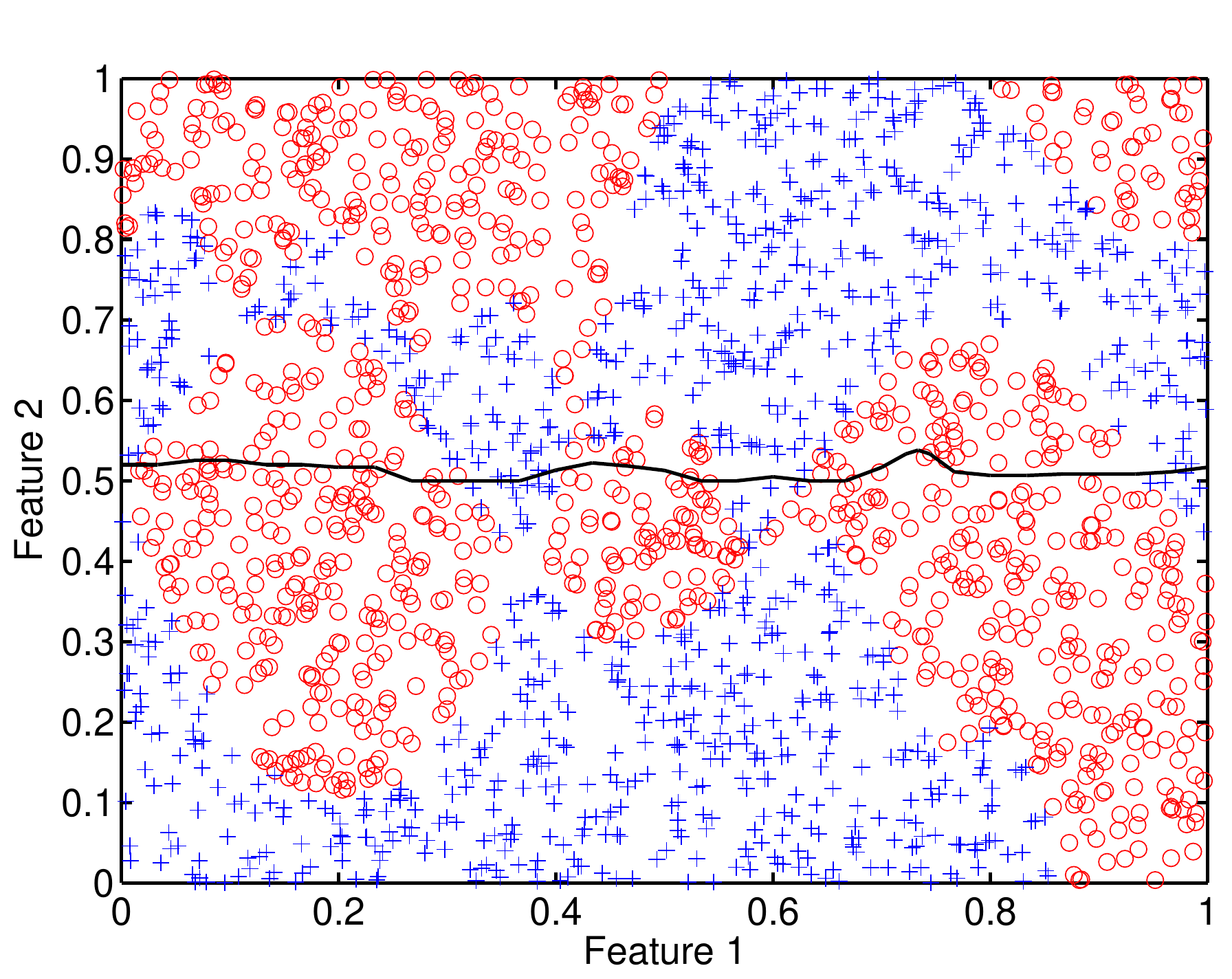}} 
	\subfigure[Averaging decision]{\includegraphics[width=3.2in,clip=]{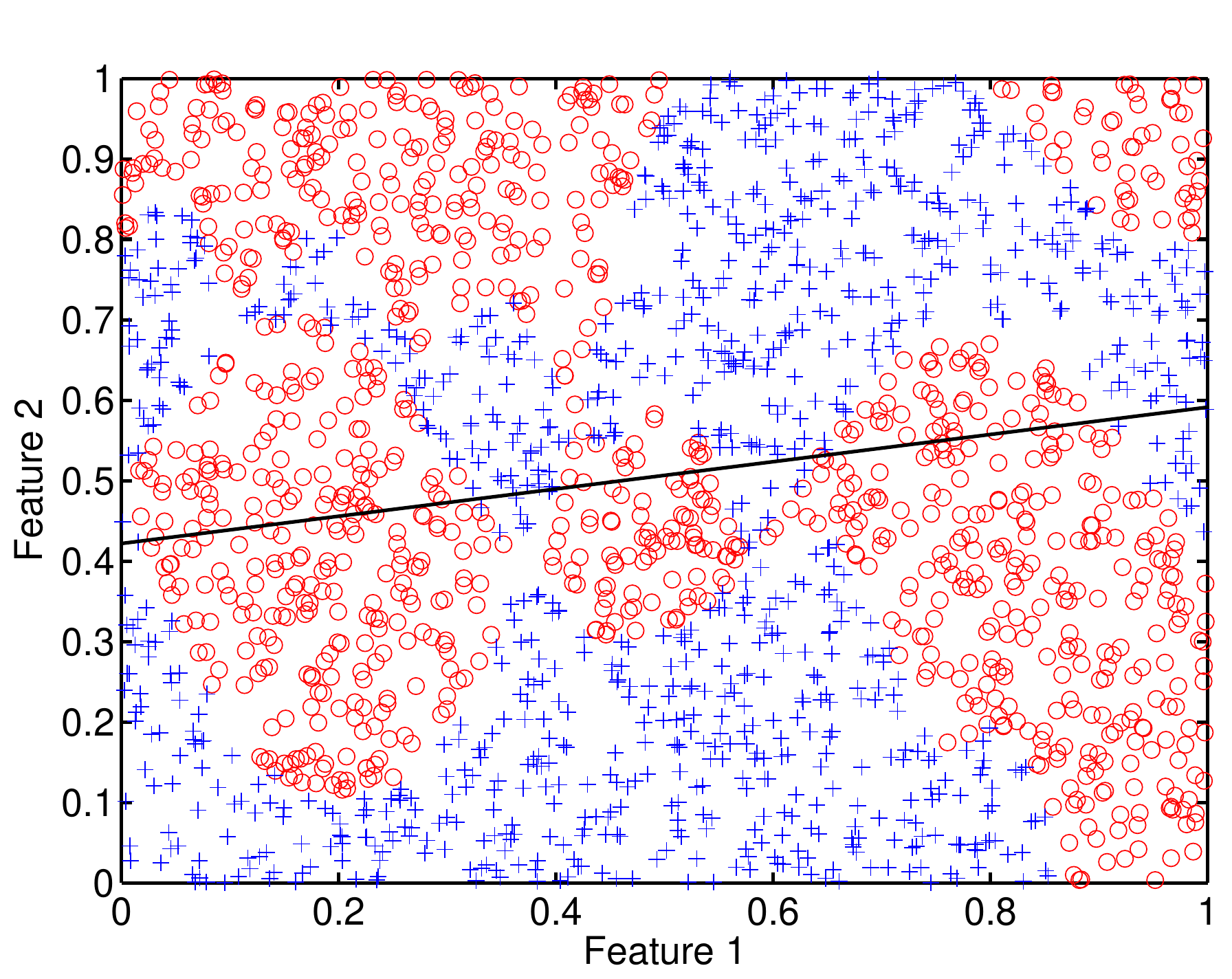}} 
	\subfigure[Maximum decision]{\includegraphics[width=3.2in,clip=]{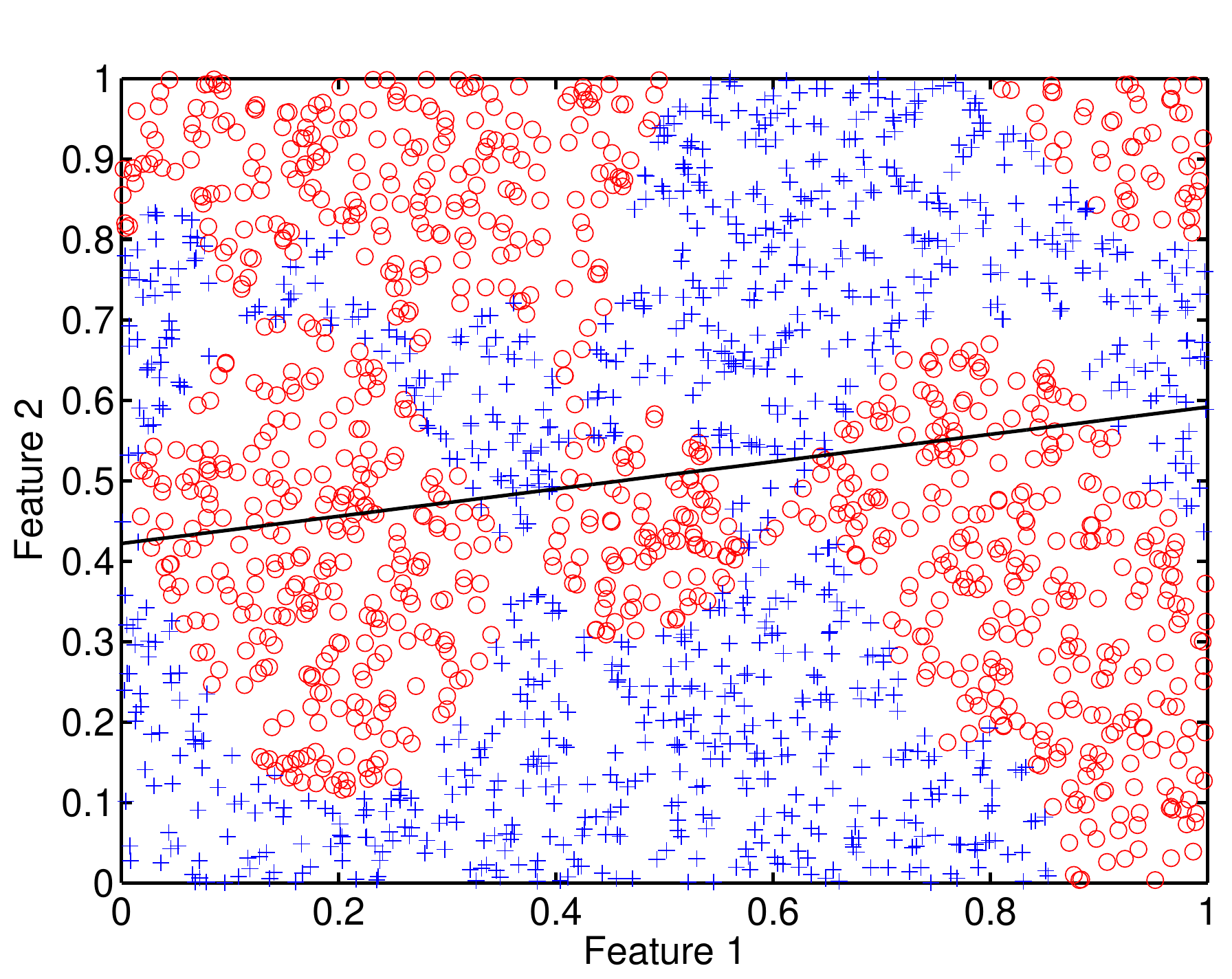}} 
	\subfigure[Product decision]{\includegraphics[width=3.2in,clip=]{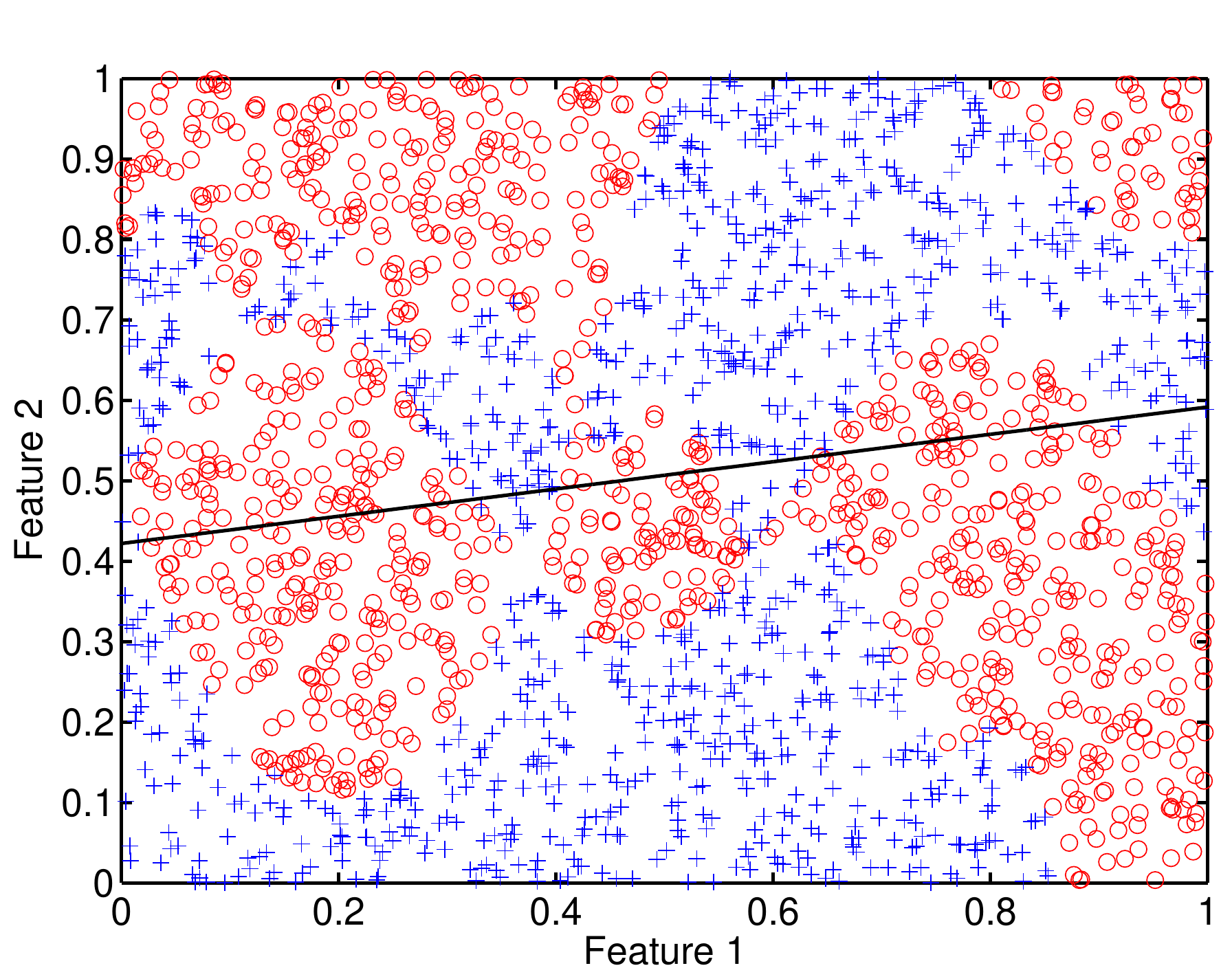}} 
	\subfigure[Adaboost decision]{\includegraphics[width=3.2in,clip=]{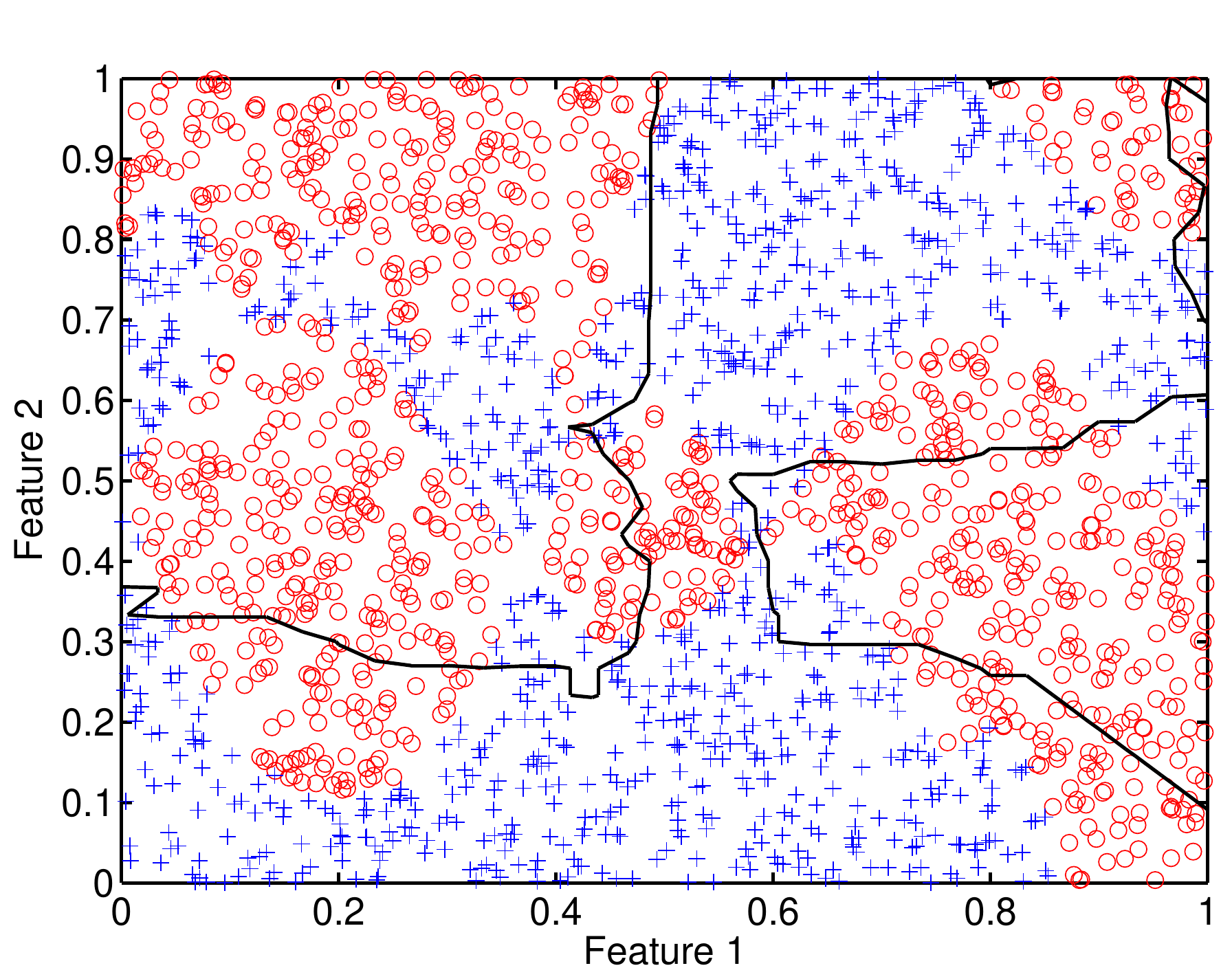}}
	 	
	\caption{Decision boundaries generated by each ensemble method. The pool of classifiers is composed of 100 Perceptron classifiers.}
	\label{fig:resultsP2StaticPerceptron100}	  
\end{figure}

\begin{figure}[H]
	\centering
	\subfigure[Voting decision]{\includegraphics[width=3.2in,clip=]{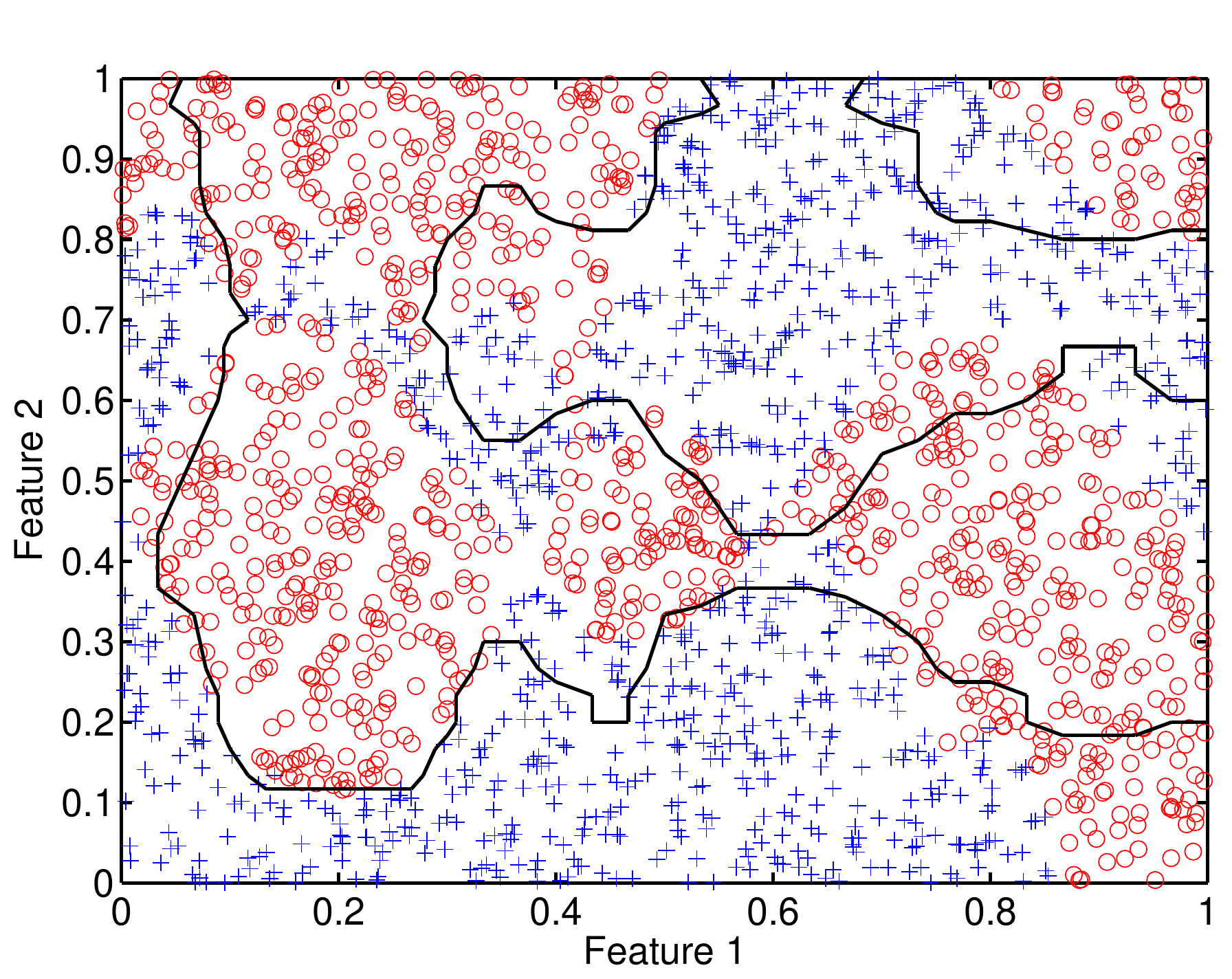}} 
	\subfigure[Averaging decision]{\includegraphics[width=3.2in,clip=]{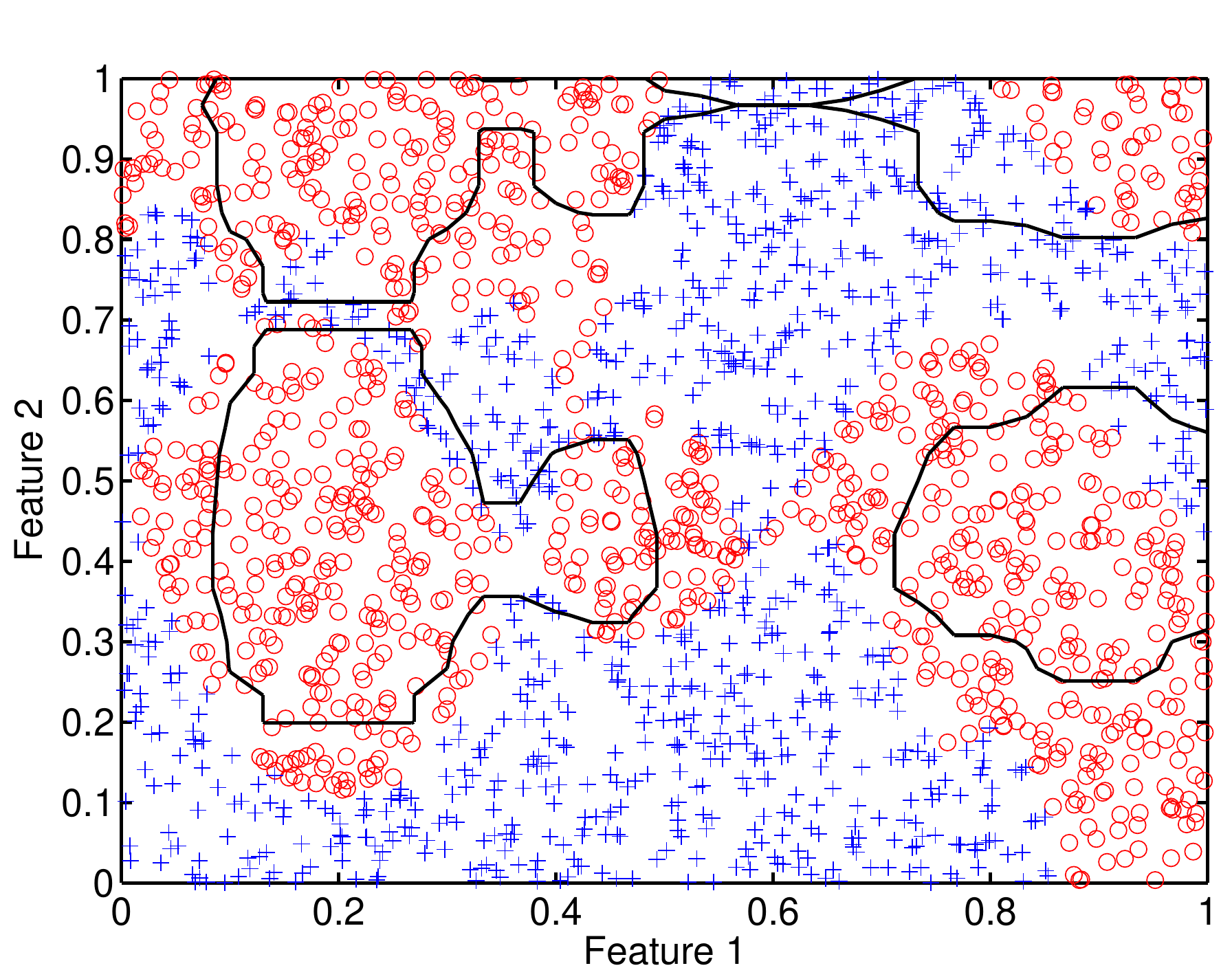}} 
	\subfigure[Maximum decision]{\includegraphics[width=3.2in,clip=]{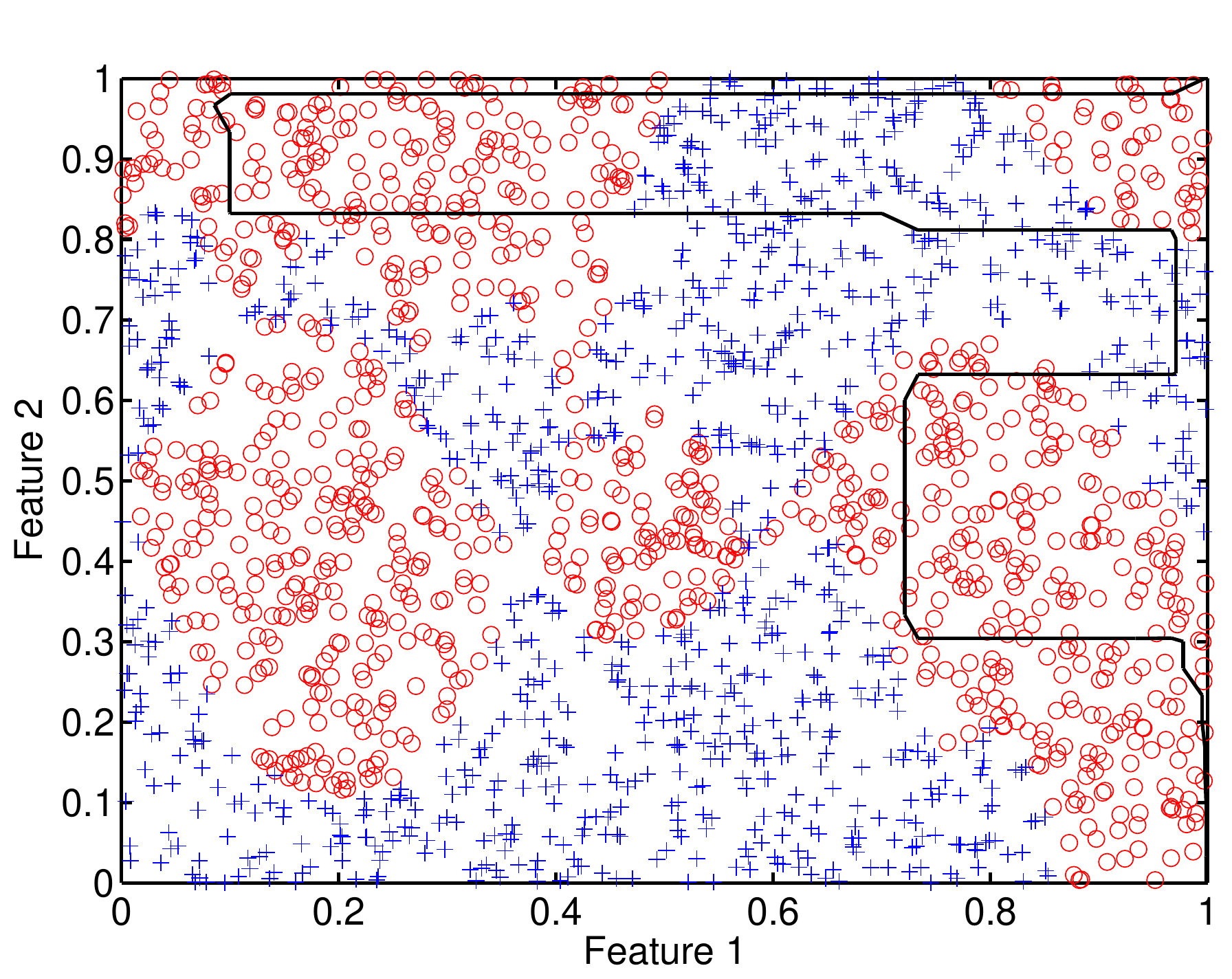}} 
	\subfigure[Product decision]{\includegraphics[width=3.2in,clip=]{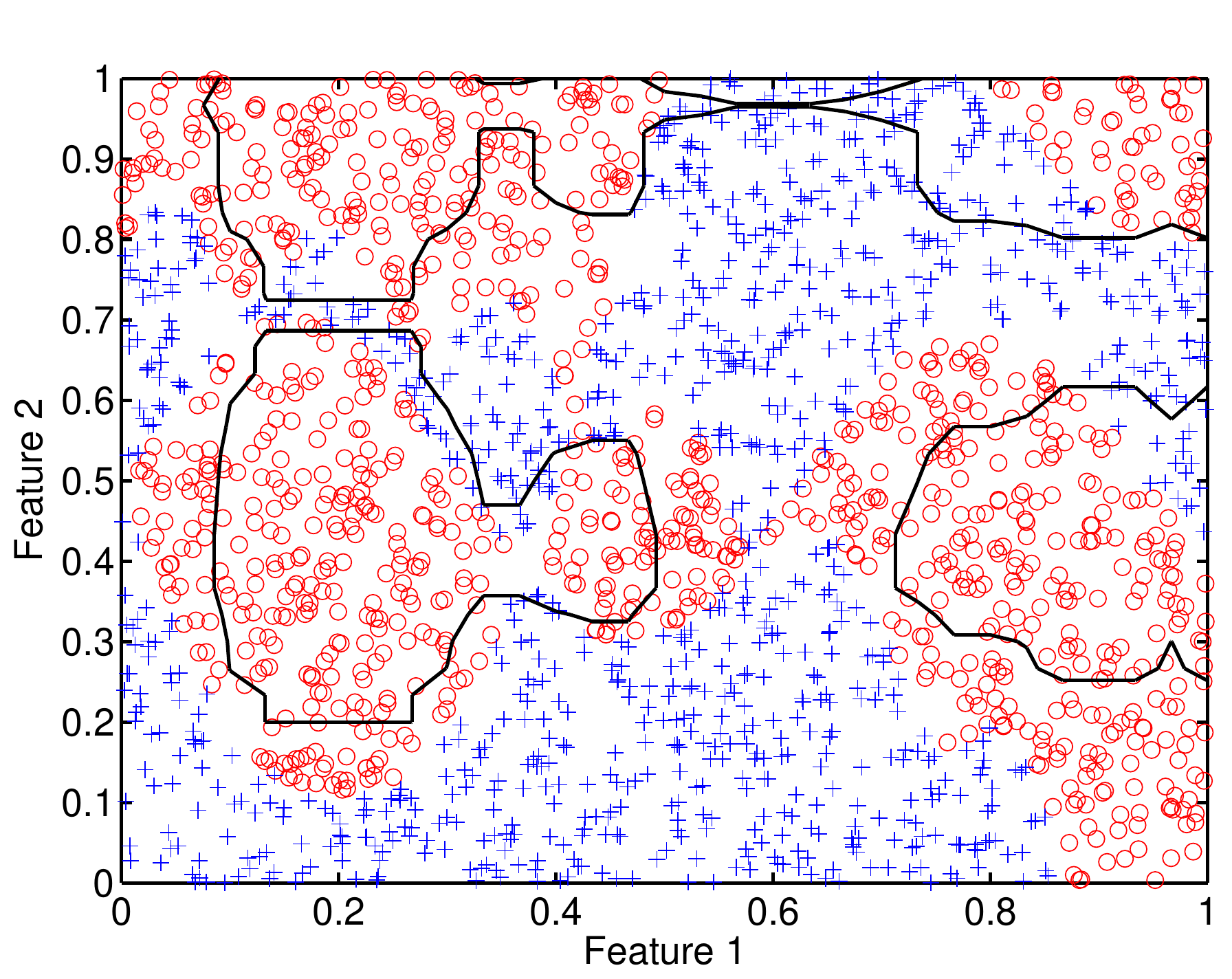}} 
	\subfigure[Adaboost decision]{\includegraphics[width=3.2in,clip=]{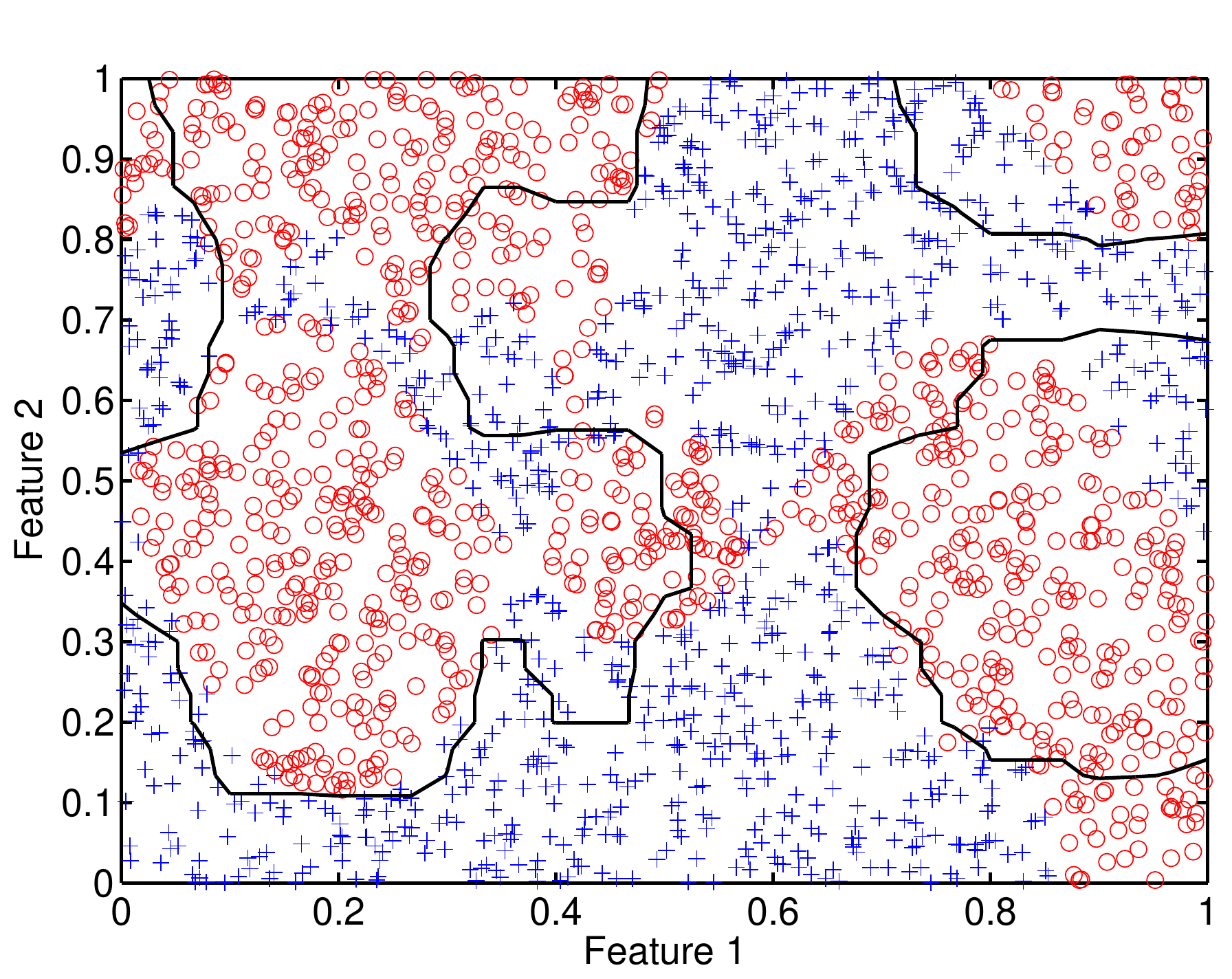}} 	
	\caption{Decision boundaries generated by each ensemble method. The pool of classifiers is composed of 100 Decision stumps classifiers.}
	\label{fig:resultsP2StaticDecisionStumps}	  
\end{figure}

Figures~\ref{fig:StaticPerceptron} and~\ref{fig:StaticStumps} illustrate the accuracy rates of static combination techniques by varying the size of the pool of classifiers. Furthermore, the decision boundaries for the static combination techniques are shown in Figures~\ref{fig:resultsP2StaticPerceptron100} and~\ref{fig:resultsP2StaticDecisionStumps} for Perceptrons and Decision Stumps, respectively.

Even when the size of the pool is increased to 100 base classifiers (Figure~\ref{fig:resultsP2StaticPerceptron100}), the static combination techniques cannot approximate the decision of the P2 problem. The performance using Decision Stumps as base classifiers is significantly better than that using Perceptrons for the static combination rules, especially considering the AdaBoost technique. This fact can be explained by the divide-and-conquer approach of decision stumps, in which each Stump is trained using a single feature. Hence, the classification task may become easier for the classifier model. However, the classification accuracy is still far from the performance obtained by the META-DES framework. Even using only 5 base classifiers, the performance of the META-DES is superior when compared to static combination techniques using up to 100 base classifiers.

%----------------------------------------------------------------------------------------------------------------------

%-----------------------------------PERCEPTRON---------------------------------------------------------------

%----------------------------------------------------------DSTUMP---------------------------------------------------------------

%----------------------------------------------------------------------------------------------------------------------

\subsection{Single classifier models}

In this section, we show the results of classical classification models for the P2 problem. We evaluate three classifier models: MLP Neural Network, Support Vector Machines with Gaussian Kernel (SVM) and Random Forest classifier. These classifiers were selected based on a recent study~\cite{delgado14a} that ranked the best classification models in a comparison considering a total of 179 classifiers over 121 classification datasets. All the classifiers were evaluated using the Matlab PRTOOLS toolbox~\cite{PRTools}. The parameters of each classifier were set as follows:

\begin{enumerate}

\item MLP Neural Network LM: The validation data was used to select the number of nodes in the hidden layer. We used a configuration with 100 neurons in the hidden layer. The training process was performed using the Levenberg-Marquadt algorithm. The training process was stopped if its performance on the validation set decreased or failed to improve for five consecutive epochs.

\item MLP Neural Network RPROP: The validation data was used to select the number of nodes in the hidden layer. We used a configuration with 100 neurons in the hidden layer. The training process was performed using the Resilient Backpropagation algorithm~\cite{rprop} since this algorithm presented both a faster convergence and better classification performance in many applications~\cite{Cruz2012}. The training process was stopped if its performance on the validation set decreased or failed to improve for five consecutive epochs.

\item SVM: A radial basis SVM with a Gaussian Kernel was used. A grid search was performed in order to set the values of the regularization parameter $c$ and the Kernel spread parameter $\gamma$. 

\item Random Forest: We vary the number of trees from 25 to 200 at 25 point intervals. The configuration with the highest performance on the validation dataset is used for generalization. Since there are only two features in the P2 problem, a decision stump is used (depth = 1). 

\end{enumerate}

\begin{figure}[!ht]
	\centering
	\subfigure[Levemberg-Marquadt NN]{\includegraphics[width=3.2in,clip=]{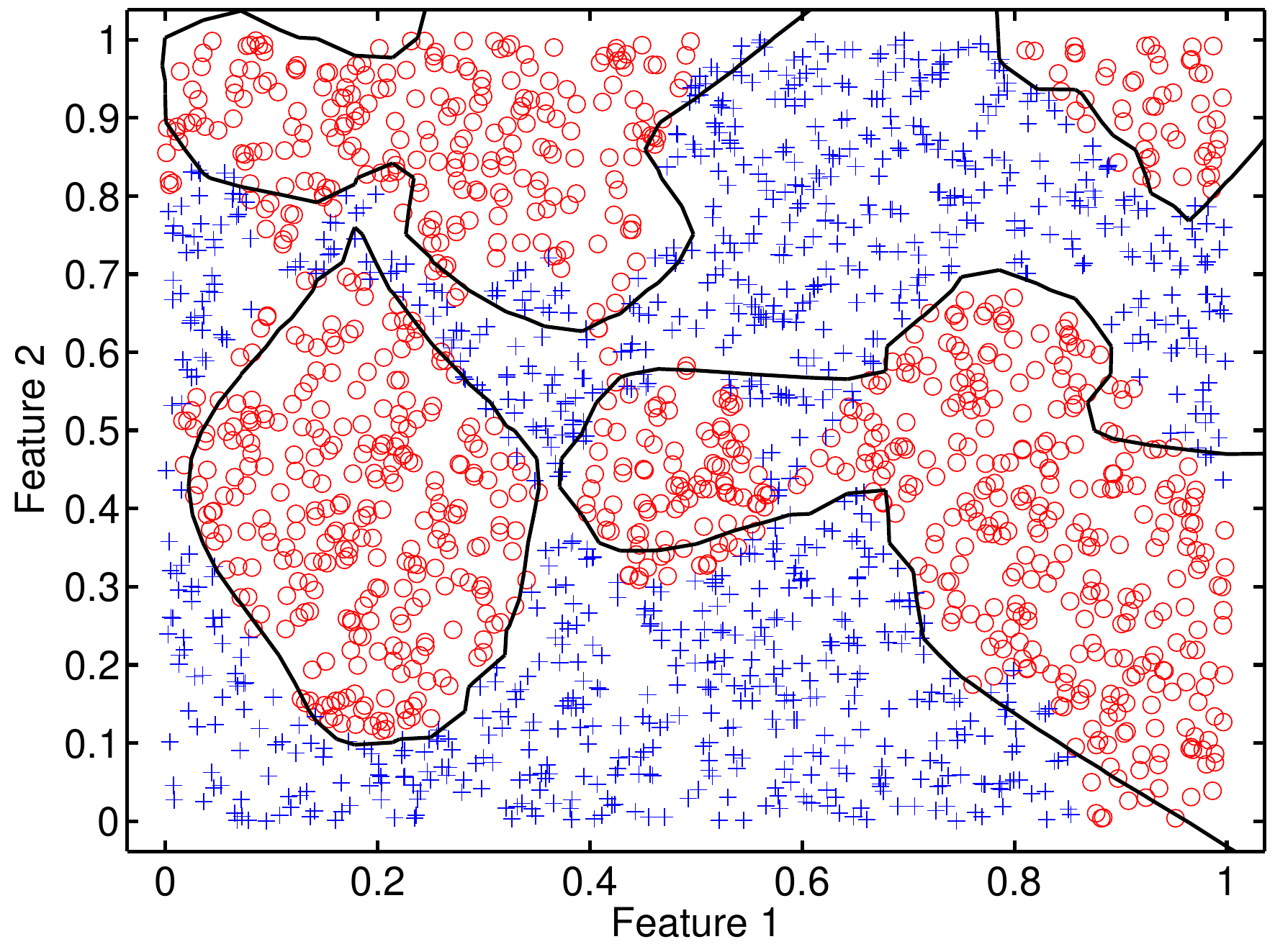}} 
	\subfigure[Resilent Backpropagation NN]{\includegraphics[width=3.2in,clip=]{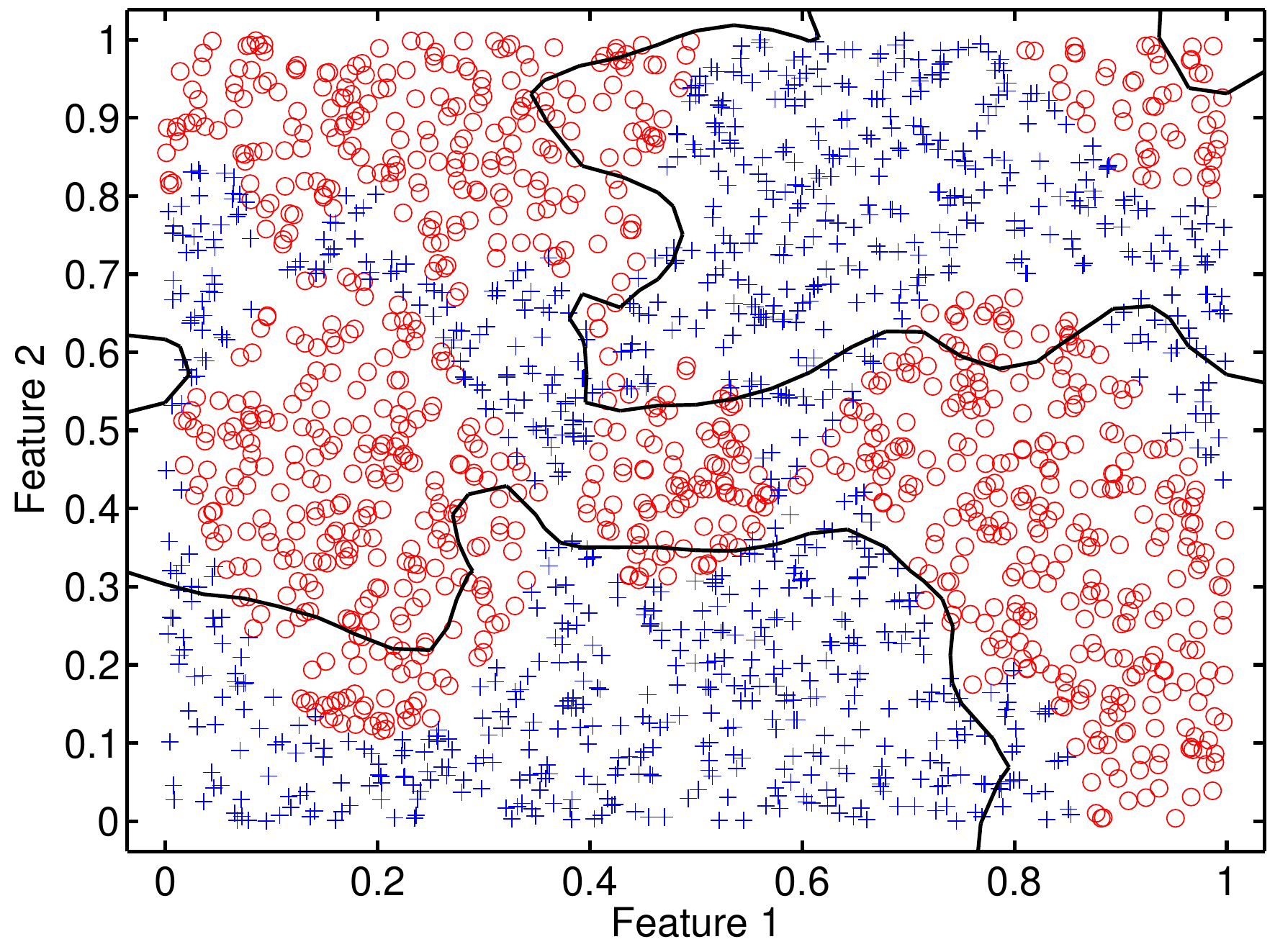}} 
	\subfigure[Random Forest]{\includegraphics[width=3.2in,clip=]{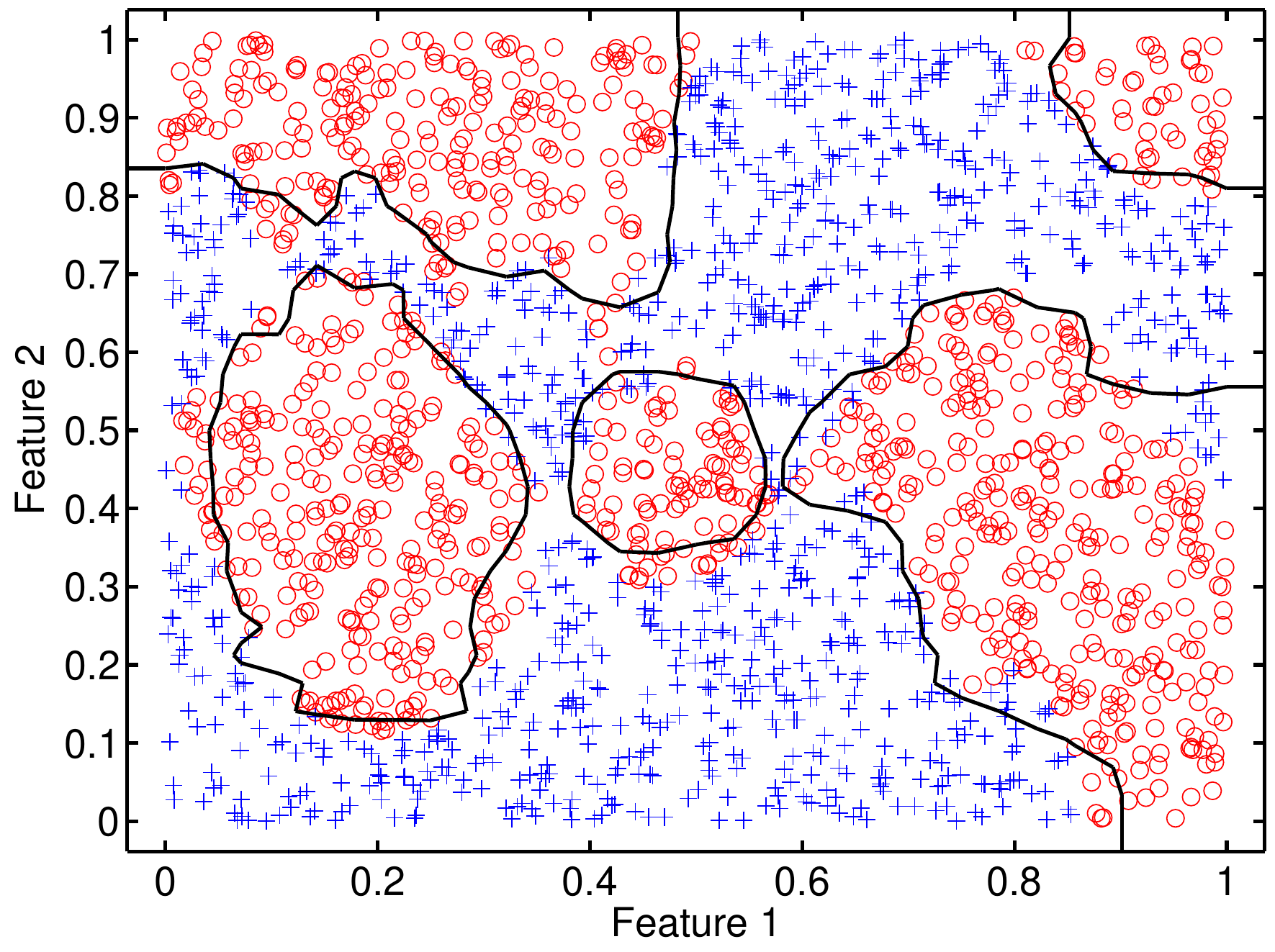}} 
	\subfigure[Radial Basis SVM]{\includegraphics[width=3.2in,clip=]{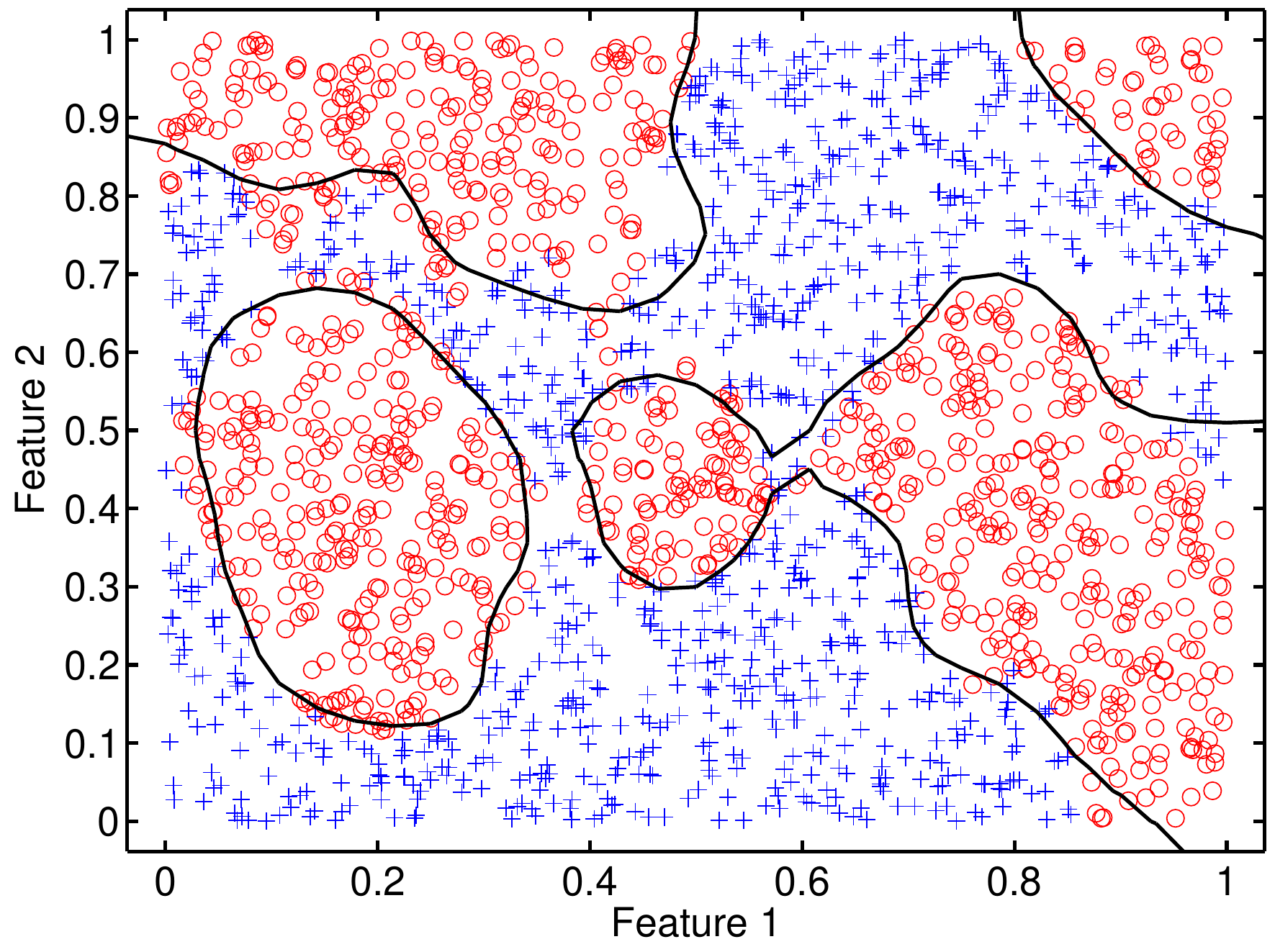}} 
	\caption{Decision boundaries obtained using a single classifier. (a) MLP-NN with 100 neurons in the hidden layer trained using Levemberg-Marquadt ($90\%$ accuracy). (b) MLP-NN with 100 neurons in the hidden layer trained using Resilient Backpropagation ($77\%$ accuracy). (c) Random Forest classifier ($91\%$ accuracy). Support Vector Machine with a Gaussian kernel (d) ($93\%$ accuracy).}
	\label{fig:decisiontClassifiers}	  
\end{figure}

Since these classifiers do not require a meta-training stage, in these experiments, we merge the training ($\mathcal{T}$) and meta-training set ($\mathcal{T}_{\lambda}$) into a single training set, thereby training the classifiers with 1000 samples. The samples in the dynamic selection dataset (DSEL) are used for the validation dataset. The decision boundary obtained by each classifier is presented in Figure~\ref{fig:decisiontClassifiers}. The MLP neural network trained with Levemberg-Marquadt obtained a recognition accuracy of $90\%$, while that trained with Resilient Backpropagation algorithm obtained $77\%$. The SVM obtained a recognition accuracy of $93\%$, and the random forest classifier achieved $91\%$. The classification accuracy of these single classifier models is lower than the performance of the META-DES using a pool of either five Perceptrons or five Decision Stumps. This result can be explained by the complex nature of the P2 problem. It is difficult to properly train a strong classifier to learn the separation between the two classes. These classifiers might require more training samples in order to obtain better generalization performance. %1000 samples for training and 500 for validation is not enough for the training of a strong single classifiers. 

%They need more training samples to obtain a better approximation of the P2 problem decision.

 %(**This experiment may be biased towards the SVM and random forest classifiers since a grid optimization procedure was performed to find the best values for the hyper-parameters. We did not use any optimization at all for the META-DES framework**).

\section{Conclusion}
\label{sec:conclusion}

In this work, we perform a DEEP analysis of the META-DES framework using linear classifiers. The analysis is conducted using the P2 problem, which is a complex non-linear problem with two classes having multiple class centers. We demonstrate that using the META-DES framework, we can approximate the complex non-linear distribution of the P2 problem using few linear classifiers. The accuracy rate provided by the best linear classifiers trained for this problem is around 50\%. We demonstrate that using static combination techniques, it is impossible to approximate the complex decision frontier of the P2 problem. Because of the complex nature of the P2 problem, for every test sample, there is high disagreement between the predictions made by the base classifier. Since there is no consensus regarding the correct label for the test sample, the static combination techniques end up making random decisions. Even using techniques that assign weights to the base classifiers, such as AdaBoost, the classification accuracy using 100 base classifiers is still very different from the performance of the META-DES framework. Classifiers that are not experts in the local region where the query instance is located end up negatively influencing the decision of the system. Using dynamic selection, the decisions of the base classifiers that are not experts for the given query sample are not taken into account. Only the most competent classifiers are selected to the predict the label of the query sample.

The size of the pool of classifiers did not have a significant influence on the recognition rate. This finding can be explained by the fact that using only 5 base classifiers, the Oracle performance of the Pool is at 100\%. In other words, there is at least one base classifier that predicts the correct class label for every testing sample. The crucial element here is the criteria used to estimate the level of competence of the base classifiers in order to always select those that predict the correct class label for a given test sample. Moreover, we noticed a performance drop when using decision stumps as base classifiers when more than 25 base classifiers are used. These results indicate that increasing the number of base classifiers in the pool does not always lead to greater classification accuracy. Thus, one aspect of the framework that must be further investigated is how many base classifiers should be trained in the overproduction phase for a given classification problem. 

We evaluate the impact of the pool of classifiers and the size of the dynamic selection dataset (DSEL) that is used in dynamically estimating the level of competence of the base classifier. Experimental results show that the size of the dynamic selection dataset has a higher impact on classification performance. This can be explained by the fact the majority of the meta-features proposed for the META-DES framework are extracted from instances in DSEL that are similar to the query sample, considering both the feature space and the decision space. With more samples in the DSEL, the probability of selecting samples that are similar to the query sample in both the feature space and in the decision space for extracting the meta-features is higher. Hence, a better estimation of the competence of the base classifiers is achieved. The results found in this analysis should be considered as a guideline for future work on the META-DES and for other dynamic ensemble selection based on local accuracy information in general.

Furthermore, the META-DES framework presented a higher classification accuracy for the P2 problem than did the classical single classifier model. This finding may be attributed to the complex nature of the P2 problem, since a classifier such as an SVM or an MLP neural network may require more training samples for a better generalization performance. Using dynamic selection through the META-DES framework we can approximate the complex decision of the P2 problem using less training data.

It is important to mention that there is still room for improvement in the META-DES framework. Using five base classifiers, the accuracy rate obtained by the META-DES is around $95\%$, while the Oracle performance is close to $100\%$. Future works will involve the definition of new meta-features in order to achieve a behavior that is closer to the ideal dynamic selection technique (Oracle).

\appendix 

\subsection{Plotting decision boundaries}
\label{sec:decisionboundary}

When dealing with dynamic classifier or ensemble selection, for each classification sample $\mathbf{x}_{j,test}$, a specific ensemble or base classifier is selected to perform the classification. Thus, a grid is generated over the 2D image. The grid is generated in the same interval as the P2 classification problem [0, 1] for both axes. Each point on the 2D grid is passed down to the dynamic selection technique in order to predict its label. After every point on the 2D grid is evaluated, the MATLAB contour plot is used to separate the points that were classified between the two classes. It is important to mention that the number of points on the grid influences the definition of the decision boundary. A high number of points in the grid leads to a more precise decision boundary. In our experiment, we use a $100 \times 100$ grid, for a total of 10,000 points, in order to have a more precise decision boundary map. For the static combination rules and classification models the decision boundaries are plotted using the $plotc$ function from the PRTOOLS Matlab Toolbox~\cite{PRTools}.

\subsection{Ensemble Generation}

\begin{figure}[!ht]

	\centering
	\subfigure[5 Perceptrons]{\includegraphics[width=3.2in,clip=]{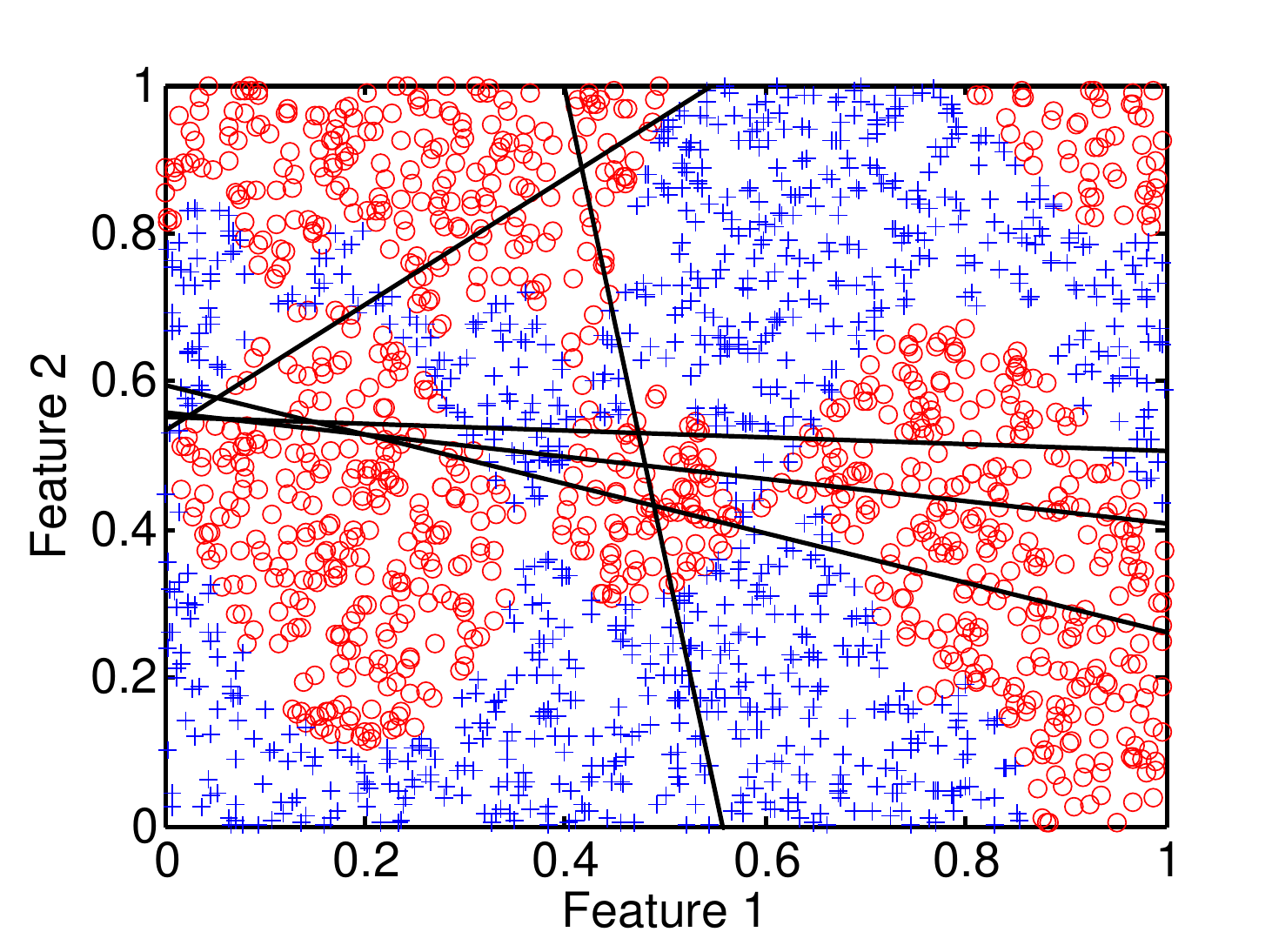}} 
	\subfigure[10 Perceptrons]{\includegraphics[width=3.2in,clip=]{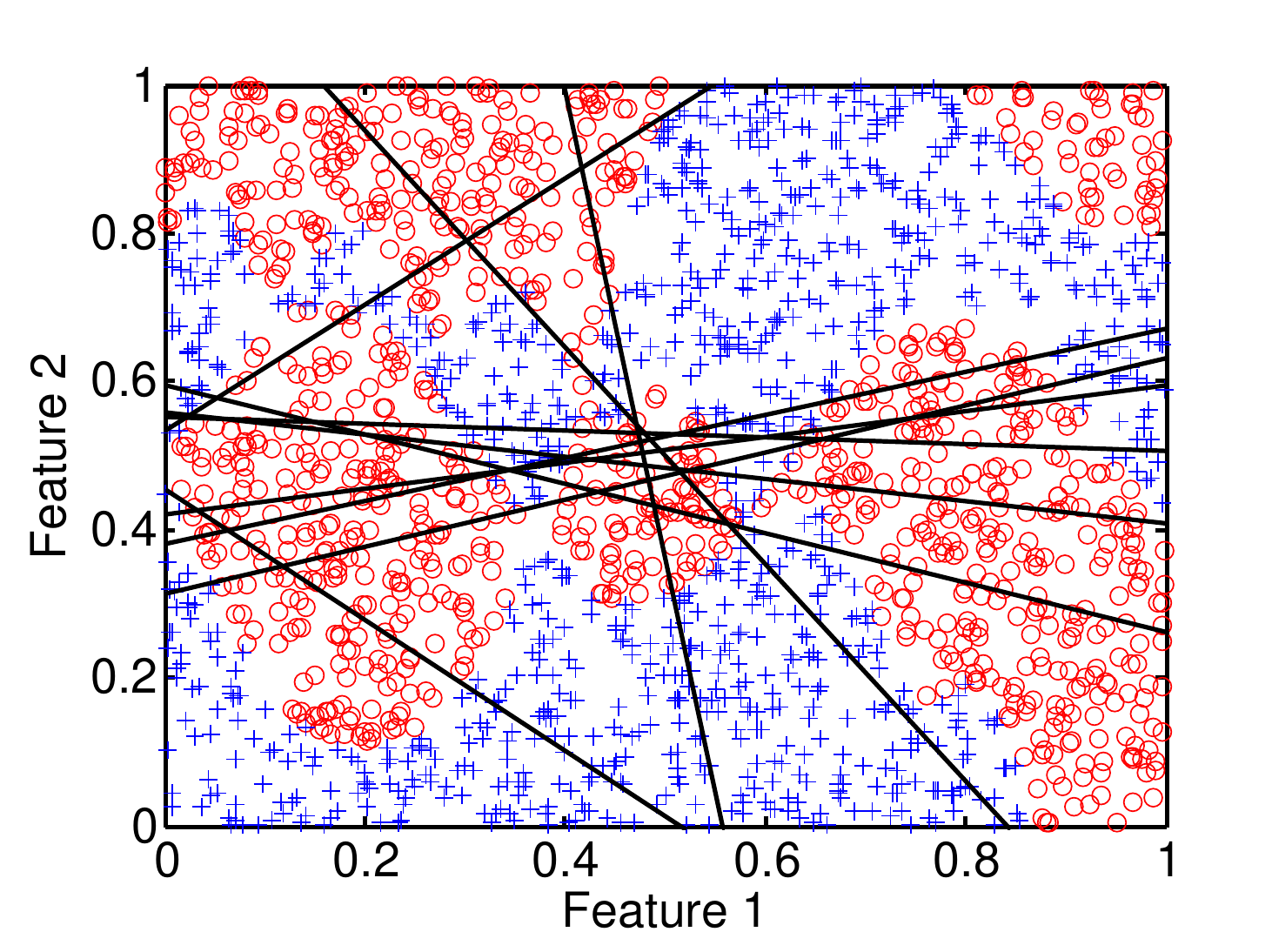}} 
	\subfigure[25 Perceptrons]{\includegraphics[width=3.2in,clip=]{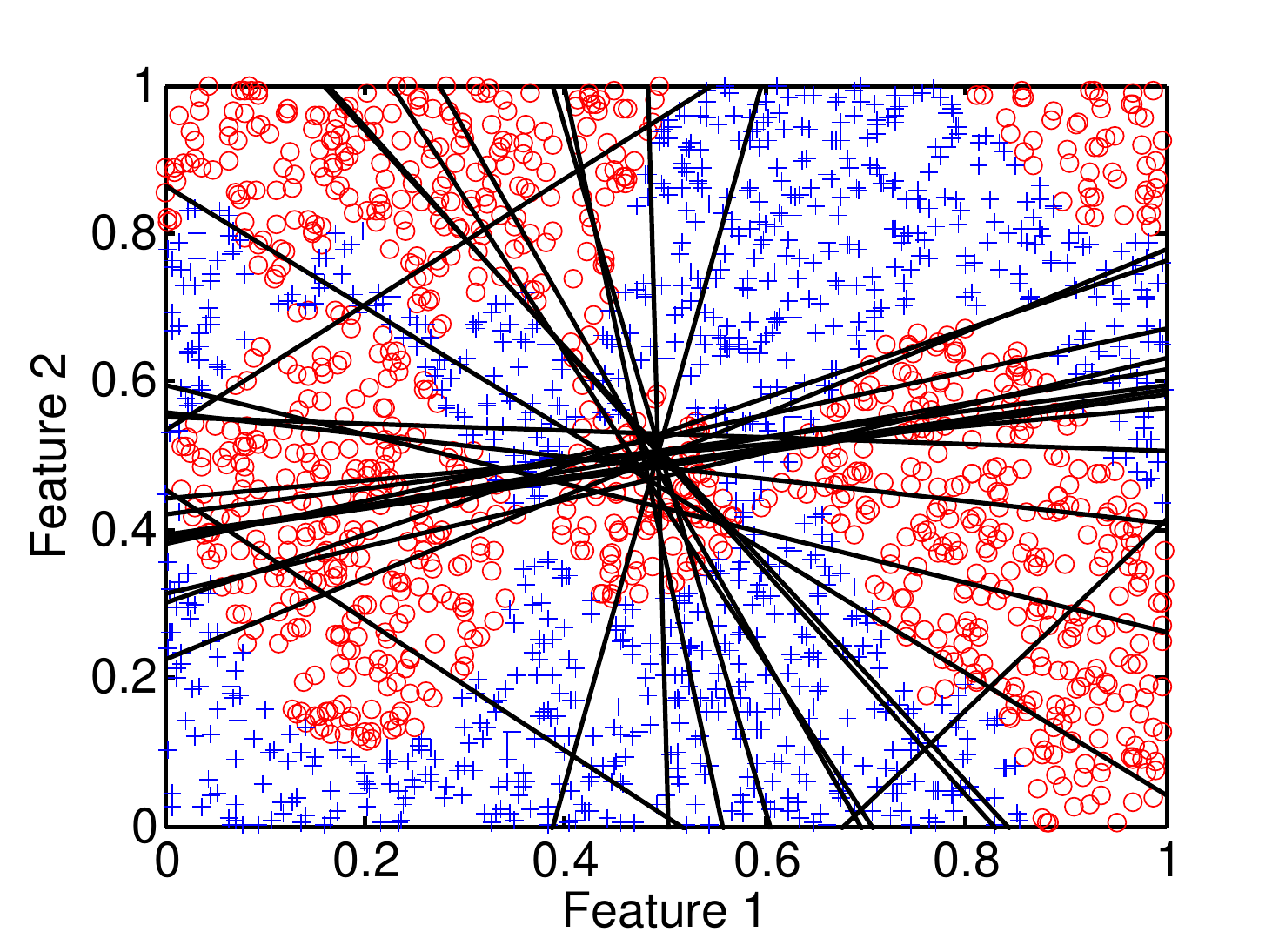}} 
	\subfigure[50 Perceptrons]{\includegraphics[width=3.2in,clip=]{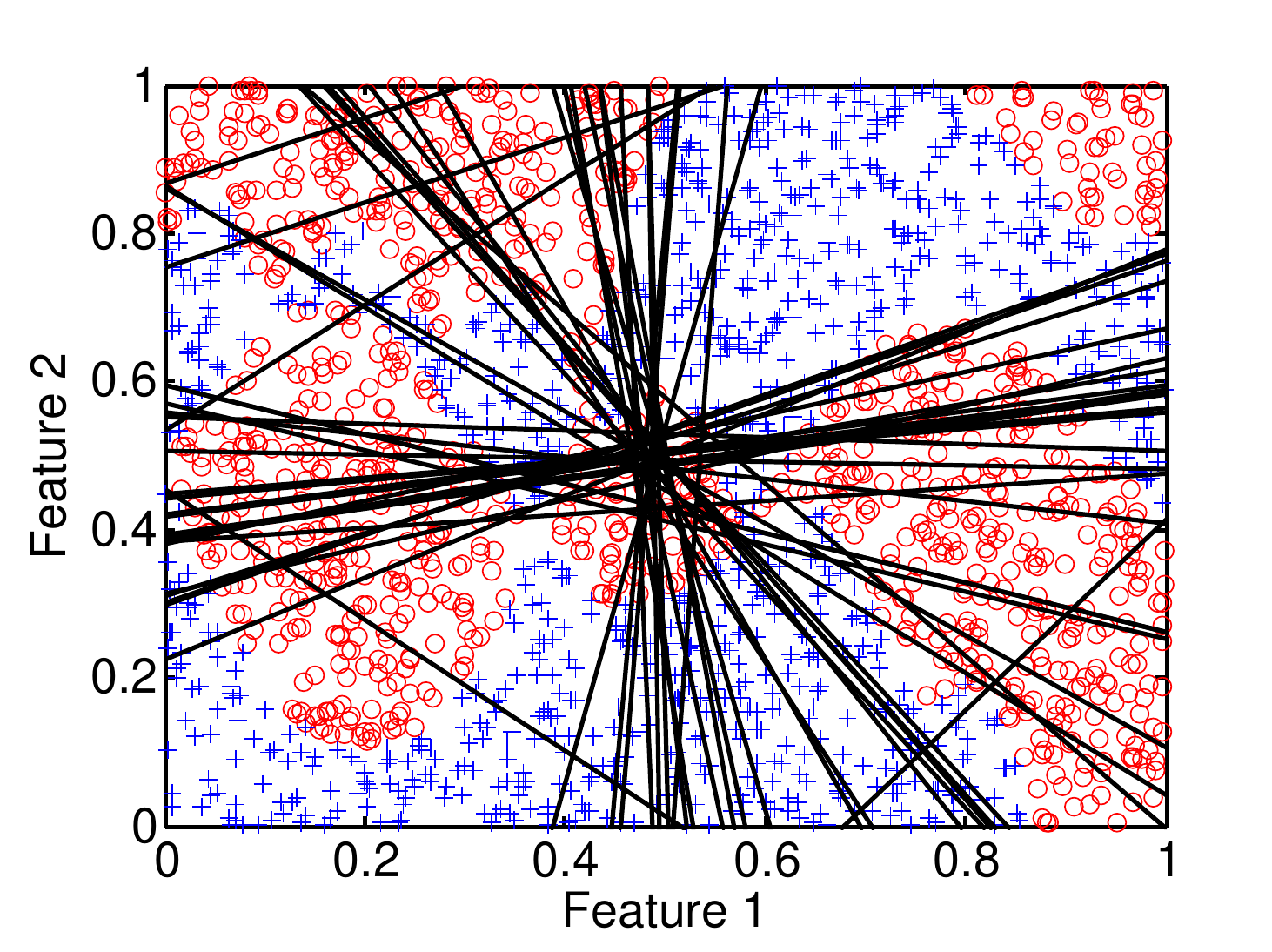}} 
	\subfigure[75 Perceptrons]{\includegraphics[width=3.2in,clip=]{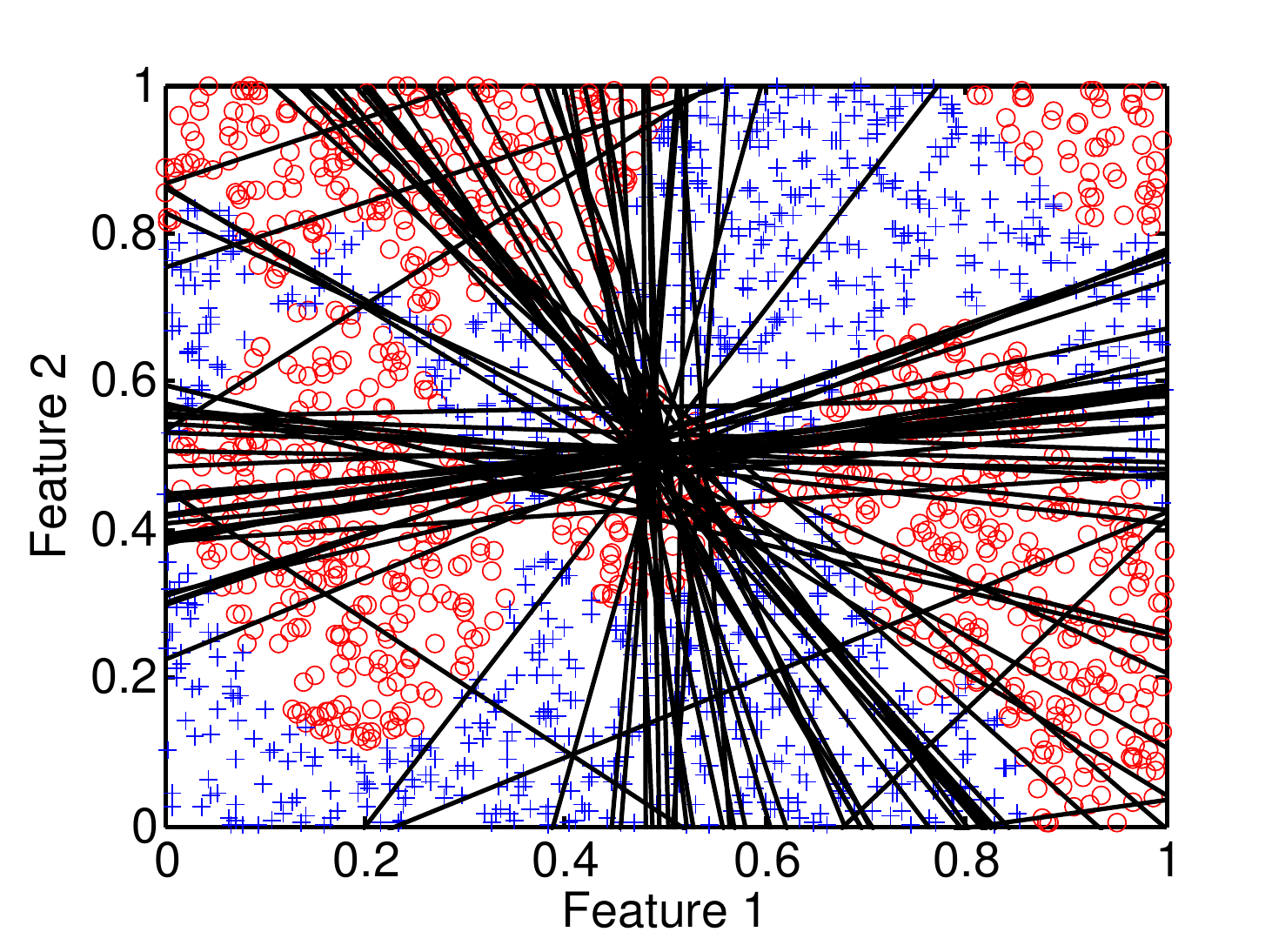}} 
	\subfigure[100 Perceptrons]{\includegraphics[width=3.2in,clip=]{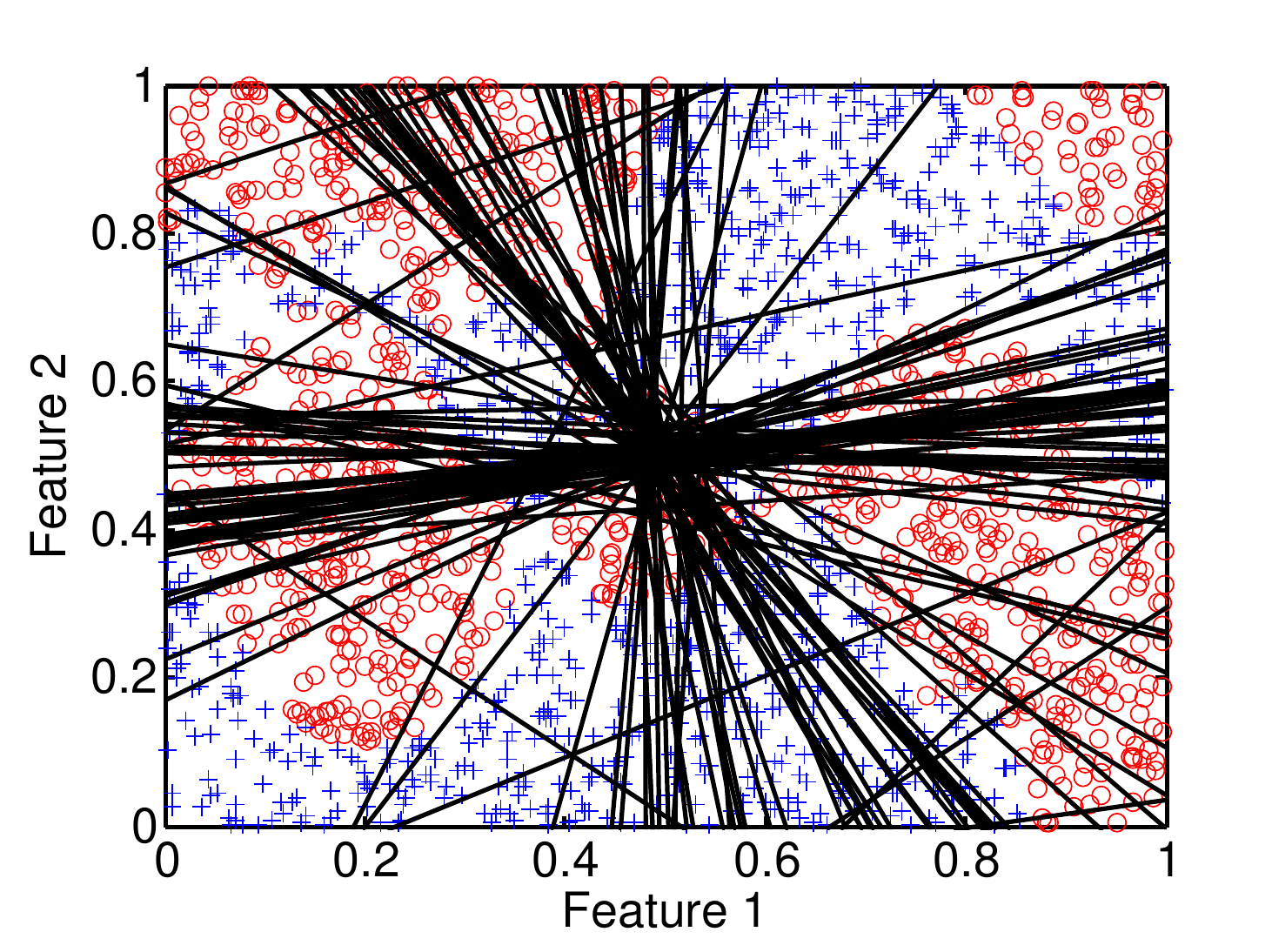}} 	
	
	\caption{Base classifiers generated during the overproduction phase. The Bagging technique is used to generate the pool of classifiers.}
	\label{fig:ensemblePerceptron}	  
\end{figure}

\begin{figure}[!ht]

	\centering
	\subfigure[5 Stumps]{\includegraphics[width=3.2in,clip=]{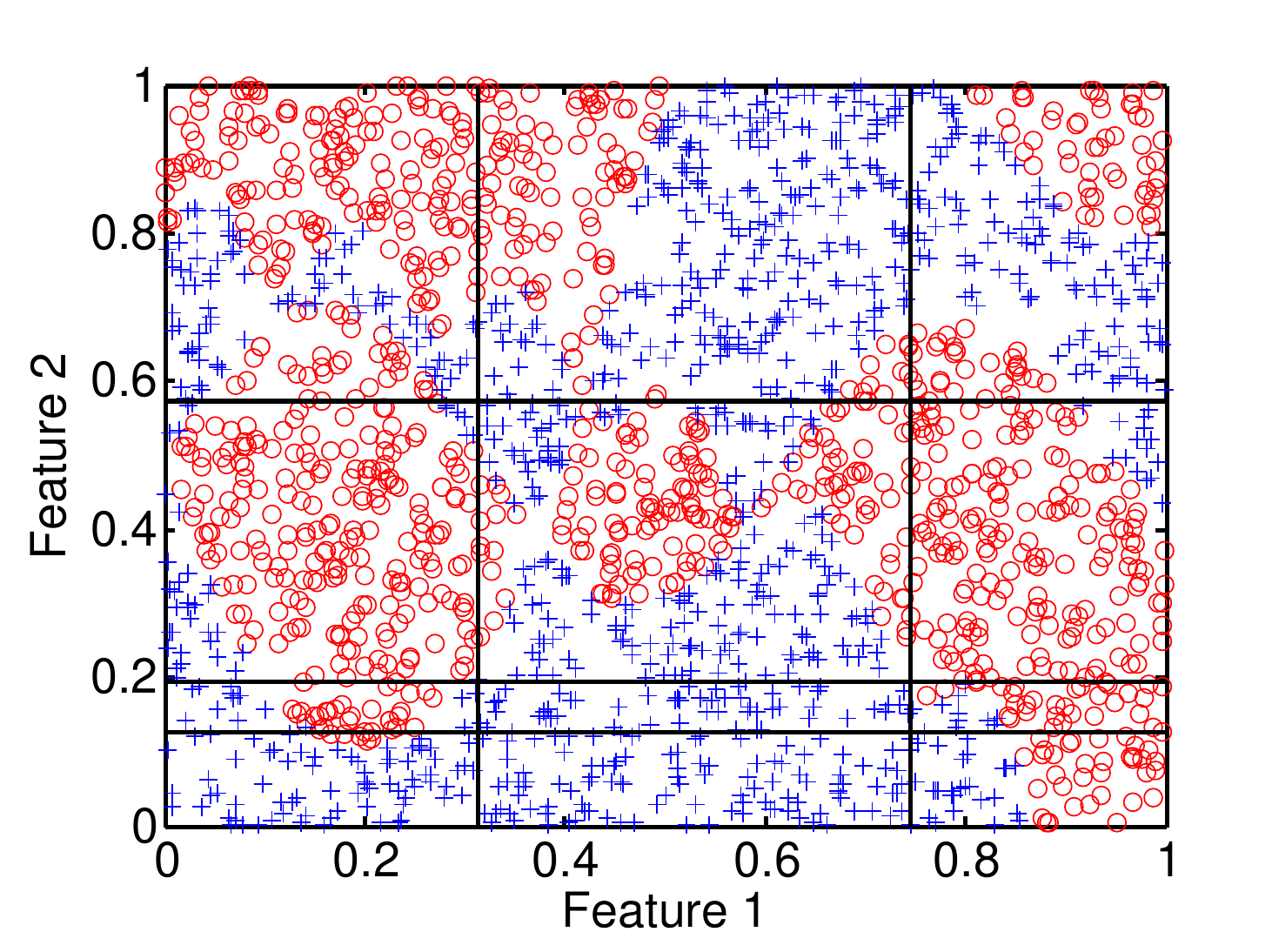}} 
	\subfigure[10 Stumps]{\includegraphics[width=3.2in,clip=]{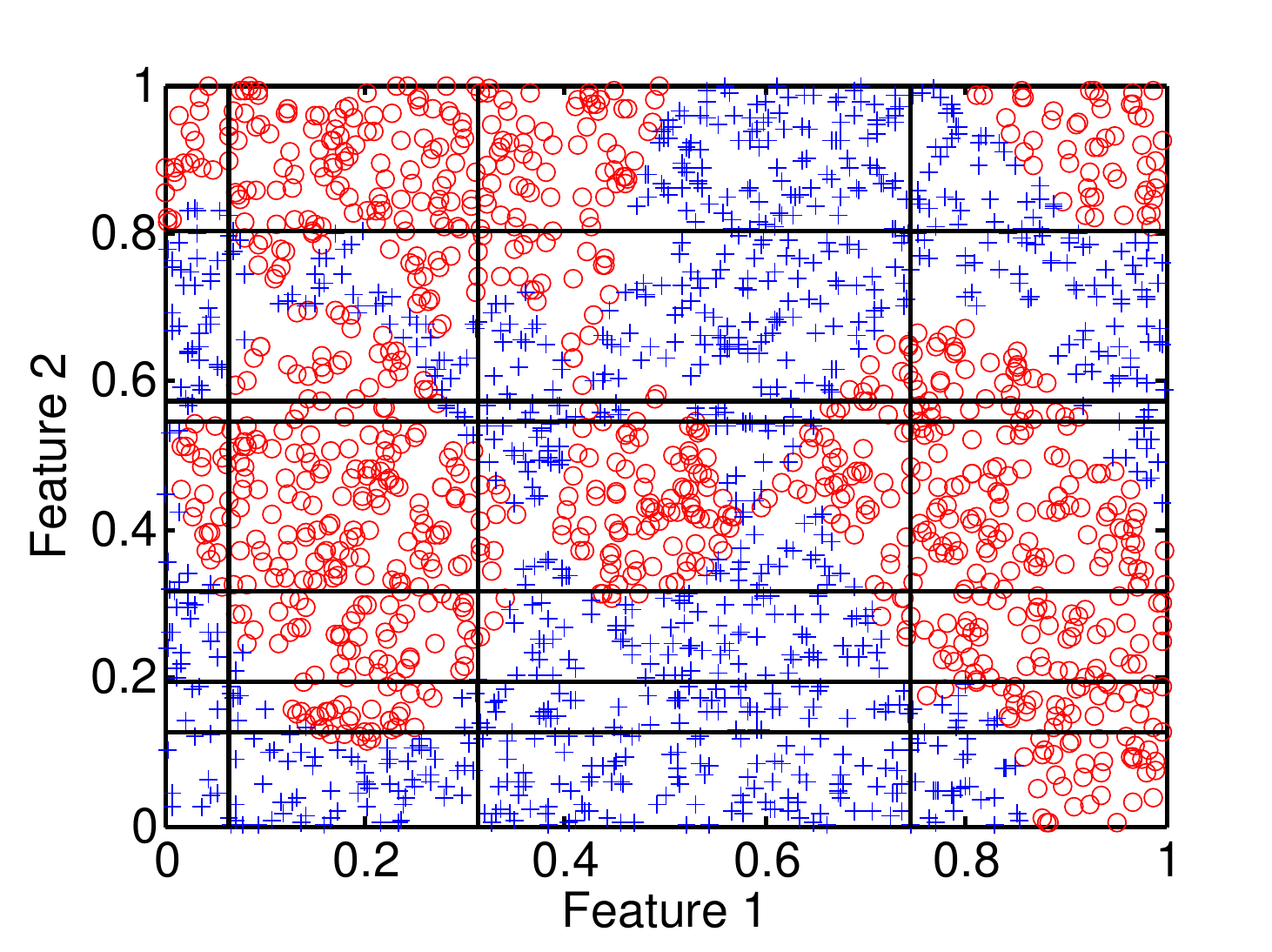}} 
	\subfigure[25 Stumps]{\includegraphics[width=3.2in,clip=]{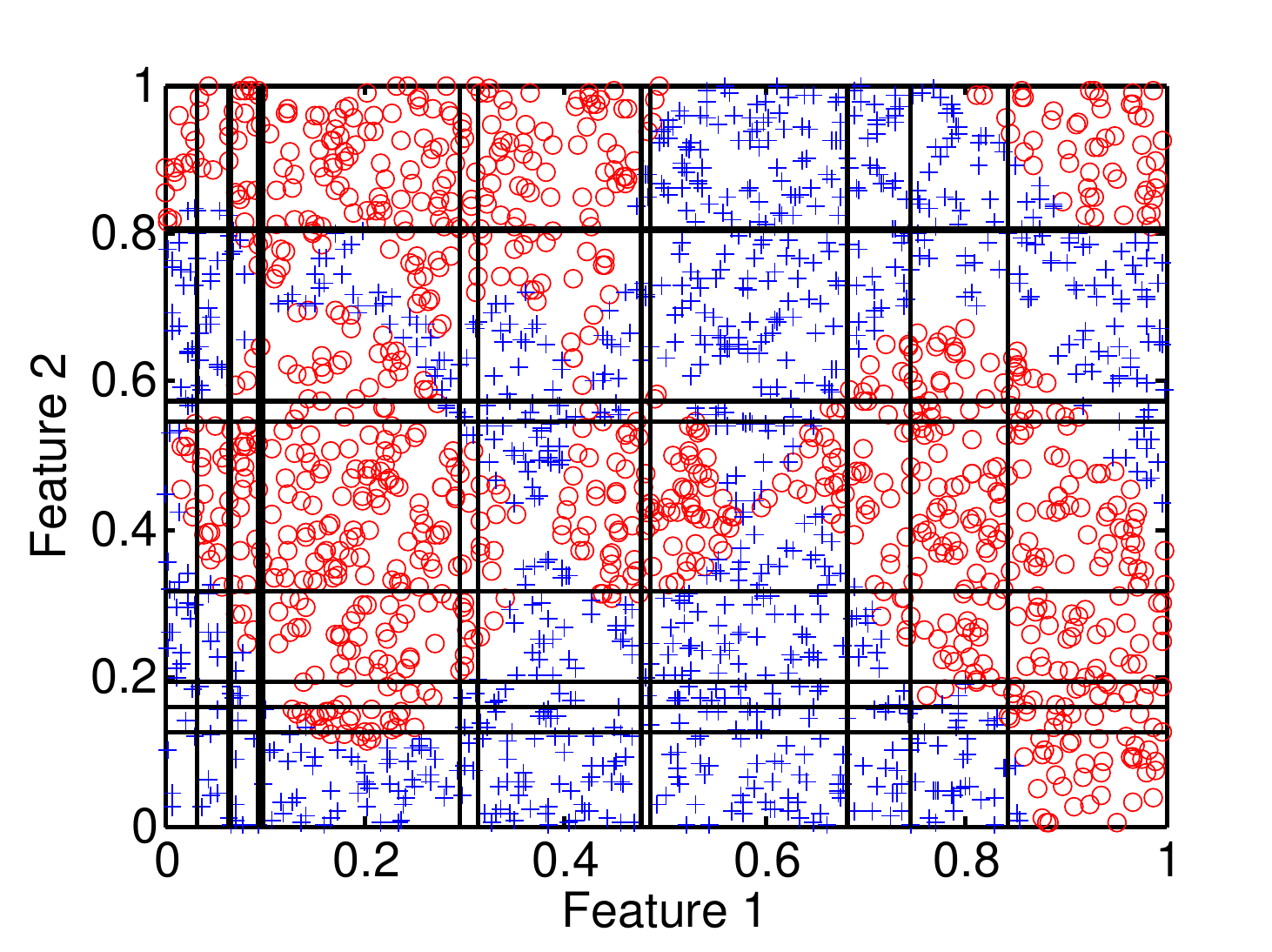}} 
	\subfigure[50 Stumps]{\includegraphics[width=3.2in,clip=]{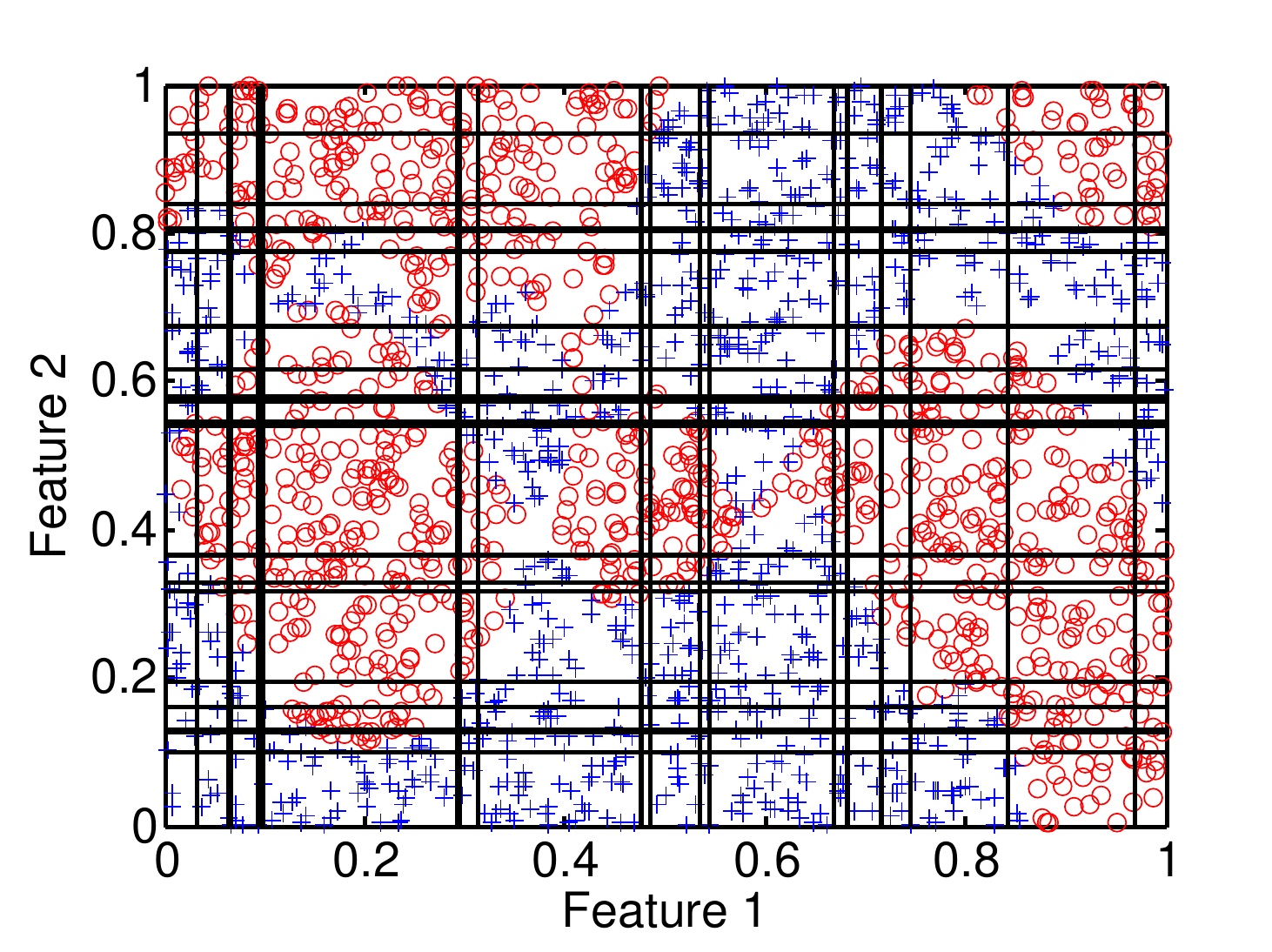}} 
	\subfigure[75 Stumps]{\includegraphics[width=3.2in,clip=]{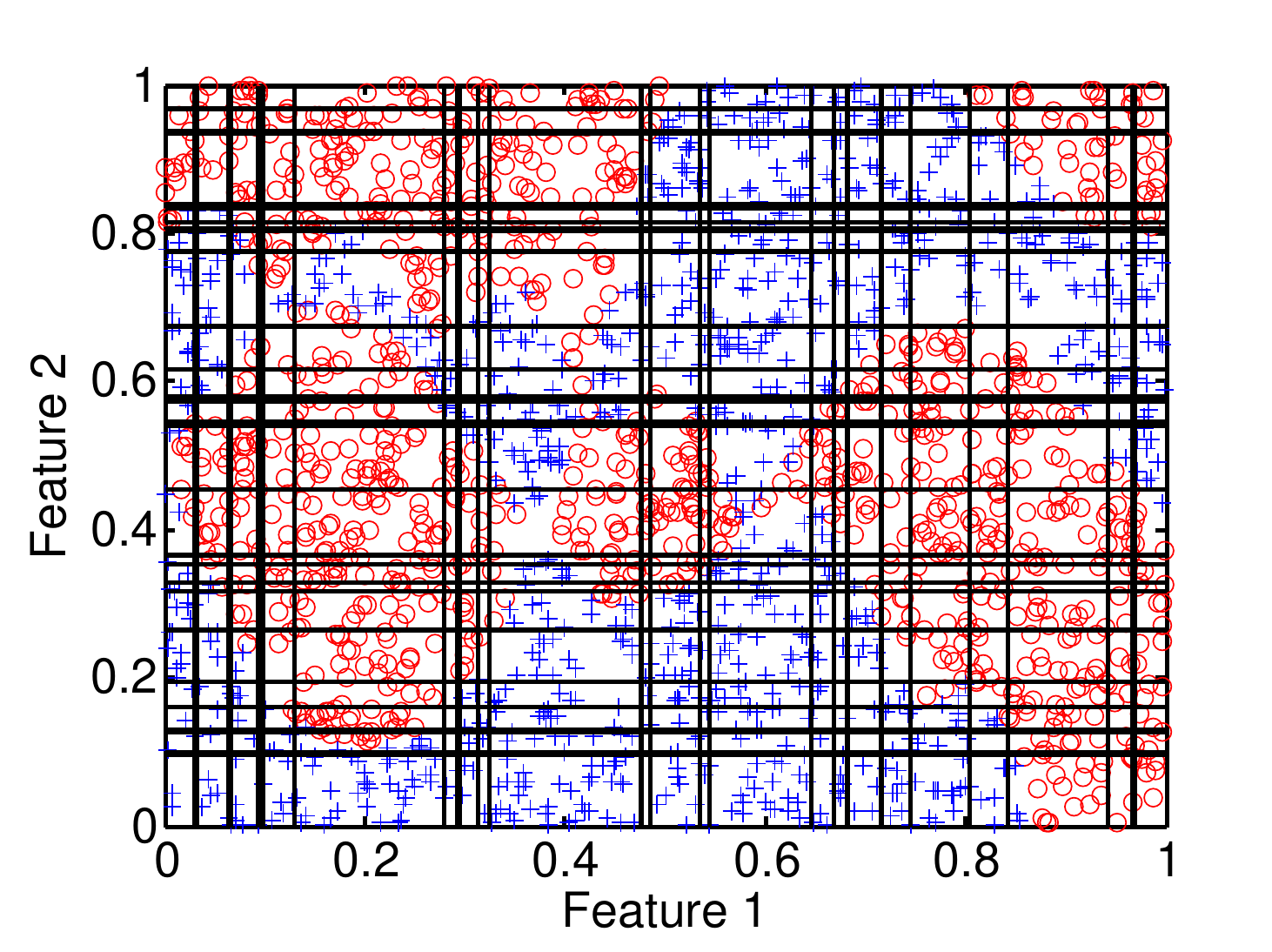}} 
	\subfigure[100 Stumps]{\includegraphics[width=3.2in,clip=]{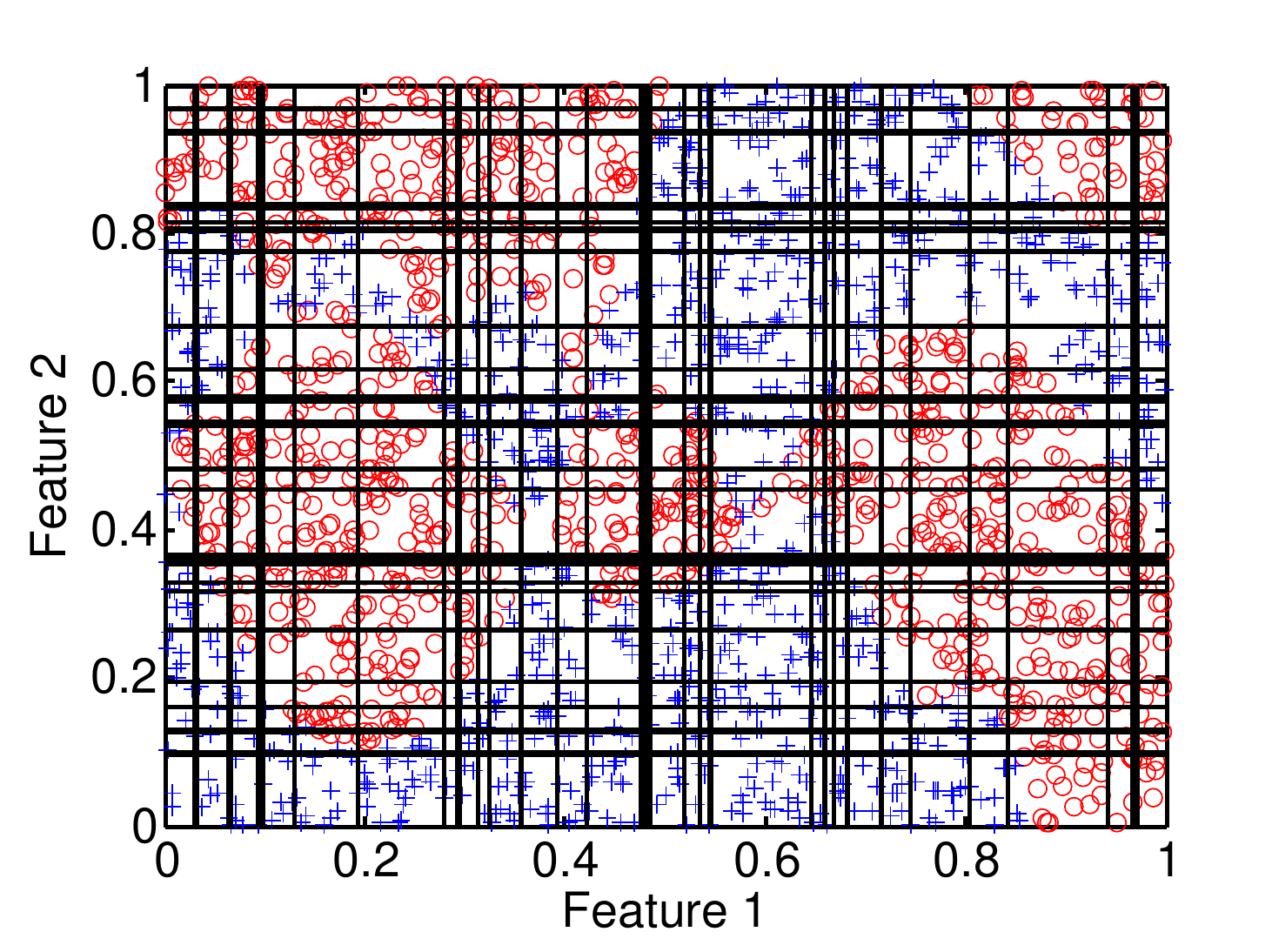}} 	
	
	\caption{Decision Stumps classifiers generated during the overproduction phase. The Bagging technique is used to generate the pool of classifiers.}
	\label{fig:ensembleStumps}	  
\end{figure}

Figures~\ref{fig:ensemblePerceptron} and~\ref{fig:ensembleStumps} illustrate the pool of classifiers generated with bagging using Perceptrons and Decision Stumps, respectively. We consider a pool of 5, 10, 25, 50, 75 and 100 base classifiers. Considering a pool size of 100 classifiers, we can see that most of the classifiers are in the same region. Thus, we believe the majority of classifiers are redundant. This can be explained by the fact we used bagging for the generation of the pool. In the bagging technique, the bootstraps are randomly taken from the training data, and such, there is no guarantee that a high diversity pool will be achieved. The use of techniques such as the Random Oracle~\cite{KunchevaR07}, may be considered in the future as an alternative for the generation of the pool in order to achieve higher diversity at the pool level.

%---------------------------------------------------------------------------------------------------------------------------

\subsection{Sample Selection Mechanism: consensus threshold $h_{c}$}

\begin{figure}[!ht]
	\centering
	\subfigure[Consensus threshold $h_{c} = 50\%$]{\includegraphics[width=3.2in,clip=]{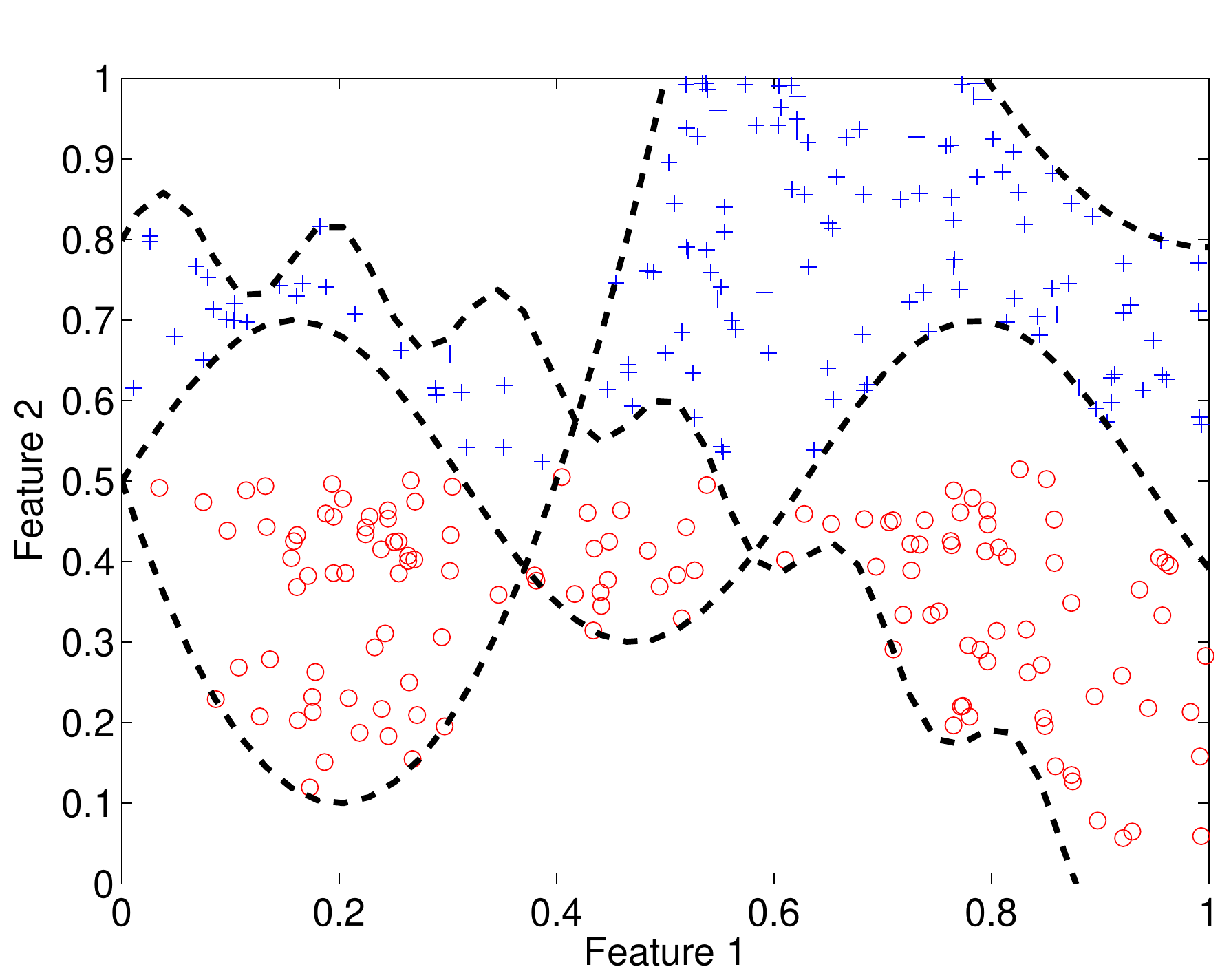}} 
	\subfigure[Consensus threshold $h_{c} = 60\%$]{\includegraphics[width=3.2in,clip=]{Images/SampleSelection/filteredDataset60PerceptronP2Border.pdf}} 
	\subfigure[Consensus threshold $h_{c} = 70\%$]{\includegraphics[width=3.2in,clip=]{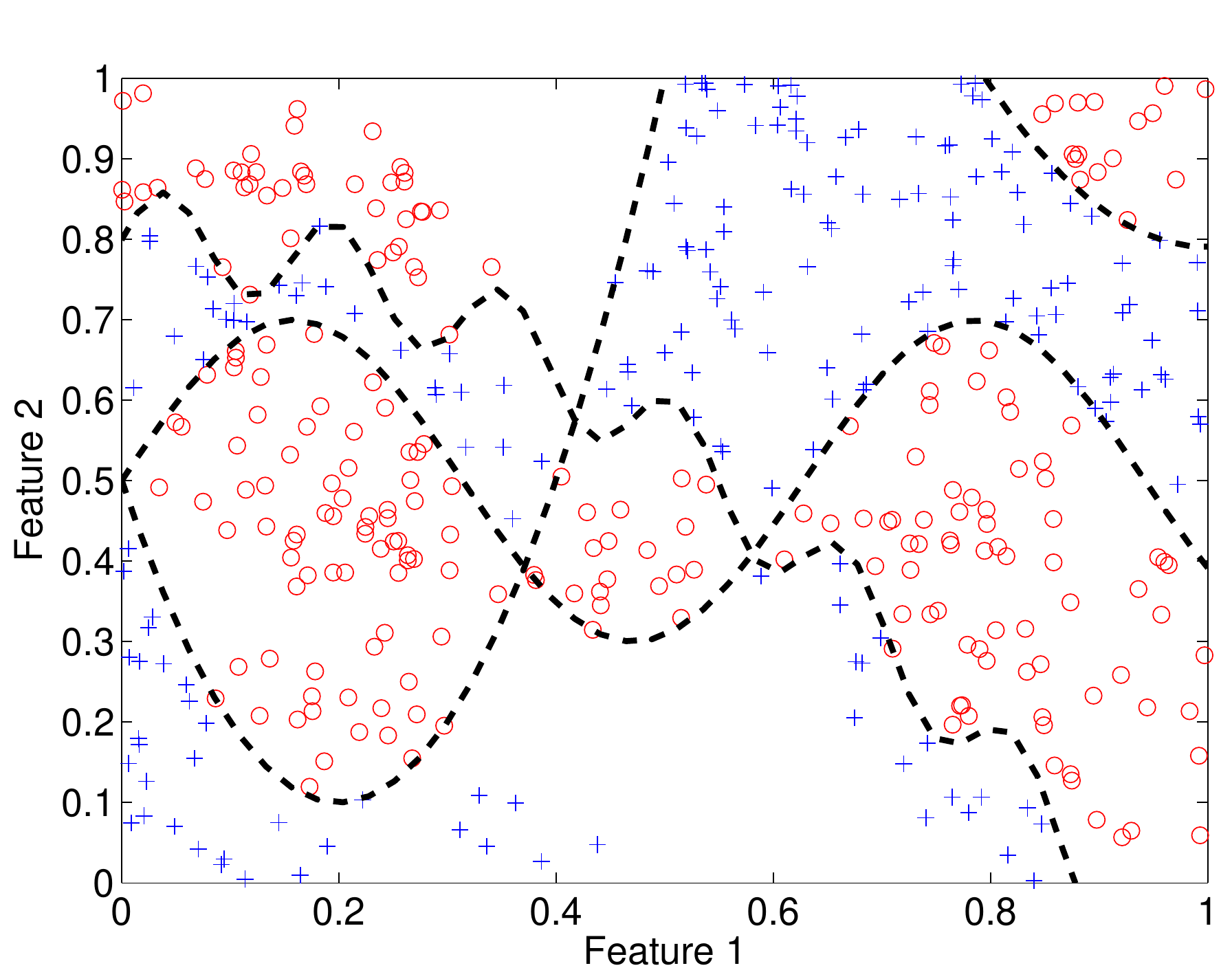}} 
	\subfigure[Consensus threshold $h_{c} = 80\%$]{\includegraphics[width=3.2in,clip=]{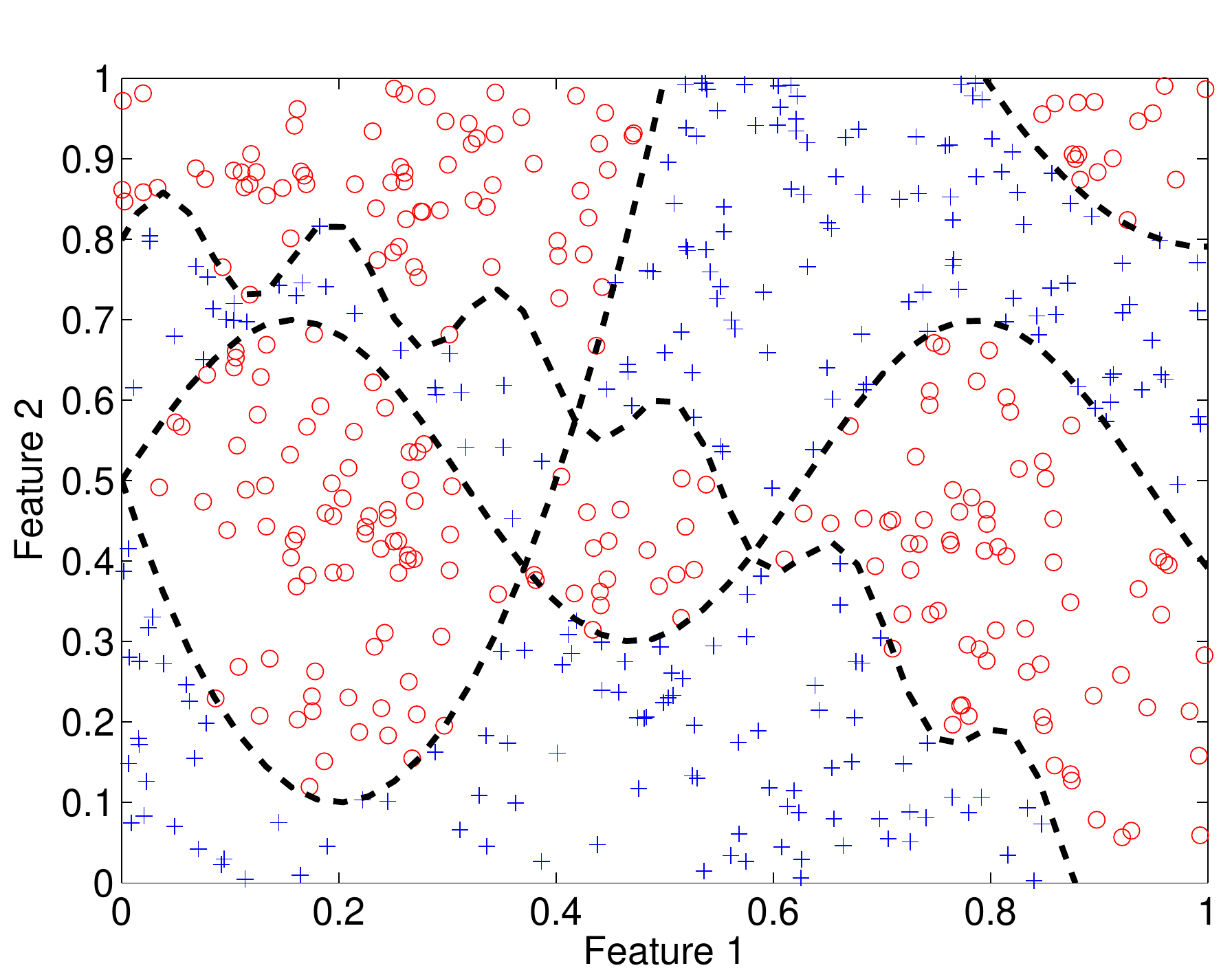}} 
	
	\caption{Meta-training dataset $\mathcal{T}_{\lambda}$ after the sample selection mechanism is applied. A pool composed of 100 Perceptrons is used.}
	\label{fig:sampleSelectionDataset}	  
\end{figure}

\begin{figure}[!ht]
	\centering
	\subfigure[Consensus threshold $h_{c} = 50\%$]{\includegraphics[width=3.2in,clip=]{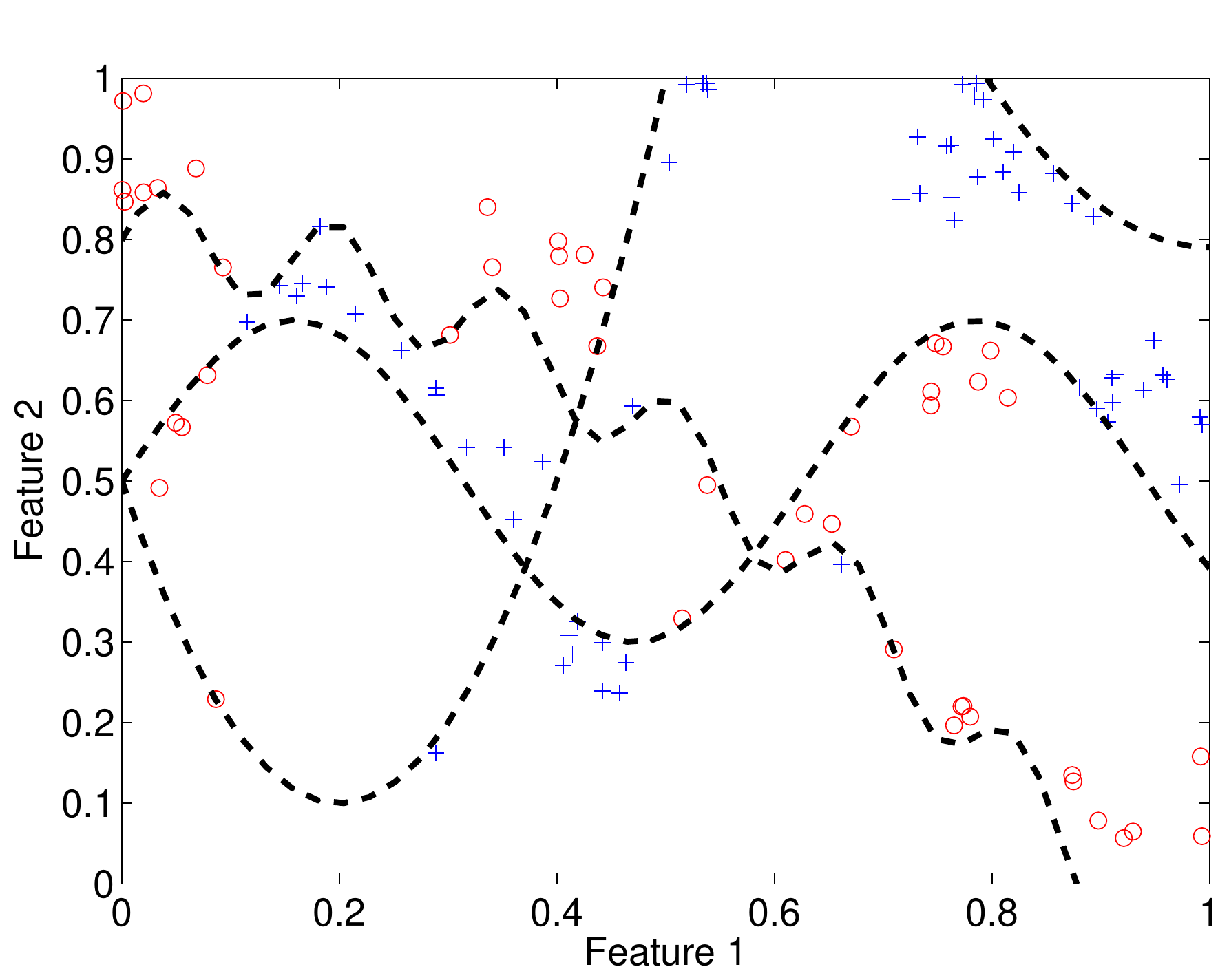}} 
	\subfigure[Consensus threshold $h_{c} = 50\%$]{\includegraphics[width=3.2in,clip=]{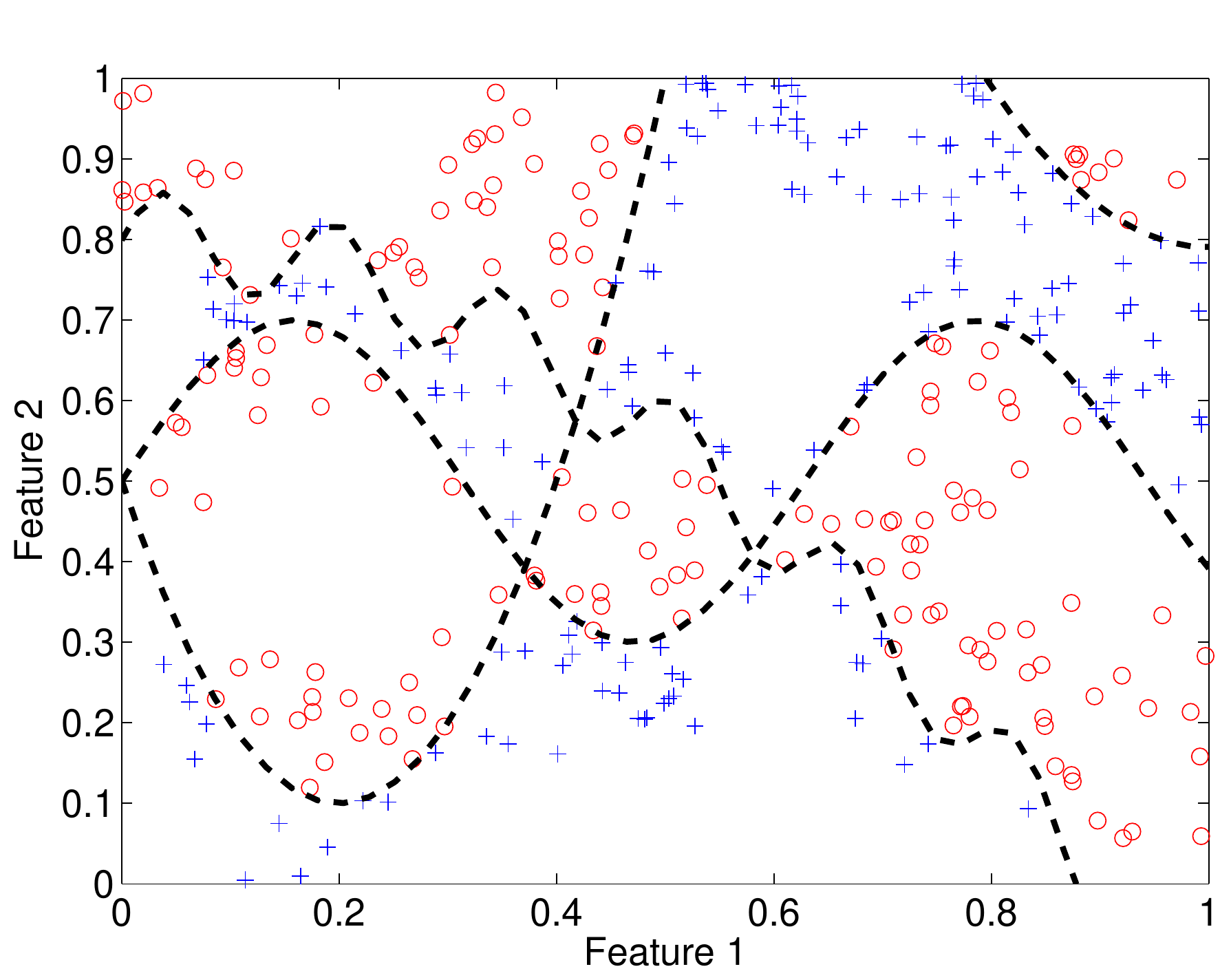}} 
	\subfigure[Consensus threshold $h_{c} = 50\%$]{\includegraphics[width=3.2in,clip=]{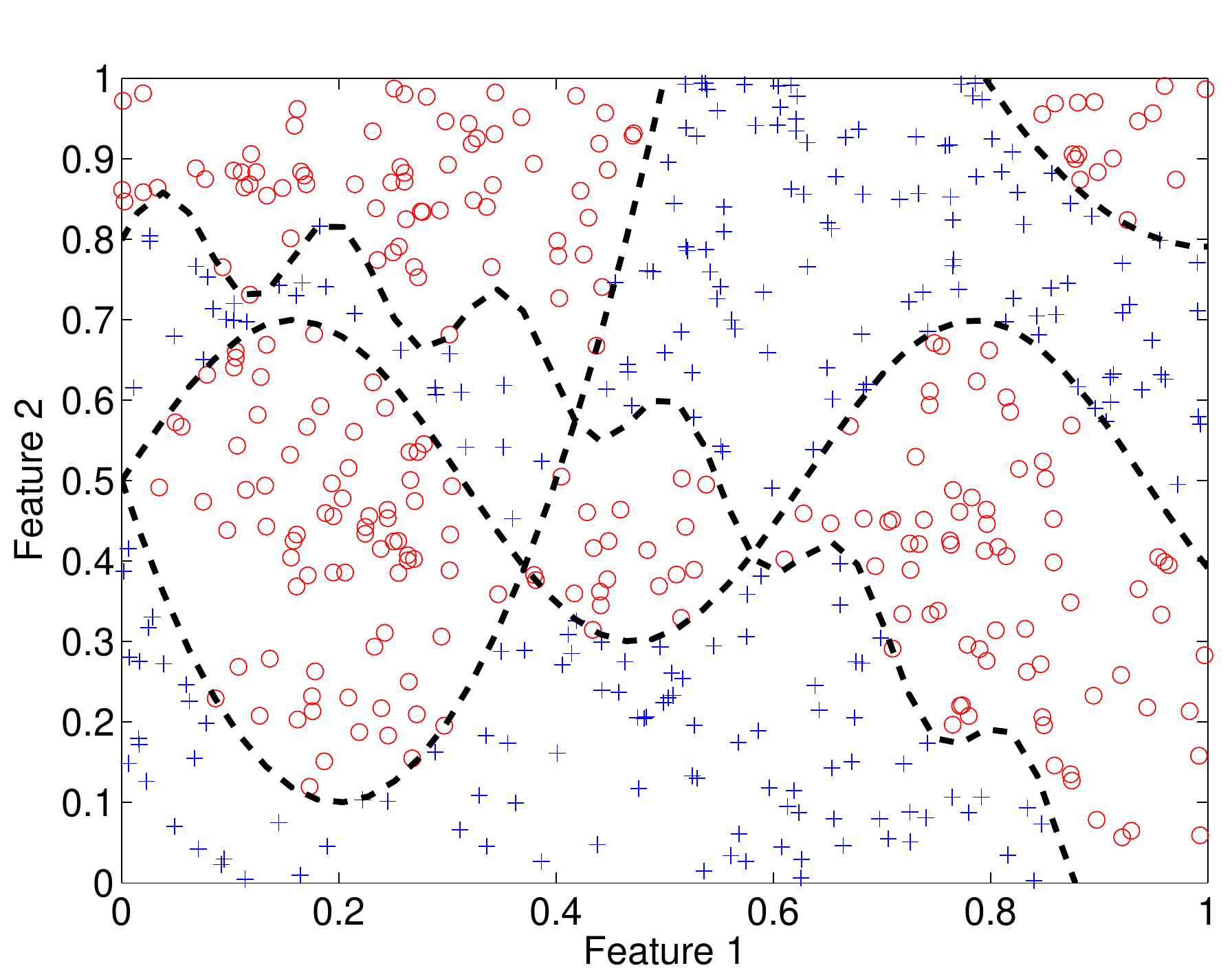}} 
	\subfigure[Consensus threshold $h_{c} = 50\%$]{\includegraphics[width=3.2in,clip=]{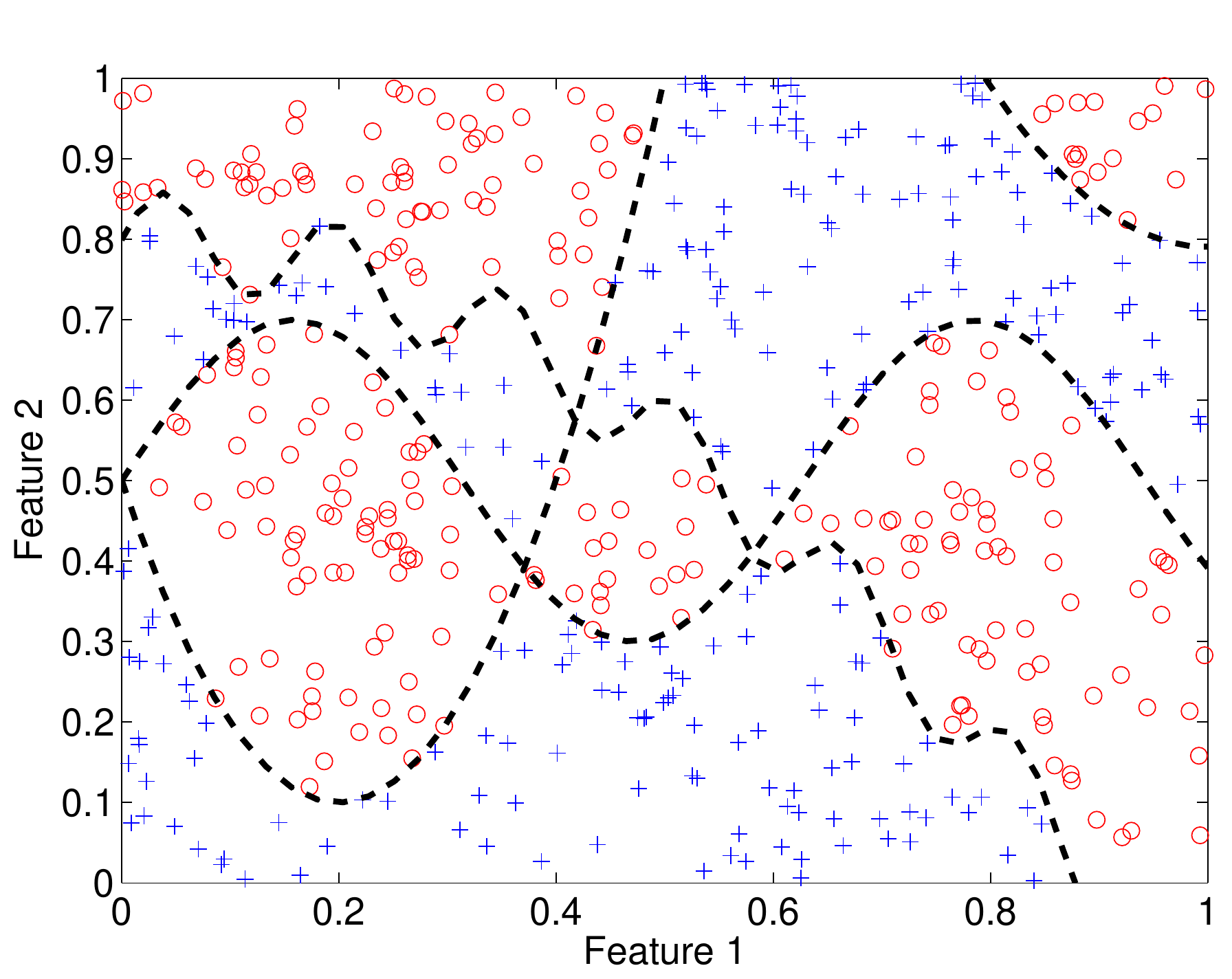}} 
	
	\caption{Meta-training dataset $\mathcal{T}_{\lambda}$ after the sample selection mechanism is applied. A pool composed of 100 Decision Stumps is used.}
	\label{fig:sampleSelectionDatasetStump}	  
\end{figure}

In this section, we show the results of the sample selection mechanism by varying the value of the threshold $h_{c}$. Since the sample selection mechanism depends on the base classifier (i.e., the consensus among the pool), we show the result of the sample selection mechanism using both Perceptrons and Decision Stumps Figures~\ref{fig:sampleSelectionDataset} and~\ref{fig:sampleSelectionDatasetStump} respectively. 

Samples close to the decision boundary are the ones more likely to be selected for the training of the meta-classifier. Hence, the sample selection mechanism focuses on samples that are close to the decision boundaries thus, are harder to predict its correct label. This principle is similar to the support vectors in the SVM, where samples close to the decision boundaries are used to achieve the best separating hyperplanes. In our case, the samples close to the decision boundary are used to train the meta-classifier in order to distinguish between a competent classifier from an incompetent one in cases where a disagreement exists between the base classifiers in the pool.

%We stopped the consensus threshold at $h_{c} = 80\%$, and all samples were selected for the training of the meta-classifier. This fact can be explained by the complexity of the decision frontier of P2 problem, including  multiple class centers. For the majority of samples there is no consensus in the pool for the correct label. Hence, the majority of instances are passed down for the training of the meta-classifier.

\subsection{Size of the dynamic selection dataset (DSEL)}
\label{sec:dselsize}

Figure~\ref{fig:ensemblePerceptron} shows the dynamic selection dataset (DSEL) generated with different sizes. The figures show the exact distributions of the dataset DSEL used to evaluate the performance of the META-DES framework according to its size (Section~\ref{sec:effectDSEL}). The size of the DSEL has a significant impact on the performance of the META-DES framework~\ref{fig:2DValidation}. This can be explained by the fact the meta-features are extracted based on the neighborhood of the query sample $\mathbf{x}_{j,test}$ projected in DSEL. 

\begin{figure}[!ht]
	
	\centering
	\subfigure[DSEL 50 Samples]{\includegraphics[width=3.2in,clip=]{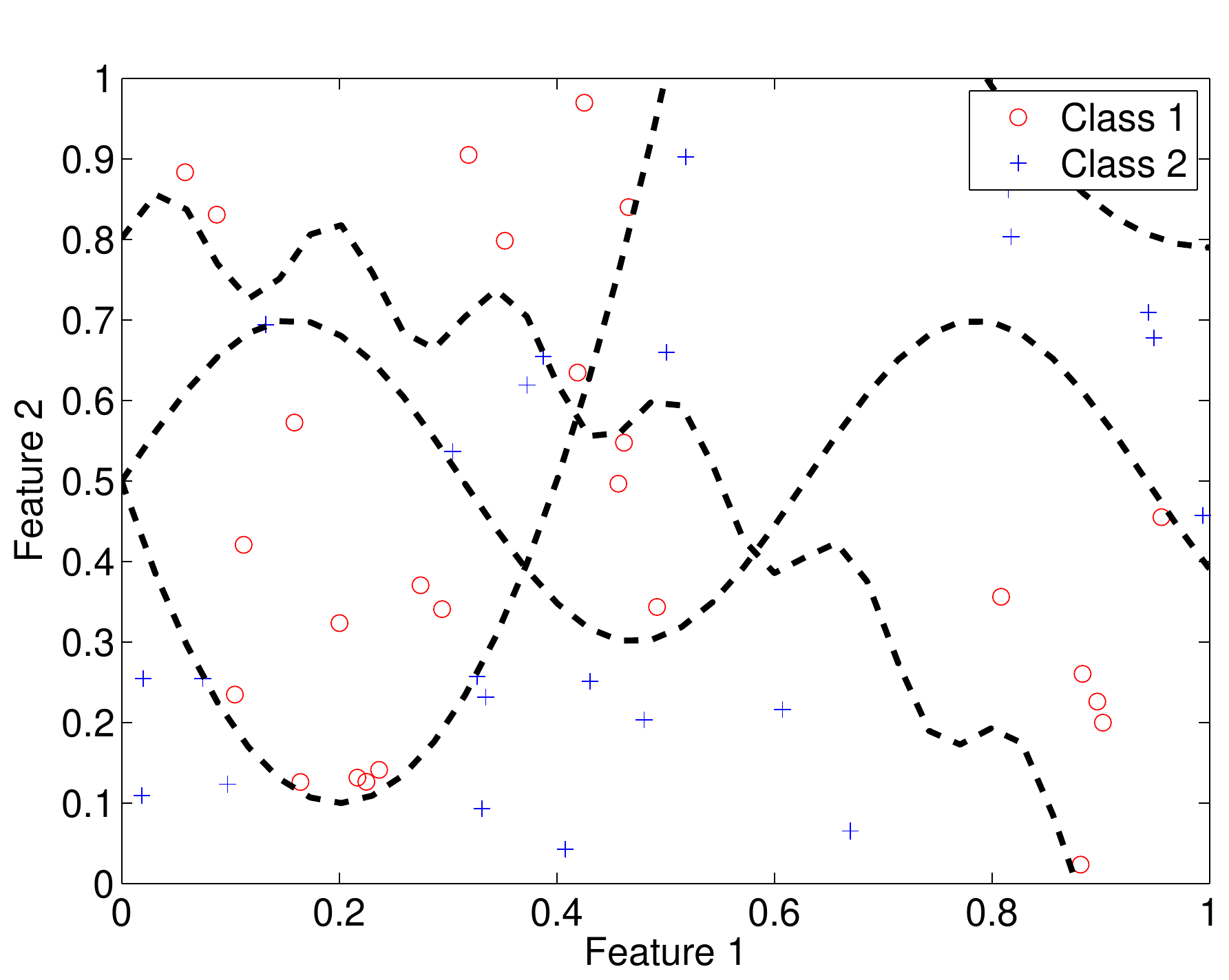}} 
	\subfigure[DSEL 100 Samples]{\includegraphics[width=3.2in,clip=]{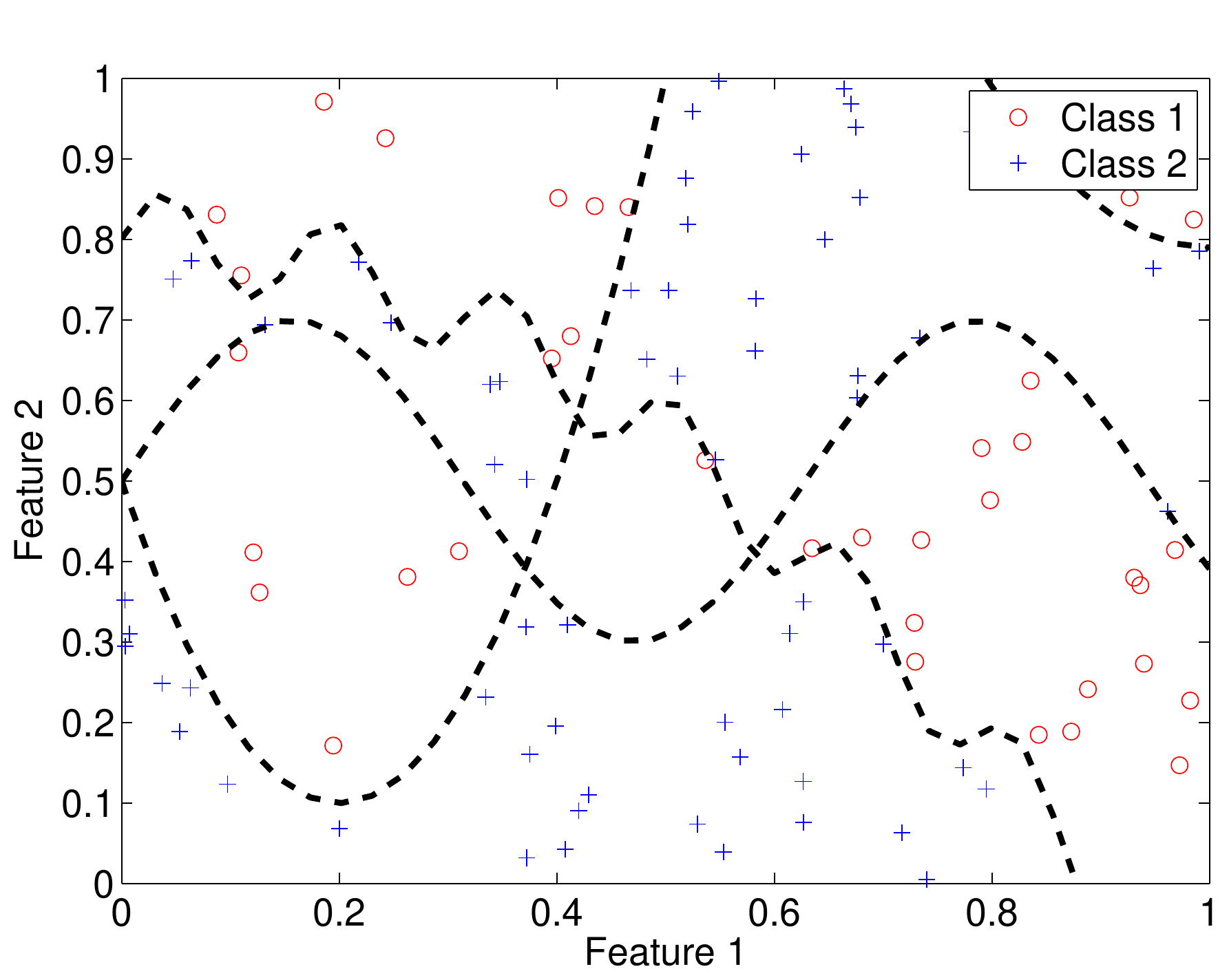}} 
	\subfigure[DSEL 150 Samples]{\includegraphics[width=3.2in,clip=]{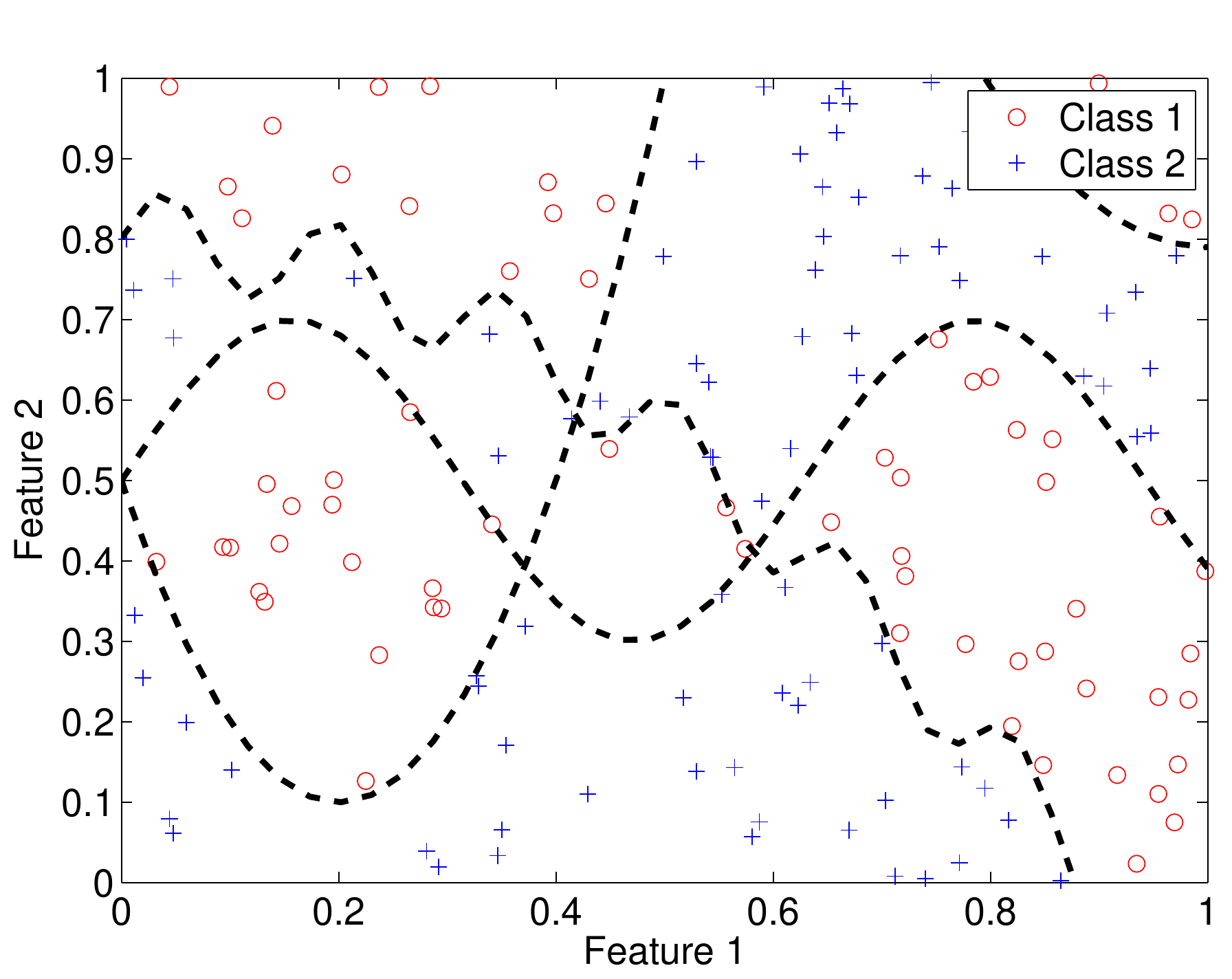}} 
	\subfigure[DSEL 200 Samples]{\includegraphics[width=3.2in,clip=]{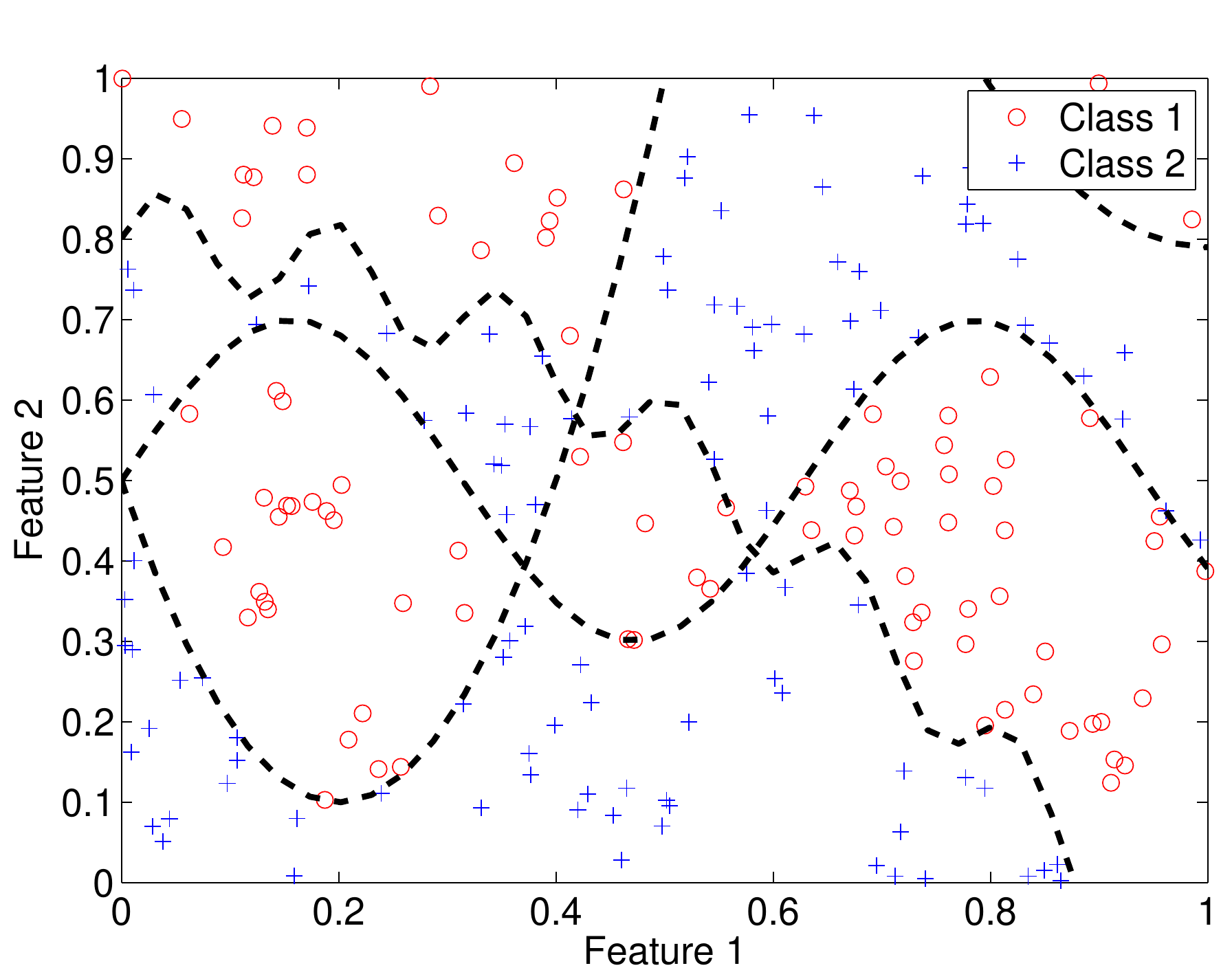}} 
	\subfigure[DSEL 250 Samples]{\includegraphics[width=3.2in,clip=]{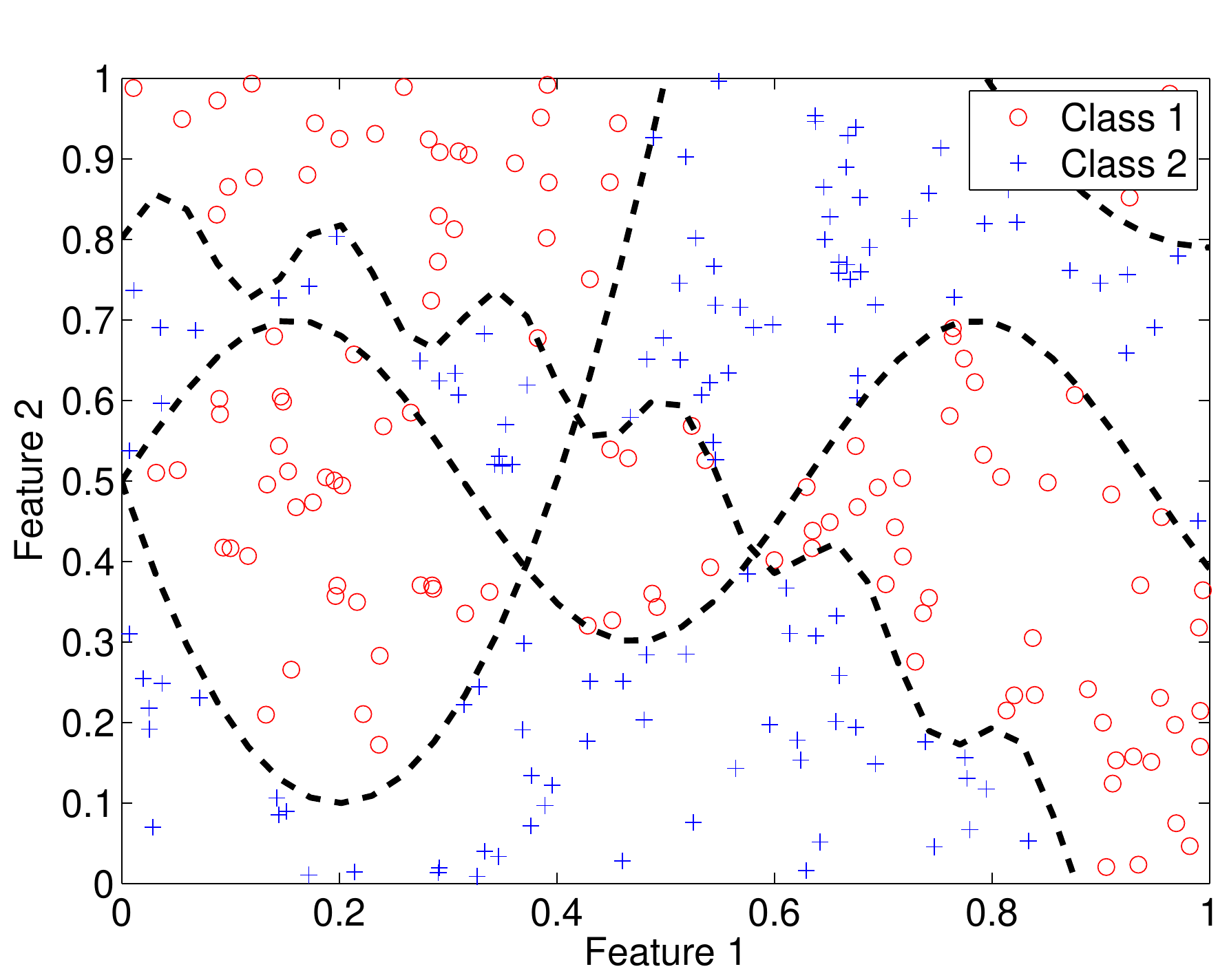}} 
	\subfigure[DSEL 500 Samples]{\includegraphics[width=3.2in,clip=]{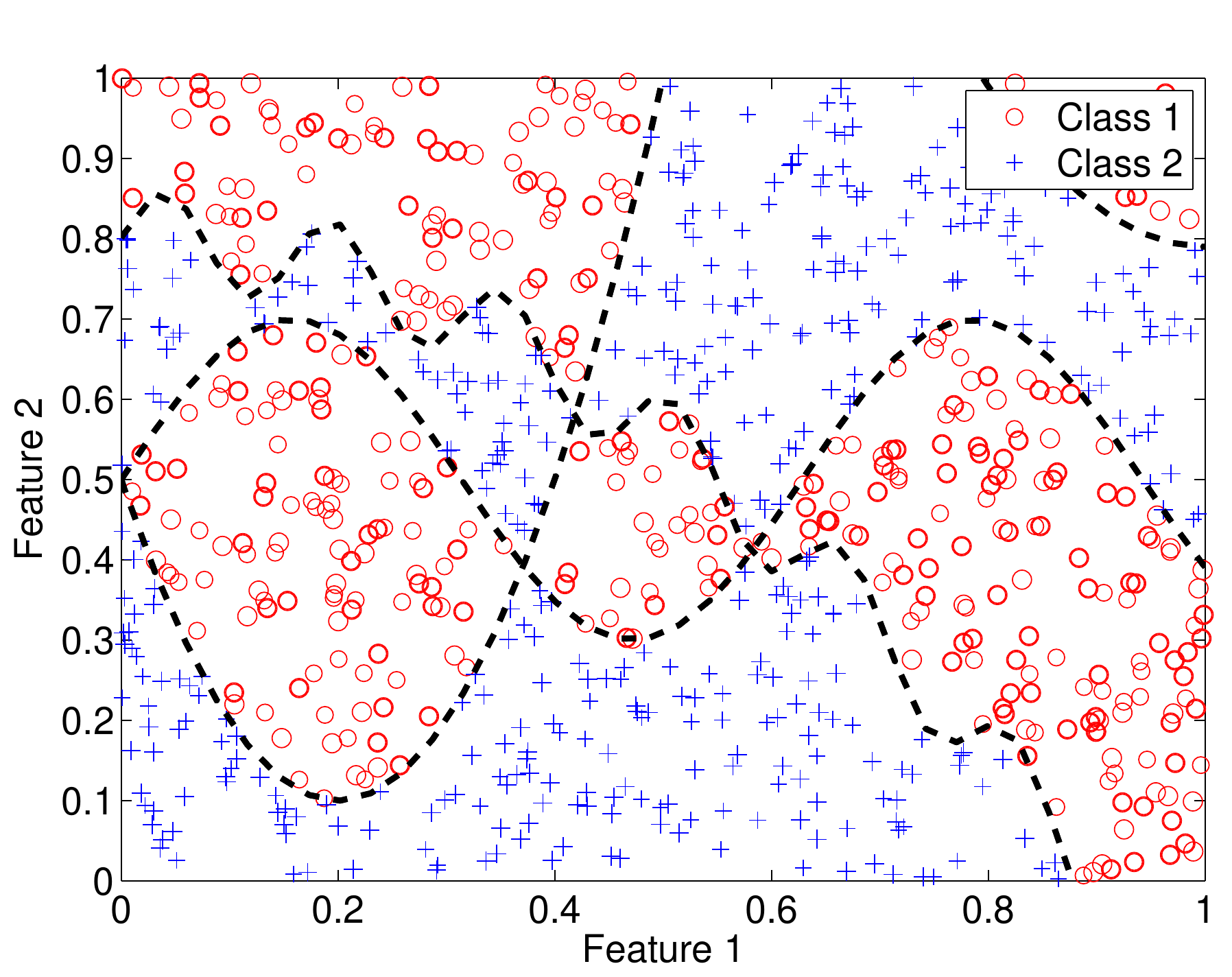}} 	
	%\subfigure[DSEL 1000 Samples]{\includegraphics[width=3.2in,clip=]{Images/DSEL/DSEL1000P2Border.eps}} 	
	
	\caption{Distributions of the dynamic selection dataset (validation), used to extract the meta-features during the generalization phase of the system.}
	\label{fig:DSEL}	  
\end{figure}

\bibliographystyle{elsarticle-num}
\bibliography{Example}

\end{document}